\RequirePackage{fix-cm}
\documentclass[twocolumn]{svjour3}          
\smartqed  
\usepackage{graphicx}

\usepackage{xspace}
\newcommand{\sys}{\textsc{VolcanoML}\xspace}
\newcommand{\ausk}{\texttt{auto-sklearn}\xspace}
\newcommand{\tpot}{\texttt{TPOT}\xspace}

\usepackage{subfigure}
\usepackage{multirow}
\usepackage{bm}
\usepackage{bbm}
\usepackage{color, soul}
\usepackage{booktabs}
\usepackage{amsmath}
\usepackage{amsfonts}
\usepackage{listings}
\usepackage{xcolor}
\usepackage{multibib}
\newcites{A}{References}
\usepackage{enumitem}

\usepackage{hyperref}
\usepackage{wrapfig}

\usepackage[linesnumbered,vlined,ruled,noend]{algorithm2e}

\newcommand{\para}[1]{{\vspace{2pt} \bf \noindent #1 \hspace{1pt}}}
\definecolor{codegray}{rgb}{0.5,0.5,0.5}

\begin{document}

\title{Efficient End-to-End AutoML via Scalable Search Space Decomposition (Extended Paper)
}


\author{Yang Li$^*$, Yu Shen$^*$, Wentao Zhang$^*$, Ce Zhang$^{\dag}$, Bin Cui}
\authorrunning{Yang Li, Yu Shen, Wentao Zhang, Ce Zhang, Bin Cui} 

\institute{$^{*}$Yang Li \at
         Institute: School of CS \& Key Laboratory of High Confidence Software Technologies (MOE), Peking University \\
         Institute: Tencent data platform, TEG, Tencent inc. \\ \email{liyang.cs@pku.edu.cn}
           \and
           $^{*}$Yu Shen \at
            Institute: School of CS \& Key Laboratory of High Confidence Software Technologies (MOE), Peking University \\ \email{shenyu@pku.edu.cn}
           \and
            $^{*}$Wentao Zhang \at
            Institute: School of CS \& Key Laboratory of High Confidence Software Technologies (MOE), Peking University \\
            Institute: Tencent data platform, TEG, Tencent inc. \\ \email{wentao.zhang@pku.edu.cn}
           \and
           $^{\dag}$ Ce Zhang \at
           Institute: Department of Computer Science, ETH Z\"urich \\    \email{ce.zhang@inf.ethz.ch}
           \and
           Corresponding author: Bin Cui
           \at
            Institute: Department of Computer Science and Technology \& Key Laboratory of High Confidence Software Technologies (MOE), Peking University \\
           Tel: +86-10-62765821 \\ \email{bin.cui@pku.edu.cn}   
}

\date{Received: date / Accepted: date}

\maketitle

\begin{abstract}
End-to-end AutoML has attracted intensive interests from both academia and industry which automatically searches for ML pipelines in a space induced by feature engineering, algorithm/model selection, and hyper-parameter tuning.
Existing AutoML systems, however, suffer from scalability issues when applying to application domains with large, high-dimensional search spaces.
We present \sys, a scalable and extensible framework that facilitates systematic exploration of large AutoML search spaces.
\sys introduces and implements basic building blocks that decompose a large search space into smaller ones, and allows users to utilize these building blocks to compose an \emph{execution plan} for the AutoML problem at hand.
\sys further supports a Volcano-style \emph{execution model} -- akin to the one supported by modern database systems -- to execute the plan constructed.
Our evaluation demonstrates that, not only does \sys raise the level of expressiveness for search space decomposition in AutoML, it also leads to actual findings of decomposition strategies that are significantly more efficient than the ones employed by state-of-the-art AutoML systems such as \texttt{auto-sklearn}.
This paper is the extended version of the initial VolcanoML paper appeared in VLDB 2021.
\end{abstract}

\section{Introduction}
In recent years, researchers in the database community have been working on raising the level of abstractions of machine learning (ML) and integrating such functionality into today's data management systems~\cite{zhang2021facilitating,zhang2022towards},
e.g., SystemML~\cite{Ghoting2011}, SystemDS~\cite{Boehm2019}, Snorkel~\cite{snorkel}, ZeroER~\cite{ZeroER}, TFX~\cite{baylor2017tfx,TFX}, Query 2.0~\cite{Query20}, Krypton~\cite{Krypton}, Cerebro~\cite{Cerebro}, ModelDB~\cite{modeldb}, MLFlow~\cite{mlflow}, DeepDive~\cite{DeepDive}, HoloClean~\cite{Holoclean},
EaseML~\cite{aguilar2021ease},
ActiveClean~\cite{ActiveClean}, and NorthStar~\cite{NorthStar}.
End-to-end AutoML systems~\cite{automl,DBLP:journals/corr/abs-1904-12054,automl_book} have been an emerging type of systems that has significantly raised the level of abstractions of building ML applications. 
Given an input dataset and a user-defined utility metric (e.g., validation accuracy), these systems automate the search of an end-to-end ML pipeline, including \textit{feature engineering}, \textit{algorithm/model selection}, and \textit{hyper-parameter tuning}. 
Open-source examples include \texttt{auto-sklearn}~\cite{feurer2015efficient}, \texttt{TPOT}~\cite{olson2019tpot}, and \texttt{hyperopt-sklearn}~\cite{komer2014hyperopt}, whereas most cloud service providers, e.g., Google, Microsoft, Amazon, etc., all provide their proprietary services on the cloud.
As machine learning has become an increasingly indispensable functionality integrated in modern data (management) systems, an efficient and effective end-to-end AutoML component also becomes increasingly important.

End-to-end AutoML provides a powerful abstraction to automatically navigate and search in a given complex search space. 
However, in our experience of applying state-of-the-art end-to-end AutoML systems in a range of real-world applications~\cite{bai2022autodc}, we find that such a system running fully automatically is rarely enough --- often, developing a successful ML application involves multiple iterations between a user and an AutoML system to iteratively improve the resulting ML artifact.

\noindent
\paragraph{\bf Motivating Practical Challenge.}

One such type of interaction, which inspires this work, is the \textit{\underline{enrichment of search space}}.
We observe that the default search space provided by state-of-the-art AutoML systems is often not enough in 
many applications. 
This was not obvious to us at all in the beginning and it is not until we finish building a range of real-world applications that we realize this via a set of concrete examples. 
For example, in one of our astronomy applications~\cite{app1}, the feature normalization function is domain-specific and not supported by most, if not all, AutoML systems. 
Similar examples can also be found when searching for suitable ML models via AutoML.
In one of our meteorology applications, we need to extend the models with meteorology-specific loss functions.
We saw similar problems when we tried to extend existing AutoML systems with pre-trained feature embeddings coming from TensorFlow Hub, to include include models that have been newly published on arXiv to enrich the Model Base~\cite{automl2}, or to support Cosine annealing as for tuning.

\paragraph{\bf Technical Challenge: Scalability over the Search Space.} 
``\textit{Why is it hard to extend the search space, as a user, in an end-to-end AutoML system}?'' The answer to this question is a complex one that is \textit{not} completely technical: 
some aspects are less technical such as engineering decisions and UX designs, however, there are also more \textit{technically fundamental} aspects. 
An end-to-end AutoML system contains an optimization algorithm that navigates a joint search space induced by \textit{feature engineering}, \textit{algorithm selection}, and \textit{hyper-parameter tuning}.
Because of this joint nature, the search space of end-to-end AutoML is complex and huge while the enrichment is only going to make it even larger.
As we will see, handling such a huge space is already challenging for existing systems, and further enriching it will make it even harder to scale.

\begin{figure}[htb]
\centering
\scalebox{0.68}{
\includegraphics[width=1\linewidth]{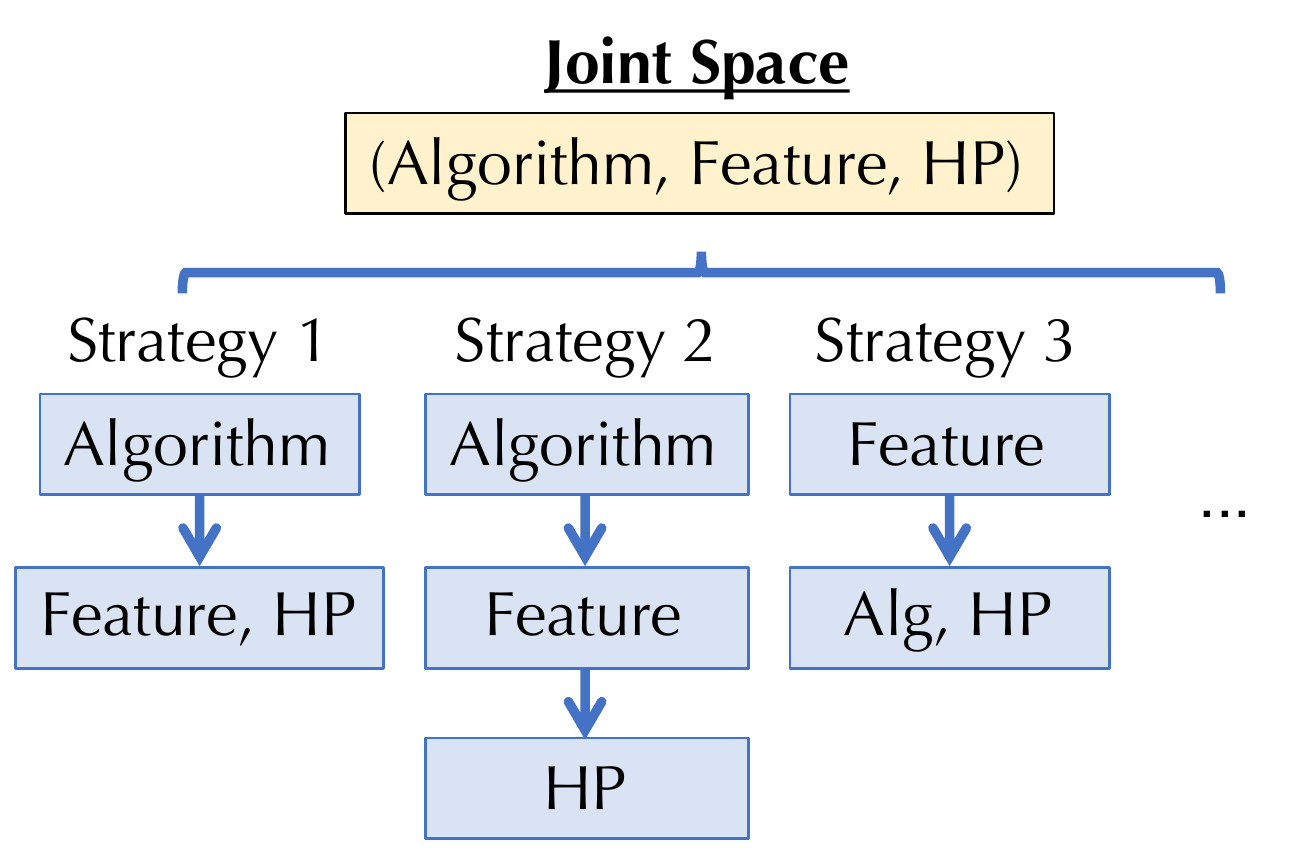}
}
\caption{Different decomposition choices.}
\label{fig:decomposition}
\end{figure}


Many existing systems such as \texttt{auto-sklearn}~\cite{feurer2015efficient} and \texttt{TPOT}~\cite{olson2019tpot} deal with the entire composite search space \textit{jointly}, which naturally leads to the scalability bottleneck.
Decomposing a joint space has been explored for some subspaces (e.g., only algorithm and hyper-parameters as in ~\cite{liu2019automated,li2020efficient}), however, none of them has been applied to a search space as large as that of end-to-end AutoML. 
One challenge is that there exist many different ways to decompose the same space (See Figure~\ref{fig:decomposition}), as shown above, but only \textit{some} of them can perform well. 
\textit{Without a structured, high-level abstraction for search space decomposition to explore different strategies, it is very hard to scale up an end-to-end AutoML system to accommodate the search space that will only get larger in the future.}

\paragraph{\bf Summary of Contributions.}
The initial version of this paper~\cite{li2021volcanoml} appeared in VLDB 2021, where we focused on designing the system, \sys, which is \textit{scalable to a large search space}. In this paper, we make the following four additional contributions:
First, we provide the automatic execution plan generation module (in Section~\ref{sec:plan_generation}) to enrich the proposed framework, and discuss the advantages and underlying problems in this direction.
Second, we propose the meta-learning based components for the building blocks (in Section~\ref{sec:metalearning}) to further speed up \sys.
Third, we conduct a comprehensive set of experiments (in Section~\ref{sec:experiments}) to demonstrate the effectiveness and efficiency of \sys, and provide the results about automatic plan generation and meta-learning based acceleration.
Finally, we provide more details about system components, implementations (interfaces) and search spaces in Section~\ref{sec:appendix} of the appendix.
Our technical contributions are as follows.

\paragraph{C1. System Design: A Structured View on Decomposition.} 

The main technical contribution of \sys is to provide a flexible and principled way of decomposing a large search space into multiple smaller ones.
We propose a novel system abstraction: a set of \textit{\sys building blocks} (Section~\ref{sec:building-blocks}), each of which takes charge of a smaller sub-search space whereas a \textit{\sys execution plan} (Section~\ref{sec:exec-plan}) consists of a \textit{tree} of such building blocks --- the root node corresponds to the original search space and its child nodes correspond to different subspaces. 
Under this abstraction, optimizing in the joint space is conducted as optimization problems over different smaller subspaces. 
The execution model is similar to the classic Volcano query evaluation model in a relational database~\cite{10.5555/1450931} (thus the name \sys): the system asks the root node to take one iteration in the optimization process, which \emph{recursively} invokes one of its child nodes to take one iteration on solving a smaller-scale optimization problem over its own subspace; this recursive invocation procedure will continue until a leaf node is reached.
This flexible abstraction allows us to explore different ways that the same joint space can be decomposed. Together with the meta-learning based optimizations (Section~\ref{sec:metalearning}), \sys can often support  more scalable search process than the existing AutoML systems such as \texttt{auto-sklearn} and \texttt{TPOT}.

\paragraph{C2. Large-scale
Empirical Evaluations.}

We conducted intensive empirical evaluations, comparing \sys with state-of-the-art systems including \texttt{auto-sklearn} and \texttt{TPOT}. 
We show that (1) under the \underline{\textit{same search space}} as \texttt{auto-sklearn},
\sys significantly outperforms \texttt{auto-sklearn} and \texttt{TPOT} --- over 30 classification tasks and 20 regression tasks --- \sys outperforms the \textit{best} of \texttt{auto-sklearn} and \texttt{TPOT} on a majority of tasks; concretely, \sys could achieve a higher balanced accuracy for classification tasks and a smaller mean square error for regression tasks given the same time budget;
(2) using an \underline{\textit{enriched search space}} with additional feature engineering operators, \sys performs significantly better than \texttt{auto-sklearn};
(3) using an \underline{\textit{enriched search space}} with an additional data processing stage and functionalities beyond what \texttt{auto-sklearn} and \texttt{TPOT} currently support (i.e., an additional embedding selection stage using pre-trained models on TensorFlow Hub), \sys can deal with input types such as images efficiently; and (4) \sys is at least comparable with and often outperforms four industrial AutoML platforms on six Kaggle competitions.

\paragraph{\bf Moving Forward.}
The \sys abstraction enables a structured view of optimizing a black-box function via decomposition. This structured view itself opens up interesting future directions. 
For example, one may wish to \textit{automatically} decompose a search space given a workload, just like what a classic query optimizer would do for relational queries.
For constrained optimizations, we also imagine techniques similar to traditional ``\textit{push-down selection}'' could be applied in a similar spirit. 
We explore the possibility of automatically searching for the best plan in Section~\ref{sec:exec-plan} and discuss the limitations of this simple strategy and the exciting line of future work 
that could follow.
While the full treatment of these aspects are beyond the scope of this paper, we hope the \sys abstraction can serve as a foundation for these future endeavors.

\section{Related Work}
\label{sec:related-work}

AutoML is a topic that has been intensively studied over the last decade. 
We briefly summarize related work in this section and readers can consult latest surveys~\cite{automl_book,automl,DBLP:journals/corr/abs-1904-12054,he2020automl} for more details.

\paragraph{\bf End-to-End AutoML.}
End-to-end AutoML, the focus of this work, aims to automate the development process of the end-to-end ML pipeline, including feature preprocessing, feature engineering, algorithm selection, and hyper-parameter tuning.
Often, this is modeled as a black-box optimization problem~\cite{Hutter2015} and solved jointly~\cite{feurer2015efficient,ThoHutHooLey13-AutoWEKA,olson2019tpot}.
Apart from grid search and random search~\cite{bergstra2012random}, genetic programming~\cite{Mohr2018,olson2019tpot} and Bayesian optimization (BO)~\cite{bergstra2011algorithms,hutter2011sequential,snoek2012practical,eggensperger2013towards,bo_survey} has become prevailing frameworks for this problem. One challenge of end-to-end AutoML is the staggeringly huge search space that one has to support and many of these methods suffer from scalability issues~\cite{li2020mfeshb,li2022hyper}.
In addition, meta-learning~\cite{DBLP:journals/corr/abs-1810-03548,li2022transbo,feurer2018scalable} systematically investigates the interactions that different ML approaches perform on a wide range of learning tasks, and then learns from this experience, to accomplish new tasks much faster.
Several meta-learning approaches~\cite{DeSa2017,Hutter2014,VanRijn2018,feurer2015efficient,li2022transfer} can guide ML practitioners to design better search spaces for AutoML tasks.

Many end-to-end AutoML systems have raised the abstraction level of ML.
\texttt{auto-weka}~\cite{ThoHutHooLey13-AutoWEKA}, \texttt{hyperopt-sklearn}~\cite{komer2014hyperopt}, and \texttt{auto-sklearn}~\cite{feurer2015efficient} are the main representatives of BO-based AutoML systems.
\texttt{auto-sklearn} is one of the most popular open-source frameworks.
\texttt{TPOT}~\cite{olson2019tpot} and ML-Plan~\cite{Mohr2018} use genetic algorithms and hierarchical task networks planning, respectively, to optimize over the pipeline space, and require discretization of the hyper-parameter space.
AlphaD3M~\cite{Drori2018} integrates reinforcement learning with Monte Carlo tree search (MCTS) to solve AutoML problems but without imposing efficient decomposition over hyper-parameters and algorithm selection. AutoStacker~\cite{Chen2018} focuses on ensembling and cascading to generate complex pipelines, and solves the CASH (Combined Algorithm Selection and Hyperparameters optimization) problem~\cite{feurer2015efficient} via random search.
ML Bazaar~\cite{smith2020machine} is a general-purpose, multi-task, end-to-end AutoML system, which pair ML pipelines with a hierarchy of AutoML strategies -- Bayesian optimization.
Furthermore, a growing number of commercial enterprises also export their AutoML services to their users, e.g., \texttt{H2O}~\cite{ledell2020h2o}, Microsoft's Azure Machine Learning~\cite{barnes2015azure}, Google Cloud's AI Platform~\cite{GooPre}, Amazon Machine Learning~\cite{liberty2020elastic} and IBM’s Watson Studio AutoAI~\cite{IBMc}.

\paragraph{\bf Automating Individual Components.}
Apart from end-to-end AutoML, many efforts have been devoted to studying sub-problems in AutoML: (1) feature engineering~\cite{Khurana2016,Kaul2017,Katz2017,Nargesian2017,Khurana2018}, (2) algorithm selection~\cite{ThoHutHooLey13-AutoWEKA,komer2014hyperopt,feurer2015efficient,efimova2017fast,liu2019automated,li2020efficient}, and (3) hyper-parameter tuning~\cite{hutter2011sequential,snoek2012practical,bergstra2011algorithms,li2018hyperband,jamieson2016non,falkner2018bohb,li2020mfeshb,swersky2013multi,kleinfbhh17,kandasamy2017multi,poloczek2017multi,hu2019multi,sen2018noisy,wu2019practical,jiang2021automated}. 
Meta-learning methods~\cite{wistuba2016two,golovin2017google,feurer2018scalable} for hyper-parameter tuning can leverage auxiliary knowledge acquired from previous tasks to achieve faster optimization.
Several systems offer a subset of functionalities in the end-to-end process. 
Microsoft's NNI~\cite{nni} helps users to automate feature engineering, hyper-parameter tuning, and model compression.
Recent work~\cite{liu2019automated} leverages the ADMM optimization framework to decompose the CASH problem~\cite{feurer2015efficient}, and solves two easier sub-problems.
Berkeley's Ray~\cite{Moritz2007} and OpenBox~\cite{openbox} provide the \texttt{tune} module~\cite{liaw2018tune,li2020mfeshb} to support scalable hyper-parameter tuning tasks in a distributed environment. 
Featuretools~\cite{kanter2015deep} is a Python library for automatic feature engineering.
Unlike these works, in this paper, we focus on deriving an end-to-end solution to the AutoML problem, where the sub-problems are solved in a joint manner. 

\paragraph{\bf Volcano Model.}
The Volcano model~\cite{Graefe1994} (originally known as the Iterator Model) is the classical evaluation strategy of an analytical DBMS query: Each relational-algebraic operator produces a tuple stream, and a consumer can iterate over its input streams. The tuple stream has three interfaces: \texttt{open}, \texttt{next} and \texttt{close}; all operators own the same interface, and the implementation is opaque.
It is a chain of iterators and data flows through them when the topmost iterator calls \texttt{next()} on the iterator below it. This results in propagation of \texttt{next()} calls till the bottom-most iterator is called. 
\section{\sys and Building Blocks}
\label{sec:building-blocks}

The goal of \sys is to enable scalability with respect to the underlying AutoML search
space. 
As a result, its design focuses on the \emph{decomposition} of a given search space.
In this section, we first introduce key building blocks in \sys, and in Section~\ref{sec:exec-plan} we describe how multiple building blocks are put together to compose a \sys \emph{execution plan} in a modular way.
Later in Section 5, we introduce additional optimizations for these building blocks.

\subsection{Search Space of End-to-End AutoML}
\label{sec:building-blocks:search-space}

We describe the search space of end-to-end AutoML following the presentation in \texttt{auto-sklearn}\cite{feurer2015efficient}.
The input to the system is a dataset $D$, containing a set of training samples. 
The user also provides a pre-defined metric, e.g., validation accuracy or cross-validation accuracy, to measure the \textit{utility} of a given ML pipeline.
The output of an end-to-end AutoML system is an ML pipeline that achieves good utility.

To find such an ML pipeline, the system searches over a large search space of possible
pipelines and picks one that maximizes the pre-defined utility. 
This search space is a composition of (1) feature engineering operators, (2) ML algorithms/models, and (3) hyper-parameters.

\begin{figure}
\begin{center}
\centerline{\includegraphics[width=0.9\columnwidth]{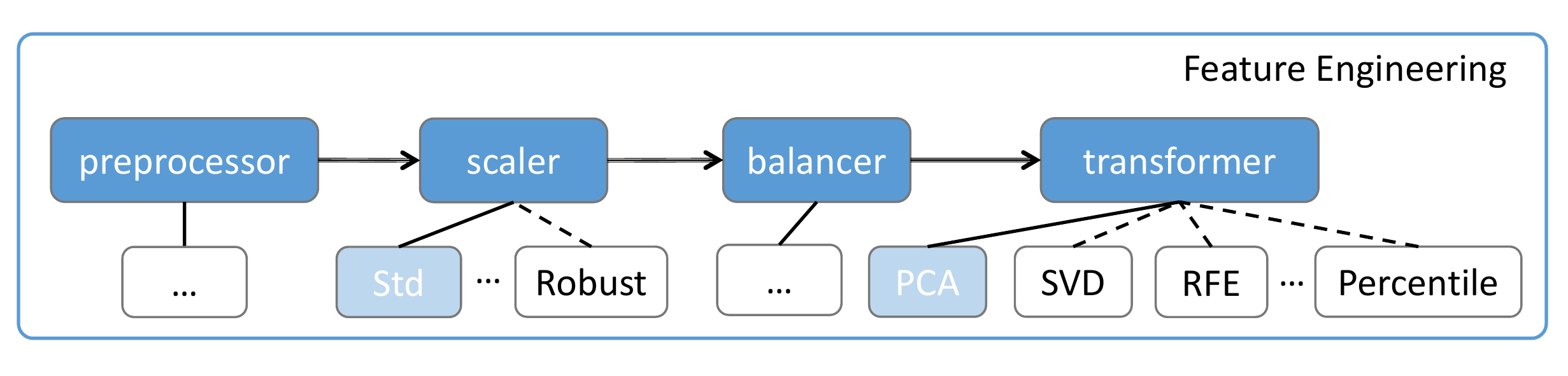}}
\caption{The search space of FE pipeline.}
\label{fe_pipeline}
\end{center}
\end{figure}

\paragraph{\bf \underline{Feature Engineering.}}
The feature engineering process takes as input a dataset $D$ and outputs a new dataset $D'$.
It achieves this by transforming the input dataset via a set of data transformations.
The pipeline for feature engineering is shown in Figure~\ref{fe_pipeline}. It comprises four sequential stages: \emph{preprocessors} (compulsory), \emph{scalers} (5 possible operators), \emph{balancers} (1 possible operators) and \emph{feature transformers} (13 possible operstors).
For each stage, the system chooses a single transformation to apply. 
For example, for \textit{feature\_transforming}, the system can choose among \texttt{no\_processing}, \texttt{kernel\_pca}, \texttt{polynomial}, \texttt{select\_percentile}, etc.


\paragraph{\bf \underline{ML Algorithms}.} 
Given a transformed dataset $D'$, the system then picks an ML algorithm to train. Since different ML algorithms are suitable for different types of tasks, the system needs to consider a diverse range of possible ML algorithms.
Taking \texttt{auto-sklearn} as an example, the search space for ML algorithms contains \texttt{Linear\_Model}, \texttt{Support\_Vector\_Machine}, \texttt{Discriminant\_Analysis}, \texttt{Random\_Forest}, etc.

\paragraph{\bf \underline{Hyper-parameters}.}
Each ML algorithm has its own sub-search space for hyper-parameter tuning --- if we choose to use a certain ML algorithm, we also have to specify the corresponding hyper-parameters. 
The hyper-parameters fall into three categories: continuous (e.g., \texttt{sub-sample\_rate} for \texttt{Random\_Forest}), discrete (e.g., \texttt{maximal\_depth} for \texttt{Decision\_Tree}), and categorical (e.g., \texttt{kernel\_type} for \texttt{Lib\_SVM}).  

\textbf{(AutoML optimization.)}
If the system makes a concrete pick for each of the above decisions, then it can compose a concrete ML pipeline and evaluate its utility.
Concretely, given a pipeline configuration that determines the details of feature engineering, algorithm and hyperparameters, we could construct a specific ML pipeline.
Then we need to train a corresponding ML model within this pipeline, and evaluate its performance on the validation set to obtain the utility of this pipeline configuration. 
This is often an expensive process
since it involves training an ML model. 
To find the optimal ML pipeline, the system evaluates the utility of different ML pipelines in an iterative manner following a \textit{search strategy}, and picks the one that maximizes the utility.
The candidates for search strategy can be random search~\cite{bergstra2012random}, grid search, genetic algorithms~\cite{Mohr2018,olson2019tpot}, Bayesian optimization~\cite{bergstra2011algorithms,hutter2011sequential,snoek2012practical}, bandits based methods~\cite{jamieson2016non,li2018hyperband}, etc.

For example, \texttt{auto-sklearn} handles the above search space \textit{jointly} and optimizes it with Bayesian optimization (BO)~\cite{bo_survey}.
Given an initial set of function evaluations, BO proceeds by fitting a surrogate model to those observations, specifically a \textit{probabilistic Random Forest} in \texttt{auto-sklearn}, and then chooses which ML pipeline to evaluate from the search space by optimizing an acquisition function that balances exploration and exploitation. 

\subsection{Building Blocks}

Unlike \texttt{auto-sklearn}, \sys decomposes the above search space into smaller subspaces.
\textbf{(Key idea.)} Instead of searching over a huge pipeline space, it could be easier for an algorithm to optimize over its subspaces.
Decomposing a joint space has been explored in many domains~\cite{liu2019automated,li2020efficient}.
The way how to decompose the pipeline space into subspaces in field of AutoML is still remains open. Next, we propose a structured and high-level abstraction to support scalable search space decomposition.
One interesting design decision in \sys is to introduce a \emph{structured abstraction} to express different \emph{decomposition strategies}.
A decomposition strategy is akin to an \emph{execution plan} in relational database management systems, which is composed of \emph{building blocks} akin to relational operators.
A building block itself can be viewed as an \emph{atomic} decomposition strategy.
We next present the details of the building blocks implemented by \sys, and we will introduce how to use these blocks to compose \sys execution plans in Section~\ref{sec:exec-plan}.

\paragraph{\underline{Goal}.}
The \textit{goal} of \sys is to solve:
\begin{equation}
\label{eq:def:goal}
\min_{x_1,...,x_n} f(x_1,...,x_n; D),    
\end{equation}
where $x_1,...,x_n$ is a set of $n$ variables and each of them has domain $\mathbb{D}_{x_i}$
for $i \in [n]$.
Together, these $n$ variables define a search space $(x_1,...,x_n) \in \prod_i \mathbb{D}_{x_i}$.
$D$ corresponds to the input dataset, which is a \textit{set} of input samples. 
In AutoML, the variables $x_1,...,x_n$ are actually the pipeline hyperparameters, and the search space is the complete pipeline search space, which is a composition of feature engineering
operators, ML algorithms/models, and hyper-
parameters. 
The optimization objective $f$ is to minimize the validation loss (e.g., classification error), i.e., the objective function $f(\cdot)$ in Formula~\ref{eq:def:goal}.
In our setting, $f(\cdot)$ is a black-box function that we can only evaluate (but not exploiting the derivative), and the objective is to solve $\operatorname{min}f(\cdot)$ as quickly as possible.
Given a fixed $\bm{c}$ (i.e., a concrete ML pipeline) in the composite domain $\bm{c} \in \prod_i \mathbb{D}_{x_i}$, we use the notation
$f(\bm{c}; D)$
as the value of evaluating $f$ by substituting $(x_1,...x_n)$ with $\bm{c}$.

\paragraph{\underline{Subgoal}.} 
One key decision of \sys is to solve the optimization problem on a search space by decomposing it into multiple smaller subspaces, each of which will be solved by one \textit{building block}. 
We define optimizing over each of these smaller subspaces as a \textit{subgoal} of the original problem.
Formally, a subgoal $g$ is defined by two components: $\bm{\bar{x}}_g \subseteq \{x_1,...x_n\}$ as a subset of variables, and 
$\bm{\bar{c}}_g \in \prod_{x_i \in \bm{\bar{x}}_g} \mathbb{D}_{x_i}$ as an assignment in the domain of all variables in $\bm{\bar{x}}_g$. 
Let $\bm{\bar{x}}_{-g} = \{x_1,...,x_n\} - \bm{\bar{x}}_g$ be all variables that are \textit{not} in $\bm{\bar{x}}_g$.

Each subgoal defines a function $f_g$ over a smaller search space, which is constructed by \textit{substituting} all variables in $\bm{\bar{x}}_g$ with $\bm{\bar{c}}_g$:
\begin{equation}
\begin{aligned}
    f_g = & f[\bm{\bar{x}}_g \slash \bm{\bar{c}}_g]: \\
    & {\bm{z}} \in \prod_{x_i \in \bm{\bar{x}}_{-g}} \mathbb{D}_{x_i}  \mapsto f(\{\bm{\bar{c}}_g; \bm{z}\}; D).\\
\end{aligned}
\end{equation}
Each subgoal is a sub-problem in the ML pipeline search of AutoML such as feature engineering, algorithm selection, etc.

\paragraph{\underline{Building Block}.}
Each subgoal $g$ corresponds to one building block $B_{g, D}$, whose goal is
to solve
\begin{equation}
\min_{\bm{\bar{x}}_{-g}} f_g(\bm{\bar{x}}_{-g}; D).   
\end{equation}
A building block $B_{g, D}$ imposes several assumptions on $g$ and $D$.
First, given an assignment $\bm{\bar{c}}_{-g}$ to $\bm{\bar{x}}_{-g}$, it is able to evaluate the value of the function $f_g(\bm{\bar{c}}_{-g},D)$.
Second, given a dataset $D$, a building block has the knowledge about how to subsample a smaller dataset $\tilde{D} \subseteq D$ and then conduct evaluations on such a subset $\bm{x} \mapsto f_g(\bm{x}; \tilde{D})$. 
Third, we assume that the building block has access to a cost model about the cost of an evaluation at $\bm{x}$, $C_{g, D, \bm{x}}$.

\paragraph{\underline{Interfaces}.} 
All implementations of a building block follow an interactive optimization process. A building block exposes several interfaces. 
First, one can initialize a building block via
\begin{equation}
B_{g, D} \leftarrow \texttt{init}(f, \bm{\bar{x}}_g, \bm{\bar{c}}_g, D),
\end{equation}
which creates a building block (i.e., a new sub-problem).
Second, one can query the current best solution found in $B_{g, D}$ by
\begin{equation}
\bm{\hat{x}} \leftarrow \texttt{get\_current\_best}(B_{g, D}).    
\end{equation}
Furthermore, one can ask $B_{g, D}$ to iterate once via
\begin{equation}
\texttt{do\_next!}(B_{g, D}),    
\end{equation}
where `\texttt{!}' indicates potential change on the state of the input $B_{g, D}$.

Last but not least, one can query a building block about its \emph{expected utility} (EU) if given $K$ more budget units (e.g., seconds) via
\begin{equation}
[l, u] \leftarrow \texttt{get\_eu}(B_{g, D}, K).    
\end{equation}
By adopting a similar design principle used in the existing AutoML systems~\cite{feurer2015efficient,olson2019tpot,liu2019automated}, in \sys we estimate EU by \emph{extrapolation} into the future with more available budget.
Given the inherent uncertainty in our estimation method, rather than returning a single point estimate, we instead return a lower bound $l$ and an upper bound $u$.
We refer readers to~\cite{li2020efficient} for the details of how the lower and upper bounds are established.
Moreover, one can query a building block about its \textit{expected utility improvement} (EUI) via
\begin{equation}
\delta \leftarrow \texttt{get\_eui}(B_{g, D}).    
\end{equation}
Note that, different from EU, EUI is the expected \textit{improvement} over the current observed utility if given $K$ more budget units.
While sharing some similarity with EI in BO, EUI works on the level of optimization process (building blocks), while EI in BO is implemented for one single iteration in BO.
In \sys, we estimate EUI by taking the mean of the observed improvements from history, following Levine et al~\cite{levine2017rotting}.

\subsection{\bf Three Types of Building Blocks}

Decomposition is the cornerstone of \sys's design.
Given a search space, apart from exploring it jointly, there are two classical ways of decomposition --- to partition the search space via conditioning on different values of a certain variable (in a similar spirit of \emph{variable elimination}~\cite{dechter1998bucket}), or to decompose the problem into multiple smaller ones by introducing equality constraints (in a similar spirit of \emph{dual decomposition}~\cite{caroe1999dual}). This inspires \sys's design, which supports three types of building blocks: 
(1) \emph{joint block} that simply optimizes the input subspace using Bayesian optimization;
(2) \emph{conditioning block} that further divides the input subspace into smaller ones by conditioning on one particular input variable;
and (3) \emph{alternating block} that partitions the input subspace into two and optimizes each one \emph{alternately}.
Note that both \emph{conditioning block} and \emph{alternating block} would generate new building blocks with smaller subgoals.
We next present the implementation details for each type of building block.

\subsubsection{Joint Block}



A joint block directly optimizes its subgoal via Bayesian optimization (BO)~\cite{bo_survey}.
Specifically, BO based method - SMAC~\cite{hutter2011sequential} has been used by many applications where evaluating the objective function is computationally expensive.
It constructs a probabilistic surrogate model $M$ to capture the relationship between the input variables $\bm{\bar{x}}$ (i.e., hyperparamters in AutoML) and the objective function value $\psi$ (e.g., the validation loss), and this surrogate model is utilized to suggest a new promising configuration to evaluate for each iteration.
It then refines $M$ iteratively using past evaluation observations $(\bm{\bar{x}}, \psi)$.

Based on the BO framework, the implementation of \texttt{do\_next!} for a joint block consists of the following three steps:
\begin{enumerate}
    \item Use the surrogate model $M$ to select a configuration $\bm{\bar{x}}$ that maximizes an acquisition function. In our implementation, we use \emph{expected improvement} (EI)~\cite{jones1998efficient} as the acquisition function, which has been widely used in BO community.
    \item Evaluate the selected configuration $\bm{\bar{x}}$ and obtain its result about the objective function $f_g(\bm{\bar{x}})$ (i.e., the subgoal). 
    Due to the randomness of most ML algorithms, we assume that $f(\bm{x})$ cannot be observed directly but rather through noisy observation $\psi=f_g(\bm{\bar{x}})+\epsilon$, with $\epsilon\sim\mathcal{N}(0, \sigma^2)$, where $\mathcal{N}$ is the normal distribution.
    \item Refit the surrogate model $M$ on the observed $(\bm{\bar{x}}, \psi)$.
\end{enumerate}

\textbf{Early-Stopping based Optimization.}
For large datasets, early-stopping based methods, e.g., Successive Halving~\cite{jamieson2016non}, Hyperband~\cite{li2018hyperband}, BOHB~\cite{falkner2018bohb}, MFES-HB~\cite{li2020mfeshb}, etc, can terminate the evaluations of poorly-performing configurations in advance, thus speeding up the evaluations. 
\sys supports MFES-HB~\cite{li2020mfeshb}, which combines the benefits of Hyperband and Multi-fidelity BO~\cite{wu2019practical,takeno2020multifidelity}, to optimize a joint block, in addition to vanilla BO.

\begin{algorithm}[t]
  \small
  \SetAlgoLined
  \KwIn{A conditioning block $B_{g, D}$, times to play each arm $L$, total budget $K$.}
  \SetAlgoLined
  \caption{The \texttt{do\_next!} of conditioning block}
  \label{alg:cond:next}
  Let $B_1$, ..., $B_m$ be all active (have not been eliminated) child blocks\;
  \For{$1\leq i\leq L$}
  {
    \For{$1\leq j\leq m$}
    {
        \texttt{do\_next!}$(B_j)$\;
    }
  }
  \For{$1\leq j\leq m$}
  {
    $[l_j, u_j]\leftarrow$~\texttt{get\_eu}$(B_j, K)$\;
  }
  Eliminate child blocks that are \emph{dominated} by others, using $[l_j, u_j]$ for $1\leq j\leq m$\;
\end{algorithm}

\subsubsection{Conditioning Block}
A conditioning block decomposes its input $\bm{\bar{x}}$ into $\bm{\bar{x}} = \{x_c\} \cup \bm{\bar{y}}$, where $x_c$ is a single variable with domain $\mathbb{D}_{x_c}$.
It then creates one new building block for each possible value $d \in \mathbb{D}_{x_c}$ of $x_c$:
\begin{equation}
\min_{\bm{\bar{y}}} g_d(\bm{\bar{y}}; D) \equiv f(\{x_c=d, \bm{\bar{y}}\}; D).    
\end{equation}
As a result, $|\mathbb{D}_{x_c}|$ new (child) building blocks are created.

The conditioning block aims to identify optimal value for $x_c$, and many previous AutoML researchers have used Bandit algorithms for this purpose~\cite{liu2019automated,jamieson2016non,li2020efficient,li2020mfeshb}. 
In \sys, we follow these previous work and 
model it as a multi-armed bandit (MAB) problem, while our framework is flexible enough to incorporate other algorithms when they are available.
There are $|\mathbb{D}_{x_c}|$ arms, where each arm corresponds to a child block.
Playing an arm means invoking the \texttt{do\_next!} primitive of the corresponding child block.

Algorithm~\ref{alg:cond:next} illustrates the implementation of \texttt{do\_next!} for a conditioning block.
It starts by playing each arm $L$ times in a Round-Robin fashion (lines 2 to 4).
Here, $L$ is a user-specified configuration parameter of \sys.
In our current implementation, we set $L=5$.
We then obtain the lower and upper bounds of the expected utility of each child block by invoking its \texttt{get\_eu} primitive (lines 5 to 6), and eliminate child blocks that are dominated by others (line 7).
The elimination works as follows.
Consider two blocks $B_i$ and $B_j$: if the upper bound $u_i$ of $B_i$ is less than the lower bound $l_j$ of $B_j$, then the block $B_i$ is eliminated.
An eliminated arm/block will not be played in future invocations of \texttt{do\_next!}.

\begin{algorithm}[t!]
  \small
  \SetAlgoLined
  \KwIn{An alternating block $B_{g, D}$ with search space $\bm{\bar{x}}=\bm{\bar{y}} \cup \bm{\bar{z}}$.}
  \SetAlgoLined
  \caption{The \texttt{init} of alternating block}
  \label{alg:alternating:init}
  Initialize $\bm{\bar{y}}$ and $\bm{\bar{z}}$ with default values $\bm{\bar{y}}_0$ and $\bm{\bar{z}}_0$\;
  $B_1\leftarrow$~\texttt{init}$(f, \bm{\bar{z}}, \bm{\bar{z}}_0, D)$\;
  $B_2\leftarrow$~\texttt{init}$(f, \bm{\bar{y}}, \bm{\bar{y}}_0, D)$\;
  \For{$1\leq i\leq L$}
  {
    \texttt{do\_next}$(B_1)$\;
    $\bm{\bar{y}}_i\leftarrow$~\texttt{get\_current\_best}$(B_1)$\;
    \texttt{set\_var}$(B_2, \bm{\bar{y}}, \bm{\bar{y}}_i)$\;
    \texttt{do\_next}$(B_2)$\;
    $\bm{\bar{z}}_i\leftarrow$~\texttt{get\_current\_best}$(B_2)$\;
    \texttt{set\_var}$(B_1, \bm{\bar{z}}, \bm{\bar{z}}_i$)\;
  }
\end{algorithm}

\noindent
\textbf{Remark:} We have simplified the above elimination criterion by using the lower and upper bounds calculated given $K$ budget units for \emph{each arm}.
In fact, these $K$ budget units are \emph{shared} by all the arms, and as a result, each arm actually has fewer budget units than $K$.
Our assumption is that, $K$ is sufficiently large so that one can play \emph{all arms} until (the observed distribution of rewards of) \emph{each arm} converges.
Otherwise, the lower and upper bounds obtained may be \emph{over-optimistic}, and as a result, may lead to incorrect eliminations.
Fortunately, our assumption usually holds in practice, where arms converge relatively fast.

\subsubsection{Alternating Block}

An alternating block decomposes its input search space into $\bm{\bar{x}} = \bm{\bar{y}} \cup \bm{\bar{z}}$, and explores $\bm{\bar{y}}$ and $\bm{\bar{z}}$ in an \emph{alternating} way.
Similarly, we also model the optimization in alternating block as an MAB problem.
Algorithm~\ref{alg:alternating:init} illustrates how its \texttt{init} primitive works.
It first creates two child blocks $B_1$ and $B_2$, which will focus on optimizing for $\bm{\bar{y}}$ and $\bm{\bar{z}}$ respectively (lines 1 to 3).
It then (again) views $B_1$ and $B_2$ as two arms and plays them using Round-Robin (lines 4 to 10).
Note that, when $B_1$ optimizes $\bm{\bar{y}}$ (resp. when $B_2$ optimizes $\bm{\bar{z}}$), it uses the current best $\bm{\bar{z}}$ found by $B_2$ (resp. the current best $\bm{\bar{y}}$ found by $B_1$).
This is done by the \texttt{set\_var} primitive (invoked at line 7 for $B_2$ and line 10 for $B_1$).

One problem of our alternating MAB formulation is that the utility improvements of the two building blocks often vary dramatically in practice.
For example, some applications are very sensitive to the features being used (e.g., normalized vs. non-normalized features) while hyper-parameter tuning will offer little or even no improvement.
In this case, we should spend more resources on looking for good features instead of tuning hyper-parameters.
Our key observation is that, the \emph{expected utility improvement} (EUI) decays as optimization proceeds.
As a result, we propose to use EUI as an indicator that measures the \emph{potential} of pulling an arm further.
Algorithm~\ref{alg:alternating:next} illustrates the details of this idea when used to implement the \texttt{do\_next!} primitive.

\begin{algorithm}[t]
  \small
  \SetAlgoLined
  \KwIn{An alternating block $B_{g, D}$ with budget $K$.}
  \SetAlgoLined
  \caption{The \texttt{do\_next!} of alternating block}
  \label{alg:alternating:next}
  $\delta_1\leftarrow$~\texttt{get\_eui}$(B_1)$\;
  $\delta_2\leftarrow$~\texttt{get\_eui}$(B_2)$\;
  \If{$\delta_1\geq \delta_2$}
  {
    $\bm{\bar{z}}_{\text{best}}\leftarrow$~\texttt{get\_current\_best}$(B_2)$\;
    \texttt{set\_var}$(B_1, \bm{\bar{z}}, \bm{\bar{z}}_{\text{best}})$\;
    \texttt{do\_next}$(B_1)$\;
  }\Else{
    $\bm{\bar{y}}_{\text{best}}\leftarrow$~\texttt{get\_current\_best}$(B_1)$\;
    \texttt{set\_var}$(B_2, \bm{\bar{y}}, \bm{\bar{y}}_{\text{best}})$\;
    \texttt{do\_next}$(B_2)$\;
  }
\end{algorithm}

Specifically, Algorithm~\ref{alg:alternating:next} starts by polling the EUI of both child blocks (lines 1 and 2).
Recall that the EUI is estimated by taking the mean of historic observations.
It then compares the EUIs and picks the arm/block with larger EUI to play next (lines 3 to 10).
Before pulling the winner arm, again it will use the current best settings found by the other arm/block (lines 4 to 6, lines 8 to 10). 

\subsubsection{Discussion: Pros and Cons of Building Blocks} 
While the joint block is the most straightforward way to solve the optimization problem associated, it is difficult to scale Bayesian optimization to a large search space~\cite{wang2013bayesian,li2020efficient}.
The alternating block addresses this scalability issue by decomposing the search space into two smaller subspaces, though with the assumption that the improvements of the two subspaces are \emph{conditionally independent} of each other.
As a result, the alternating block is a better choice when such an assumption approximately holds.
The advantage of alternating block with this assumption can solve the optimization problem efficiently by decomposing the huge and joint search space into two smaller subspaces (efficiency).
While the issue behind this lies in that the alternating block cannot converge to the optimal solution (effectiveness) when the two subspaces are highly dependent.
We expect this assumption approximately hold; if not, the alternating block still has its position when dealing with the ``efficiency vs. effectiveness'' trade-off when the search space is large. 
The conditioning block is capable of pruning the search space \emph{as optimization proceeds}, when bad arms are pulled less often or will not be played anymore, with the limitation that it can only work for conditional variables that are \emph{categorical}. 
For non-categorical variables, one possible way to use conditioning blocks is to split the value range of variables. 
For example, given a numerical variable that ranges from 1 to 3, we split it into two ranges, which are [1, 2) and [2, 3). 
During the optimization iteration, we first choose one sub-range and then optimize the splitted space along with its corresponding subspace.

In addition, \sys uses bandit-based algorithms from the existing literature~\cite{levine2017rotting,li2020efficient} as default in both the alternating and conditioning block, and other bandit-based algorithms, such as successive halving~\cite{jamieson2016non}, Hyperband~\cite{li2018hyperband},  BOHB~\cite{falkner2018bohb} and MFES-HB~\cite{li2020mfeshb}, can also be used in these blocks.

\subsubsection{Discussion: Comparing Different Building Blocks}
Joint blocks are the default blocks that can be applied to all problems. 
When the search space is rather large, conditioning and alternating blocks can be helpful.
If the search space contains a categorical hyper-parameter, under which the subspace of each choice is conditionally independent with each other, 
the conditioning block can be used instead of exploring the entire space. 
If the search space can be decomposed into two approximately independent subspaces, the alternating block can be applied to this case.
As a result, a scalable system needs to be able to decompose the problem in different ways and pick the most suitable building blocks. 
This forms a \sys execution plan, which we will describe in the next section.
In Section~\ref{sec:exec-plan}, we explore the possibility of automatically choosing building blocks to use by maximizing the empirical accuracy of different execution plans, given a pre-defined set of datasets.

\subsubsection{Discussion: Continue Tuning in Conditional Block}
As introduced in Section 3.3.2, in the conditional block of \sys, we store the lower and upper bounds of the expected utility of each child block. 
\sys eliminates those potentially bad blocks based on the two bounds.
When new algorithms are added into the search space, we extend the previously survived candidate algorithm set in the conditional block with those new algorithms and play each candidate in a round-robin fashion as described in Section 3.3.2. 
After \sys evaluates those new algorithms with sufficient budget, the conditional block follows the bandit algorithm and eliminates bad candidates with low upper bounds from the candidate set.
This process still follows Algorithm 1, only with a difference that the child blocks are extended with new algorithms.
Therefore, it's quite natural and easy to support continue tuning in \sys.

\section{\sys Execution Plan}
\label{sec:exec-plan}

Given a pre-defined search space, the input of \sys is (1) a dataset $D$, (2) a utility metric (e.g., cross-validation accuracy) which defines the objective function $f$, and (3) a time budget.
\sys then decomposes a large search space into an \emph{execution plan}, following some specific \emph{decomposition strategy}.

\begin{figure}
\centering
\includegraphics[width=0.35\textwidth]{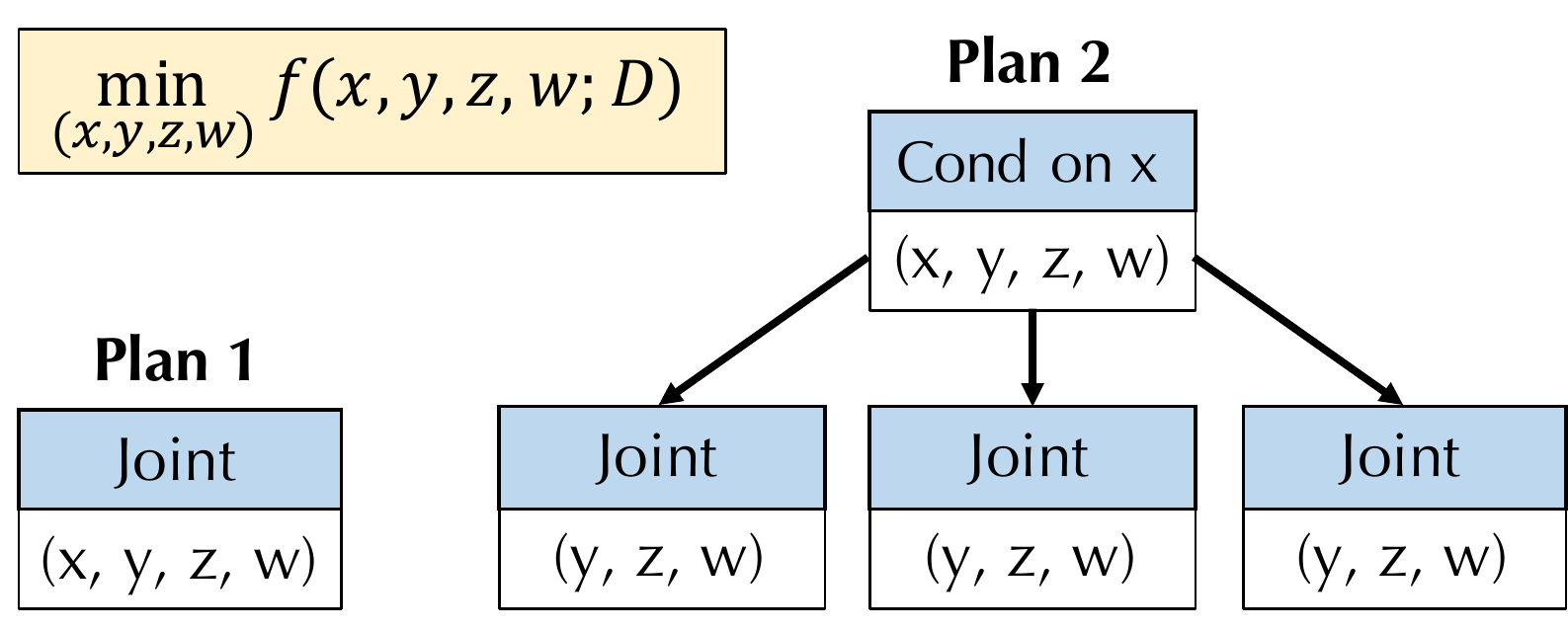}
\caption{Two different execution plans
for the same optimization problem. Each plan
corresponds to a different way to decompose 
the same search space $(x,y,z,w)$.}
\label{fig:plan}
\end{figure}

\subsection{Execution Plan}
\paragraph{\underline{\sys Execution Plan}.}
Due to the space limitation, we omit the formal definition of a \sys execution plan. 
A \sys execution plan is a \textit{tree} of building blocks.
The root node corresponds to a building block solving the problem $f$ with the entire search space, which can be further decomposed into multiple sub-problems.
For each generated sub-problem, a building block (from the three candidates) is applied to solve the corresponding problem.
In addition, all the leaf nodes must be the joint blocks.
Since joint block does not decompose the search space, it can not be in any paths from the root node to leaf node.
As an example, Figure~\ref{fig:plan} illustrates two possible execution plans
for $f(x,y,z,w;D)$.
\textbf{Plan 1} contains only a single root building block as a joint block, whereas \textbf{Plan 2} first introduces a conditioning block on $x$, and then creates one lower level of building blocks for each possible value of $x$ (in Figure~\ref{fig:plan}, we assume that $|\mathbb{D}_x|=3$).

\paragraph{\underline{\sys Execution Model}.}
To execute a \sys execution plan, we follow a Volcano-style execution that is similar to a relational database~\cite{Graefe1994} --- the system invokes the $\texttt{do\_next!}$ of the root node, which then invokes the $\texttt{do\_next!}$ of one of its child nodes, propagating until the leaf node.
At any time, one can invoke the \texttt{get\_current\_best} of the root node, which returns the current best solution for the entire search space.

\begin{figure}[t!]
\centering
\includegraphics[width=0.45\textwidth]{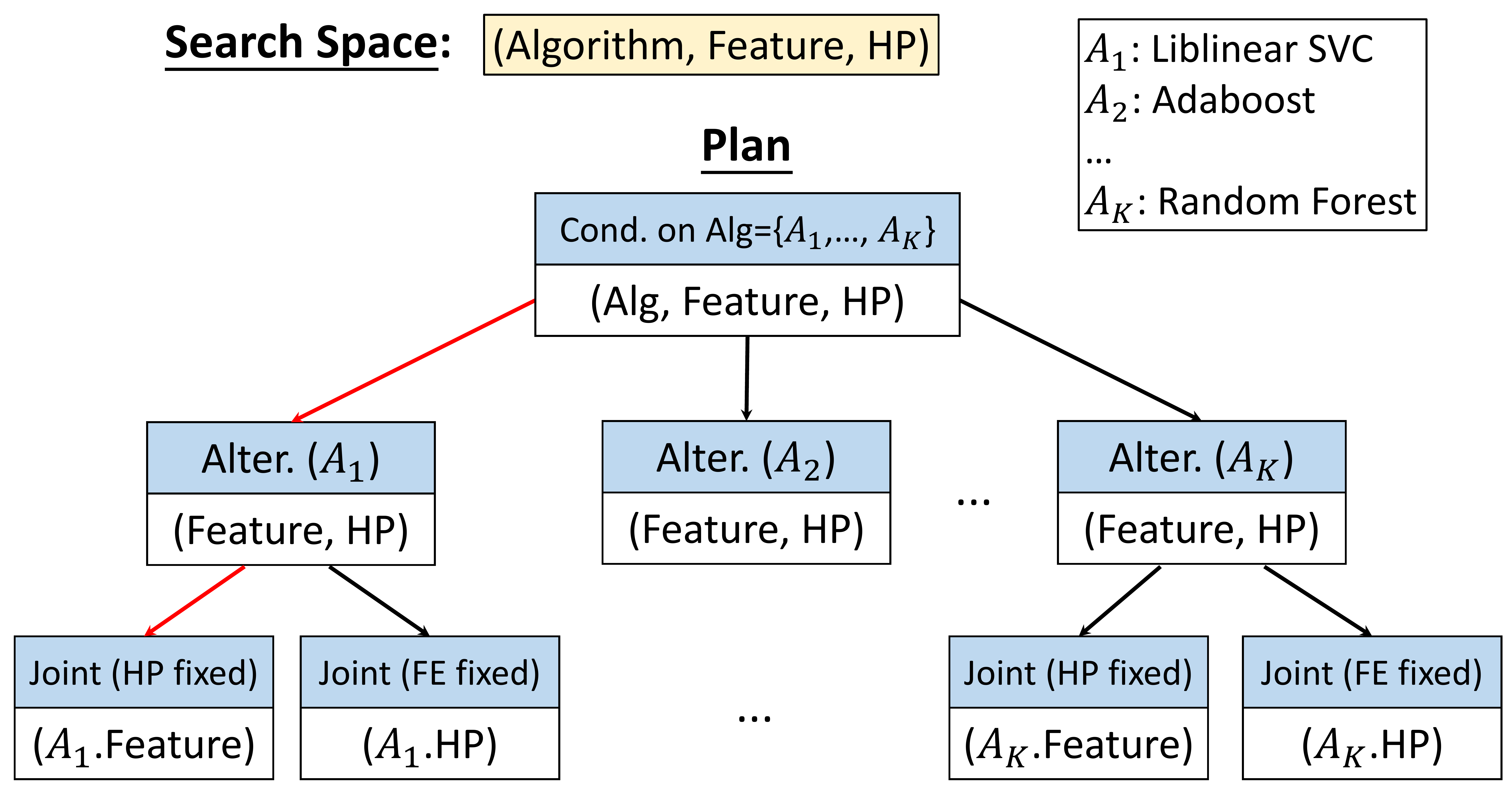}
\caption{\sys's execution plan for the same search space as explored by \texttt{auto-sklearn}. Here `Alg' and `HP' correspond to Algorithm and hyper-parameters respectively.}
\label{fig:plan_revised}
\end{figure}

\paragraph{\underline{\sys Plan for \texttt{auto-sklearn}}.}
Figure~\ref{fig:plan_revised} presents a \sys execution plan for the same search space explored by \texttt{auto-sklearn}, which consists of the joint search of \textit{algorithms}, \textit{features transformations operators}, and \textit{hyper-parameters}. 
Instead of conducting the search process in a single joint block, as was done by \texttt{auto-sklearn}, \sys first decomposes the search space via a conditioning block 
on \textit{algorithms} --- this introduces a MAB problem in which each arm corresponds to one particular algorithm.
It then decomposes each of the conditioned subspaces via an alternating block between feature engineering and hyper-parameter tuning. 
The whole subspace of feature engineering (resp. that of hyper-parameter tuning) is optimized by a joint block. Note that this execution plan is similar to the regular plan of human experts, in which experts usually try different algorithms and optimize the feature engineering operations and hyperparameters alternatively for specific well-performing algorithms.

Concretely, Figure \ref{fig:plan_revised} shows a search space for AutoML with $K$ choices of ML algorithms.
During each iteration, starting from the root node, \sys selects the child node to optimize until it reaches a leaf node and then optimizes over the subspace in the leaf node. 
As shown by the red lines in Figure \ref{fig:plan_revised}, in this iteration, \sys only tunes the feature engineering pipeline of algorithm $A_1$ while fixing its algorithm hyper-parameters.

\paragraph{\underline{Alternative Execution Plans}.}
Note that the execution plan in Figure~\ref{fig:plan_revised} is not the only possible one.
Our flexible and scalable framework in \sys allows us to explore different execution plans before reaching the proposed one, and in the next section~\ref{sec:plan_generation} we introduce the way of automatic plan generation.
The reason why we choose this plan is due to the fundamental property of the AutoML search space --- we observe that, the optimal choices of \textit{features} are different across \textit{algorithms}, which implies that we can first decompose the search space along \textit{ML algorithms}.
The improvements introduced by feature engineering and hyper-parameter tuning are largely complementary (See~\ref{appendix:obs} for more details), and thus we can optimize them \emph{alternately}.
For feature engineering (resp. hyper-parameter tuning), the subspace is small enough to be handled by a single joint block efficiently. In Section~\ref{sec:plan_generation}, we will list the possible plans of the coarse-grained level and further discuss the opportunity of automatic plan generation.

\begin{figure}[t!]
\centering
\includegraphics[width=0.42\textwidth]{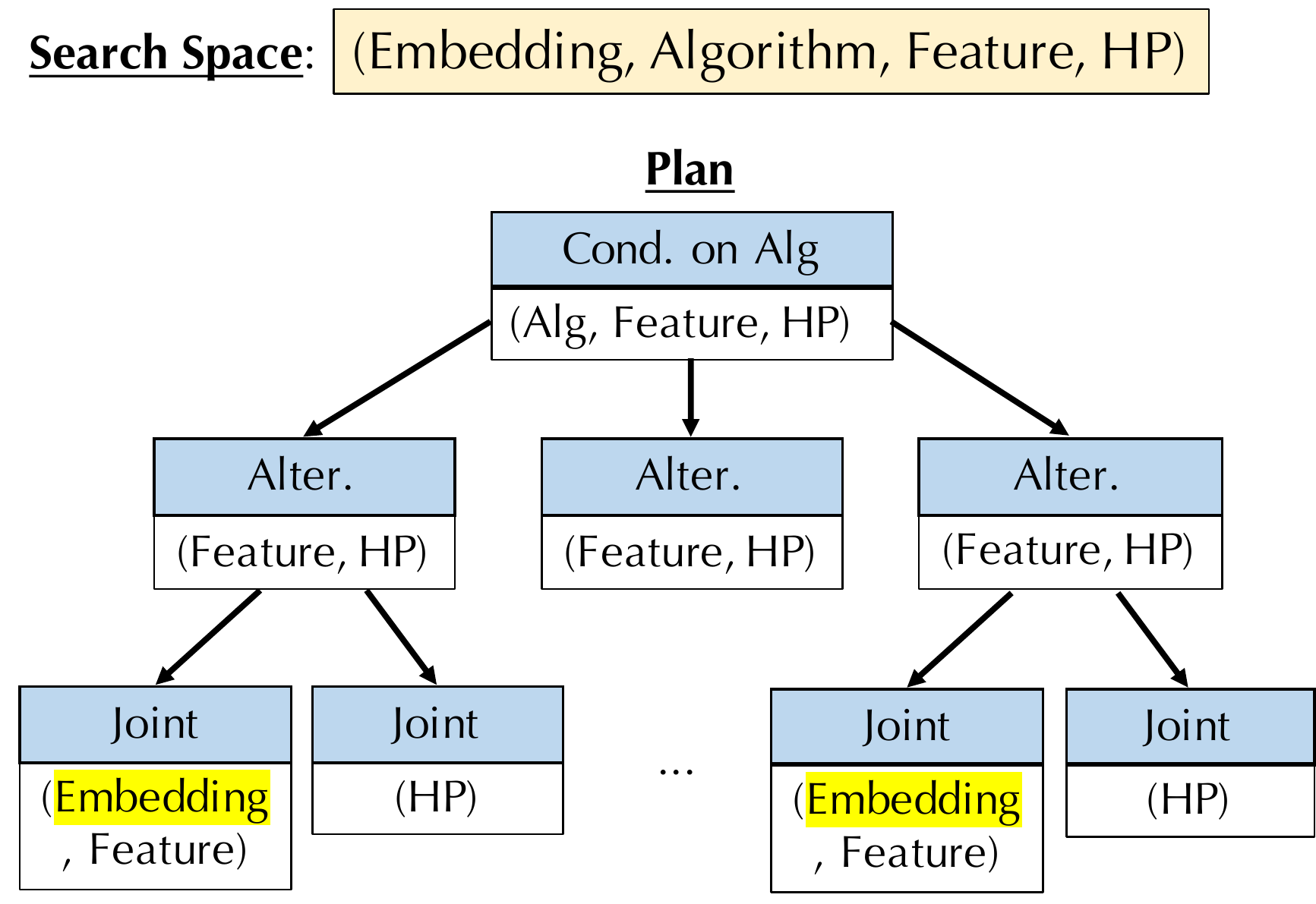}
\caption{\sys's execution plan for a larger search space enriched by an additional 
embedding selection stage.}
\label{fig:plan_embedding}
\end{figure}

\paragraph{\underline{\sys Plan for Enriched Search Space}.}
We can easily extend \sys and enable functionalities that are not supported by most AutoML systems.
For example, Figure~\ref{fig:plan_embedding} illustrates an execution plan for a search space with an additional stage --- \emph{embedding selection}. Given an input, e.g., image or text, we first choose embeddings based on a collection of TensorFlow Hub pre-trained models and then conduct algorithm selection, feature engineering, and hyper-parameter tuning. 
We use an execution plan as illustrated in Figure~\ref{fig:plan_embedding}, having the embedding selection step jointly optimized together with the feature engineering. 

\subsection{Automatic Plan Generation}
\label{sec:plan_generation}

\begin{figure*}[t!]
	\centering
		\scalebox{0.9}{	\includegraphics[width=0.8\linewidth]{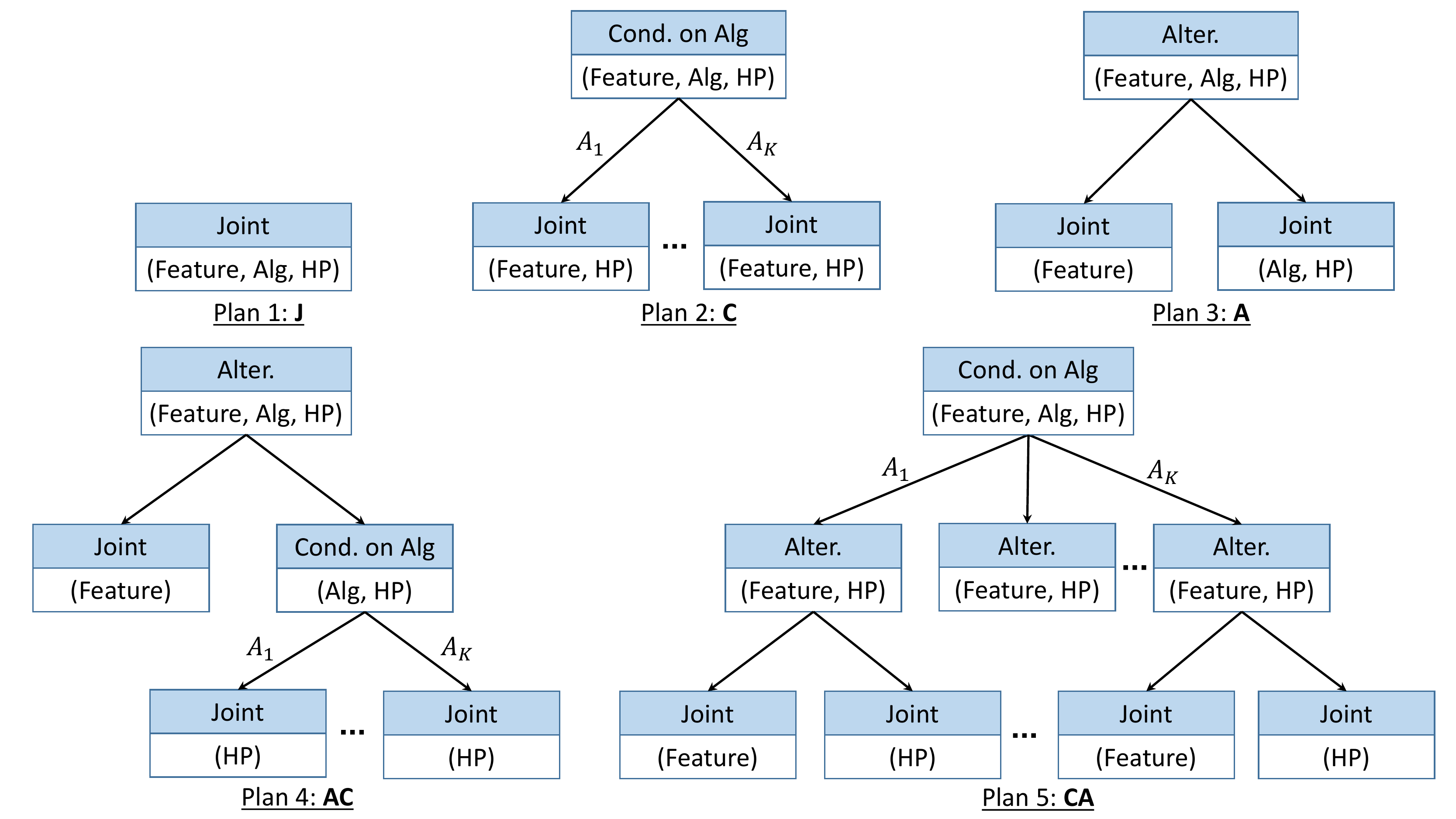}
	}
	\caption{Five execution plans in the task granularity.}
  \label{fig:execution_plans}
\end{figure*}

In principle, the design of \sys opens up the opportunity for automatic plan generation --- given a collection of benchmark datasets, one could automatically search for the best decomposition strategy of the search space and come up with a physical plan automatically.
While the complete treatment of this problem is beyond the scope of this paper, we illustrate the possibility with a straightforward strategy.
We automatically enumerate all possible execution plans in a coarse-grained level and find that our manually specified execution plan in Figure~\ref{fig:plan_revised} outperforms the alternatives.
The five execution plans are as follows:

\begin{itemize}
    \item Plan 1 - J(Joint). Optimize over the entire space using a joint block. 
    \item Plan 2 - C(Conditioning). Use a conditioning block on the choice of machine learning algorithms, and then optimize each subspace using joint blocks.
    \item Plan 3 - A(Alternating). Use an alternating block to separate the entire space into feature engineering space and combined algorithm selection and hyper-parameter tuning (CASH) space.
    \item Plan 4 - AC(Alternating then Conditioning). Use an alternating block to separate the entire space into feature engineering and CASH space, and then use a conditioning block on the choice of algorithm. 
    \item Plan 5 - CA(Conditioning then Alternating). Use a conditioning block on the choice of machine learning algorithms, and then optimize the subspace of feature engineering and algorithm hyperparameters alternately. See Plan 5 in Figure~\ref{fig:execution_plans} for more details.
    \item TPOT - \texttt{TPOT}. In essence, the execution plan of \texttt{TPOT} also uses a single joint block. The difference between \texttt{TPOT} and Plan 1 is that \texttt{TPOT} uses the evolutionary algorithm while Plan 1 uses the Bayesian optimization.
    \item AUSK - \texttt{autosklearn}. The execution plan of \texttt{autosklearn} also uses a single joint block. The difference between \texttt{autosklearn} and Plan 1 is their ensemble strategy. Concretely, \texttt{autosklearn} build the ensemble model over all the evaluated models while Plan 1 builds it over a fixed number of well-performed models as \sys does.
\end{itemize}

Indeed, automatic plan generation can find the optimal solutions with techniques like reinforcement learning.
However, one critical problem behind the automatic plan generation is the overhead introduced by constructing and searching for a new execution plan.
Here, generating execution plans may take a massive amount of training cost, and automatic plan generation may involve building and evaluating an extremely large volume of plans.
Moreover, if the user only has a limited budget, 
automated plan generation can easily run out of the budget while not providing a decent execution plan.

It is still an open question of whether we can support finer-grained partition of the search space (e.g., different plans for different subspace of features), and moreover, whether we can conduct efficient automatic plan optimization without enumerating all possible plans. 
These are exciting future directions, and we expect 
the endeavor to be non-trivial. 
We hope that this paper sets the ground for this line of research
in the future (e.g., rule-based heuristics or reinforcement learning). 

\para{Further Discussion.}
We abstract a VolcanoML execution plan as a tree of building blocks. 
The root node corresponds to a building block solving the problem with the entire search space, which can be further decomposed into multiple building blocks if necessary.
Three kinds of building blocks can be used to build the tree-structured execution plan. 
Reinforcement Learning (RL) could be a straightforward solution to generate execution plans automatically.
The key decisions involve how to define the states, rewards, and actions. 
We can define the current state by encoding the current structure of the tree and the optimization problem to decompose. 
When all leaf nodes in the tree are joint blocks, we can execute the current decomposition plan. 
And we can take the validation accuracy as the reward. 
Each action corresponds to apply a decomposition strategy to an optimization problem by adding a building block to the tree's some leaf node. 
The RL agent builds and evaluates each execution plan iteratively by trying different actions. 
The goal of the agent is to find the plan that achieves the optimal evaluation result.


\subsection{Progressive Optimization Methods}
Unlike the optimization strategy used in Figure \ref{fig:plan_revised}, the progressive methods~\cite{liu2018progressive} can optimize the search space in a top-down manner. 
Take the default tree-structured space (Plan 5) shown in Figure~\ref{fig:execution_plans} as an example, a progressive method first tries different algorithms in the conditional block while keeping all other hyperparameters by default.
After evaluating all algorithm candidates, it fixes the best algorithm and enters the search space under this algorithm.
Then, it optimizes the space of feature engineering while keeping the algorithm hyperparameters by default.
Finally, by fixing the best found feature engineering operators, it optimizes the algorithm hyperparameters and obtains the final configuration.
The main advantage of progressive methods is that, they enjoy high efficiency in exploring the space because they only need to optimize the blocks following a path from the root to the leaves.
However, they also have two weak points: 
(1) While the best algorithm is chosen by keeping other hyperparameters by default, there is a risk that the algorithm found progressively may not be the optimal one; 
(2) Only one algorithm is explored in the optimization process, and it leads to a lack of diversity in the model pool for the final ensemble.
The original optimization strategy deals with the weak points by applying the bandit-based algorithm. 
It evaluates each algorithm by trying different combinations of other hyperparameters so that it can further compute the expected utility of each algorithm. Meanwhile, since all algorithms are evaluated for some given budget, the evaluation history is diverse, which helps generate a better model ensemble.
\section{Further Optimization with Meta-learning}
\label{sec:metalearning}
One class of optimizations that
we support is \textit{meta-learning}~\cite{DBLP:journals/corr/abs-1810-03548,Vilalta2002} --- given previous runs of the system 
over similar workloads, to transfer
the knowledge and better help the
workload at hand. Depending on 
the type of different building 
blocks, we support different 
meta-learning strategies. 

\subsection{Meta-learning for Conditioning Blocks}
For conditioning block on variable $x$,
it introduces a multi-armed bandit problem 
with $|\mathbb{D}_x|$ arms, and its objective is to identify the optimal arm.
One natural meta-learning strategy is to learn, given the dataset $D$, a much smaller subset of arms $A \subseteq \mathbb{D}_x$ that includes the optimal arm. 
This could explicitly reduce the search space in the conditioning block from $\mathbb{D}_x$ to $A$.
We use a meta-learning strategy
based on RankNet~\cite{ranknet}.

During the training process, we
are given a training history
over multiple previous datasets
$D_1,...,D_n$. We are given 
the relative relationships between
different arms on different datasets
\begin{equation}
\mathcal{T} = \{(A_j, A_k, D_i): A_j, A_k \in \mathbb{D}_x\},    
\end{equation}
where $(A_j, A_k, D_i) \in \mathcal{T}$
means that $A_j$ performs better than
$A_k$ on dataset $D_i$.
We are also given a meta-feature extractor
$h_D$ for dataset and a meta-feature 
extractor $h_A$ for arms.
Both types of extractors will map a
dataset (resp. an arm)
to an $m$-dimensional real-valued vector.
The model that we are trying to learn is a multi-layer perceptron (MLP) model taking as
input a dataset embedding and
an arm embedding, with the following learning objective:
\begin{equation}
\begin{aligned}
\min_{\Theta} 
  \sum_{(D_i, A_j, A_k) \in \mathcal{T}}
   l_+ \left( \sigma (r_j^{(i)} - r_k^{(i)}) \right)
   + l_- \left( \sigma (r_k^{(i)} - r_j^{(i)}) \right) & \\
  \text{where}~~~ r_j^{(i)} = MLP(h_D(D_i), h_A(A_j); \Theta), & \\
  r_k^{(i)} = MLP(h_D(D_i), h_A(A_k); \Theta), & \\
\end{aligned}
\end{equation}
where $\sigma$ is the sigmoid function, $l_+$ is the hinge loss with positive label, and $l_-$ is the hinge loss with negative label.

During inference, the MLP with parameter $\Theta$ takes the vector that consists of $h_D$ and $h_A$ as input, and outputs a score. The best 
subset of arms can then be selected based on these scores. 

\subsection{Meta-learning for Joint Blocks}
A joint block uses BO method that can be slow when the underlying search space is large.
An intuitive optimization is to leverage BO history $H_1=\{(x_j^1,y_j^1)\}_{j=1}^{n^1}, ...,H_n$ from $n$ previous datasets $D_1, ..., D_n$. 
This motivates the meta-learning based BO that can speed up the convergence of search in the current joint block. 

When executing joint block on a new dataset, we are given the historical observations $H_1, ...,H_n$ from $n$ previous datasets on the same search space, and the observations in the current task is $H_{T}$.
We use a scalable meta-learning method, RGPE~\cite{feurer2018scalable}, to accelerate BO.
First, for each previous dataset $D_i$, we train a Gaussian process model $M_i$ on the corresponding observations from $H_i$. 
Then we build a surrogate model $M_{\text{meta}}$ to guide the search in this joint block, instead of the original surrogate $M_T$ fitted on $H_{T}$ only.
The prediction of $M_{\text{meta}}$ at point $\bm{x}$ is given by 
\begin{equation}
y \sim \mathcal{N}(\sum_{i}w_i\mu_i(\bm{x}),\sum_{i}w_i\sigma^2_i(\bm{x})),  
\end{equation}
where $w_i$ is the weight of base surrogate $M_i$, and $\mu_i$ and $\sigma^2_i$ are the predictive mean and variance from base surrogate $M_i$.
The weight $w_i$ reflects the similarity between the previous task and current task.
Therefore, $M_{\text{meta}}$ carries the knowledge of search on previous tasks, which can greatly accelerate the convergence of the search in current joint block.
We then use the following ranking loss function $L$, i.e., the number of misranked pairs, to measure the similarity between previous tasks and current task:
\begin{equation}
    L(M_i,H_T) = \sum_{j=1}^{n^{T}}\sum_{k=1}^{n^{T}}\mathbbm{1}((M_i(\bm{x}_j) < _i(\bm{x}_k) \oplus (y_j < y_k)),
    \label{rank_loss}
\end{equation}
where $\oplus$ is the exclusive-or operator, $n_T=|H_T|$, $\bm{x}_j$ and $y_j$ are the sample point and its performance in $H^T$, and $M_i(\bm{x}_j)$ means the prediction of $M_i$ on point $\bm{x}_j$.
Based on the ranking loss function, the weight $w_i$ is set to the probability that $M_i$ has the smallest ranking loss on $H_T$, that is,
$w_i = \mathbbm{P}(i=\operatorname{argmin}_{j}L(M_j, H_T))$.
This probability can be estimated using MCMC sampling.

\paragraph{\underline{Example}.} When applied to
end-to-end AutoML, the joint block is used to select configurations from a joint search space, e.g., the search of hyper-parameter configurations or features given a specific ML algorithm. 
Although the optimal configuration may be different across tasks, the performance surface of configurations in current task may be similar to some in previous tasks due to the relevancy between tasks. 
In this case, BO history on previous datasets can be utilized to guide the configuration search via the above meta-learning based BO method.

\section{Experimental Evaluation}
\label{sec:experiments}

We compare \sys with state-of-the-art AutoML systems. In our evaluation, we focus on three perspectives: (1) the \emph{performance} of \sys given the \emph{same} search space explored by existing systems, (2) the \emph{scalability} of \sys given larger search spaces, and (3) the \emph{extensibity} of \sys to integrate new components into the search space of AutoML pipelines.


\begin{figure*}
	\centering
	\subfigure[\sys vs. AUSK on CLS]{
		\scalebox{0.42}{
			\includegraphics[width=1\linewidth]{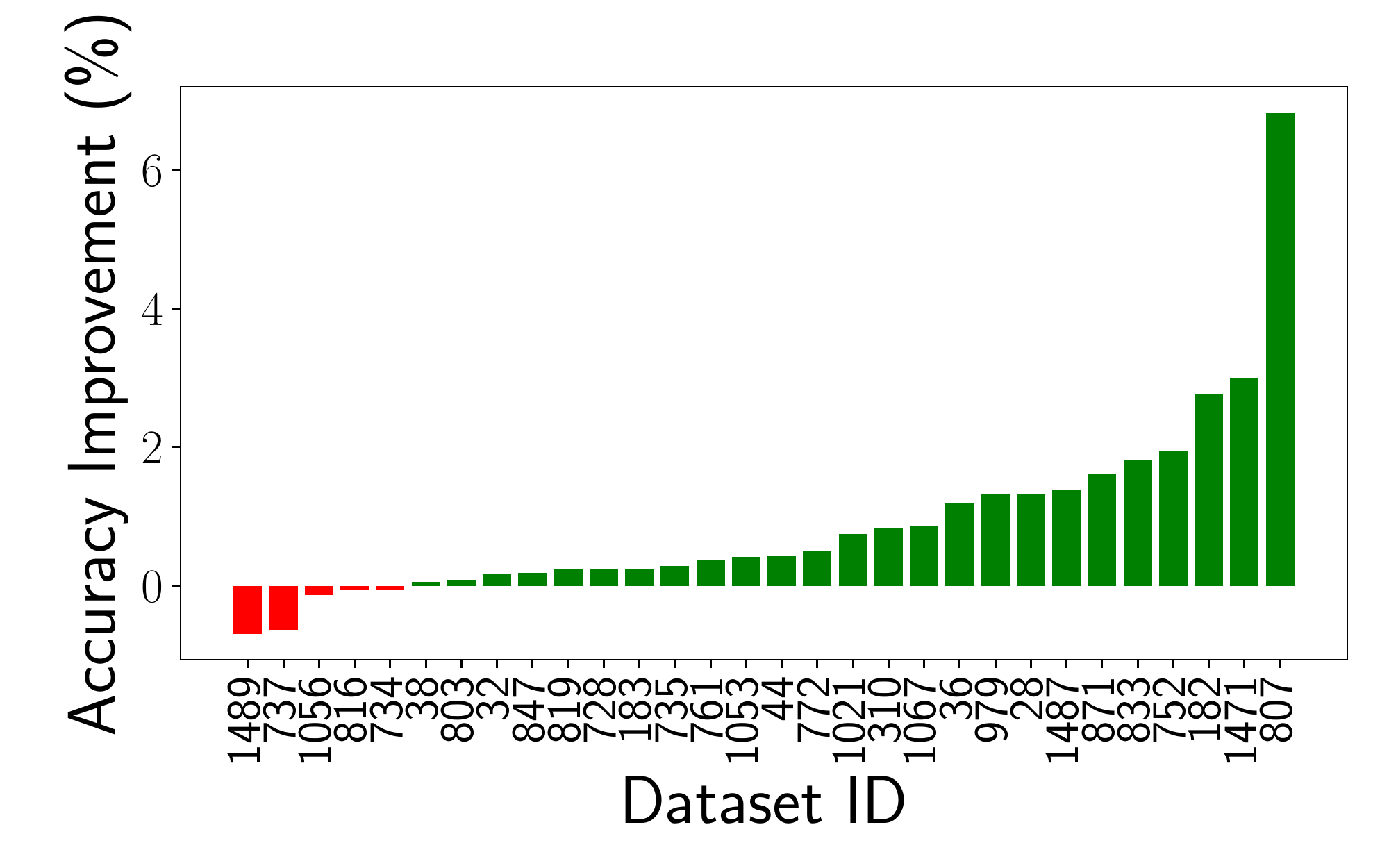}
	}}
	\subfigure[\sys vs. AUSK on REG]{
		\scalebox{0.325}{
		\includegraphics[width=1\linewidth]{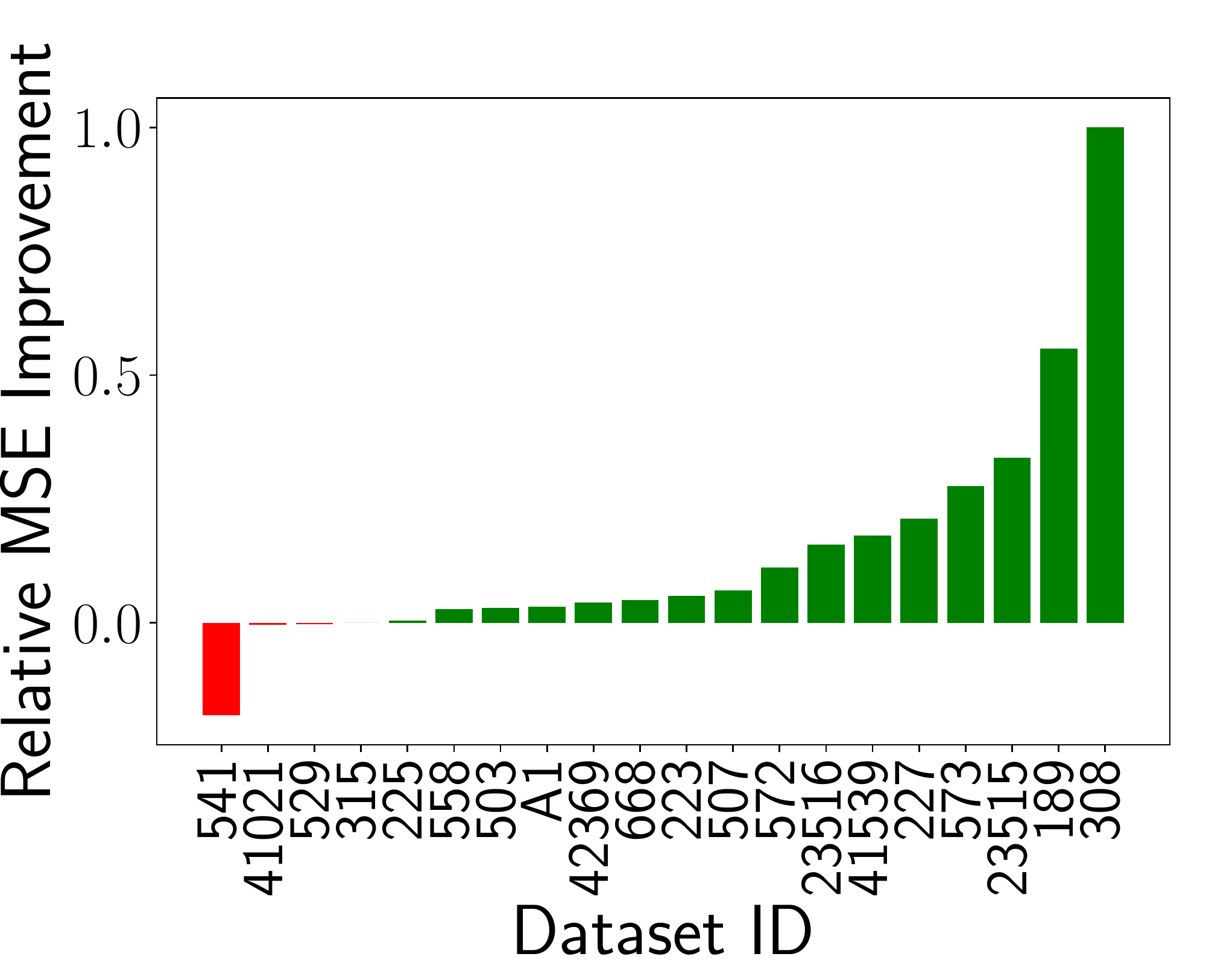}
	}}
	\subfigure[\sys vs. TPOT on CLS]{
		\scalebox{0.42}{
		\includegraphics[width=1\linewidth]{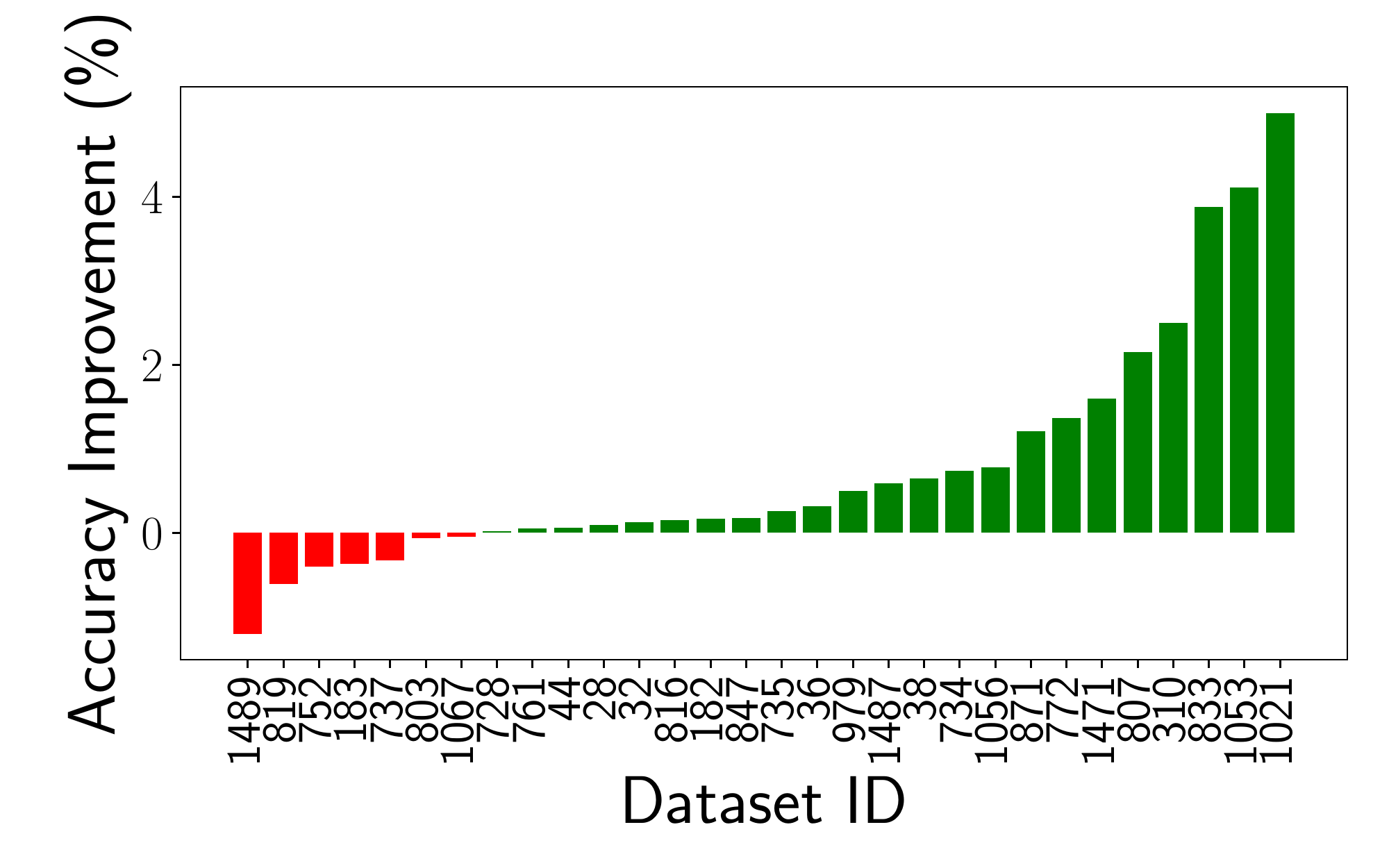}
	}}
    \subfigure[\sys vs. TPOT on REG]{
		\scalebox{0.325}{
		\includegraphics[width=1\linewidth]{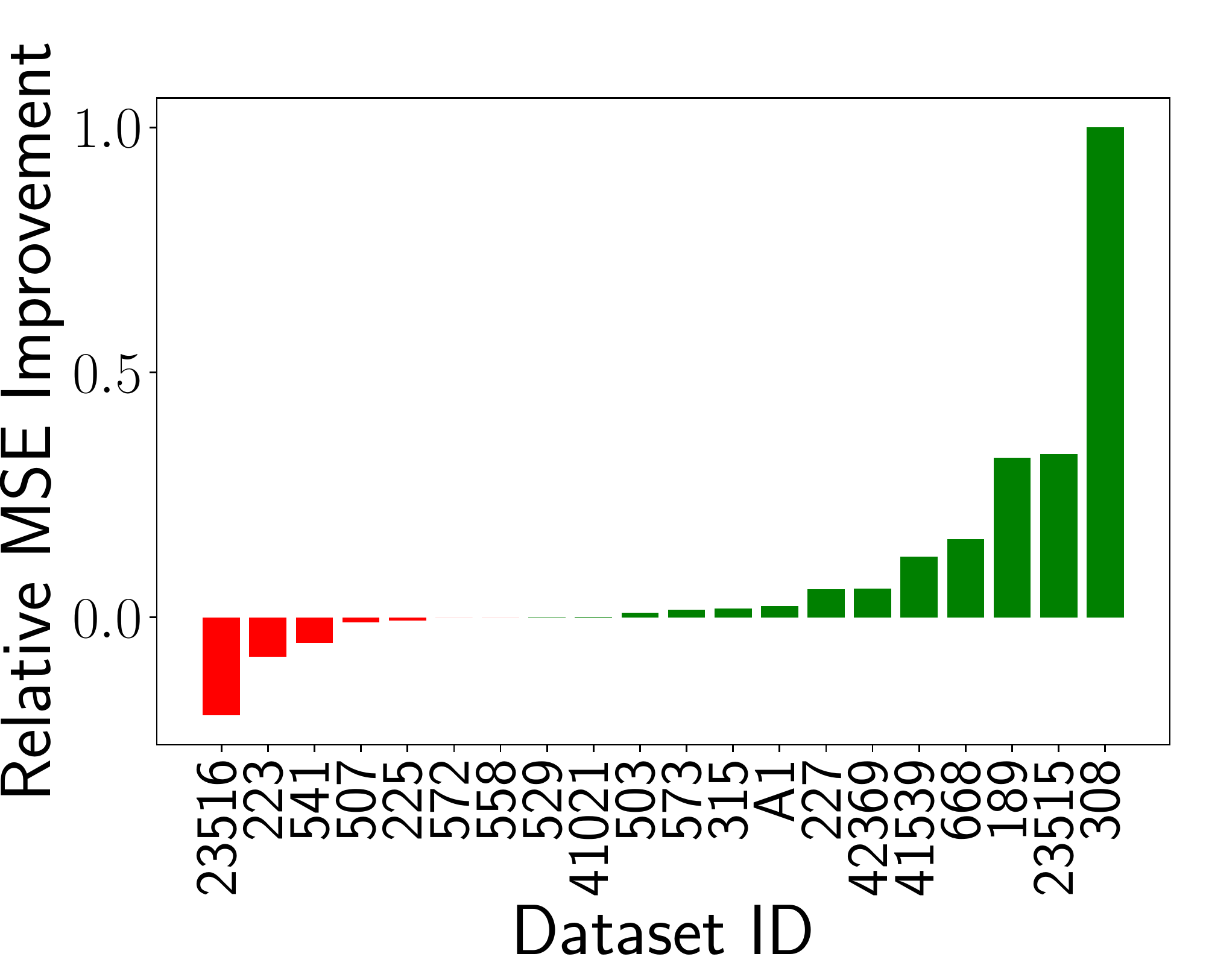}
	}}
	\caption{End-to-End results on 30 OpenML classification (CLS) datasets and 20 OpenML regression (REG) datasets.}
  \label{end2end_cmp}
\end{figure*}

\begin{figure*}
	\centering
	\subfigure[Mnist\_784]{
		\scalebox{0.42}{
			\includegraphics[width=1\linewidth]{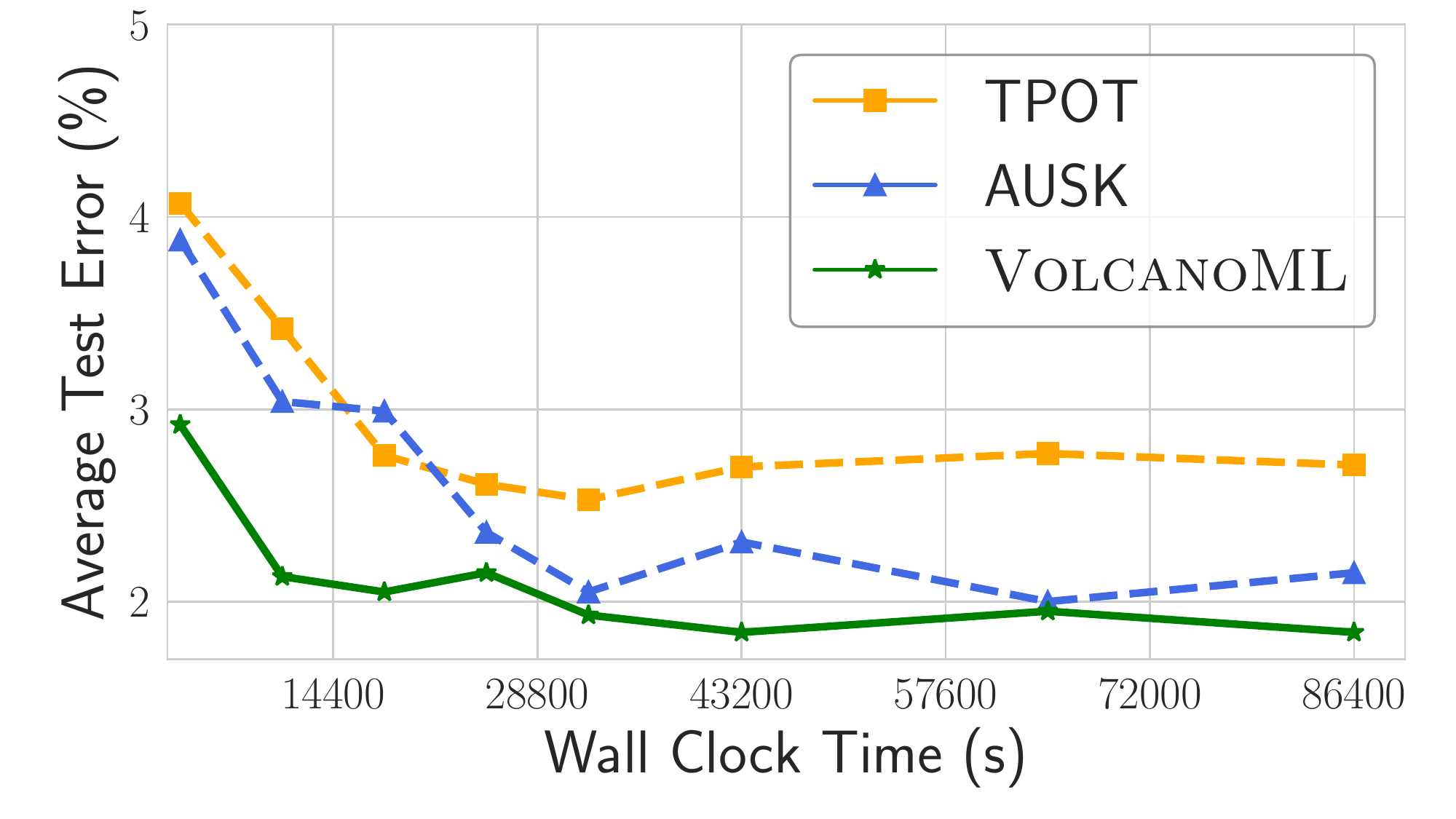}
	}}
	\subfigure[Kropt]{
		\scalebox{0.42}{
			\includegraphics[width=1\linewidth]{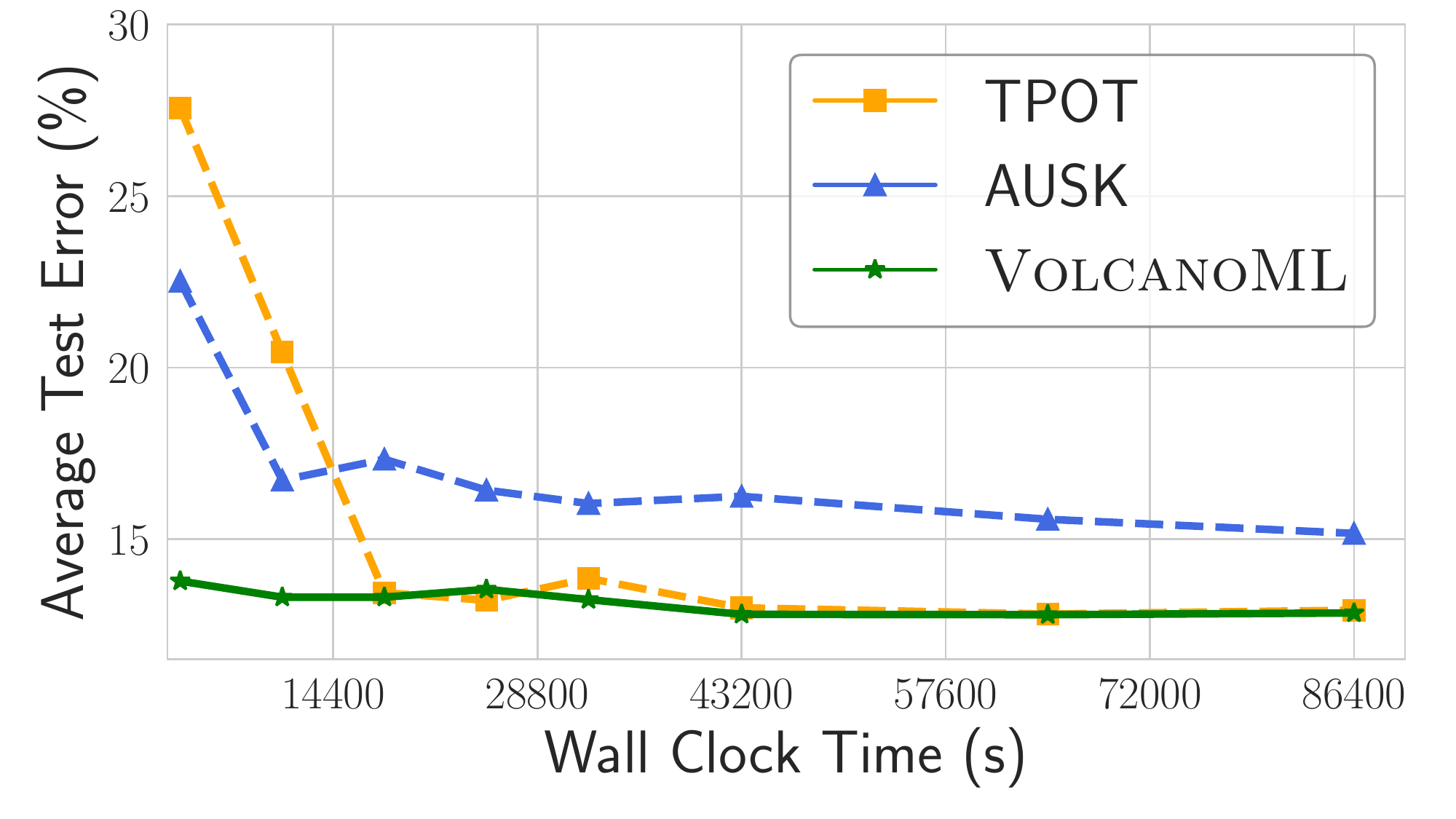}
	}}
	\subfigure[Electricity]{
		\scalebox{0.42}{
			\includegraphics[width=1\linewidth]{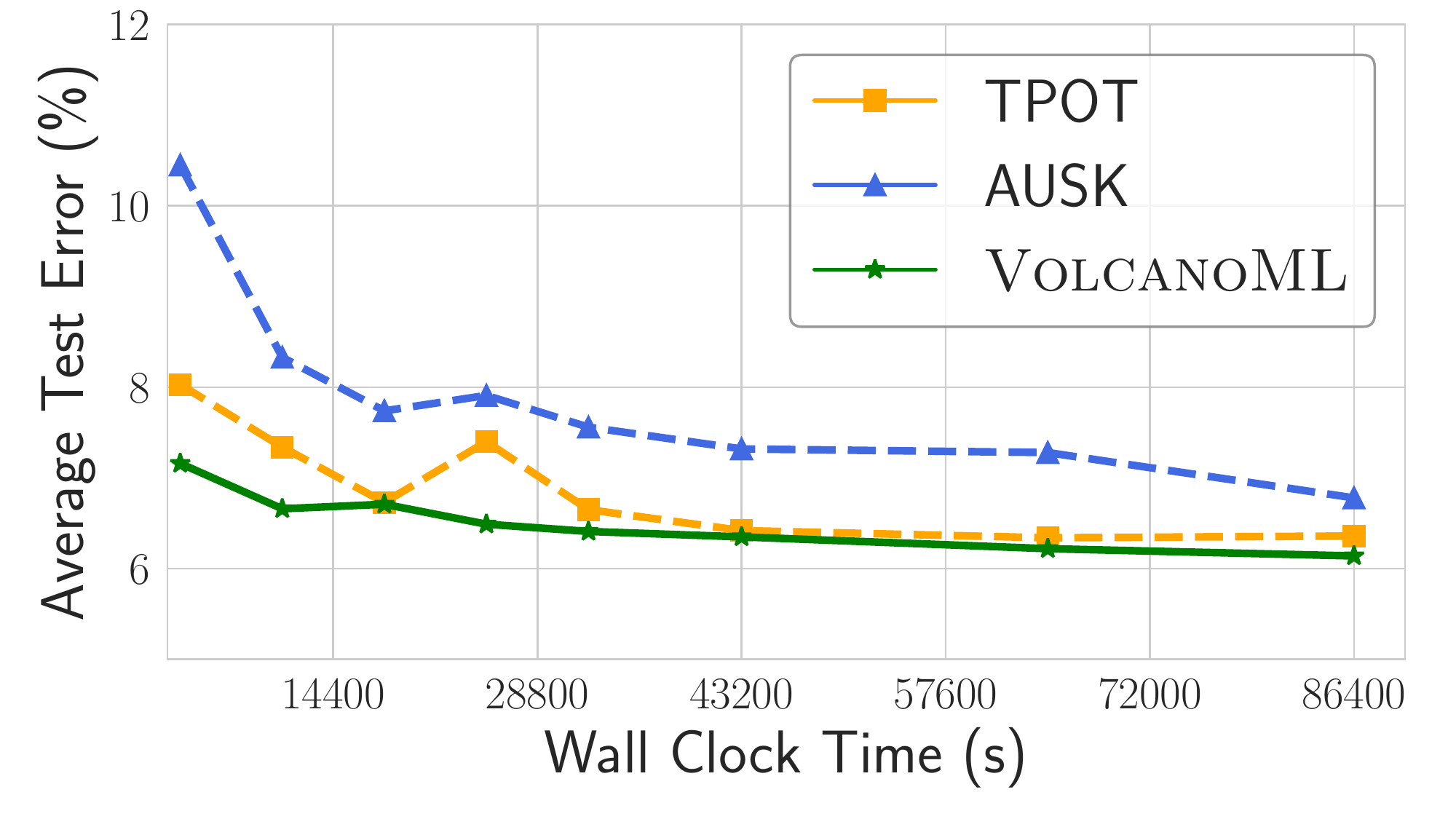}
	}}
    \subfigure[Higgs]{
		\scalebox{0.42}{
			\includegraphics[width=1\linewidth]{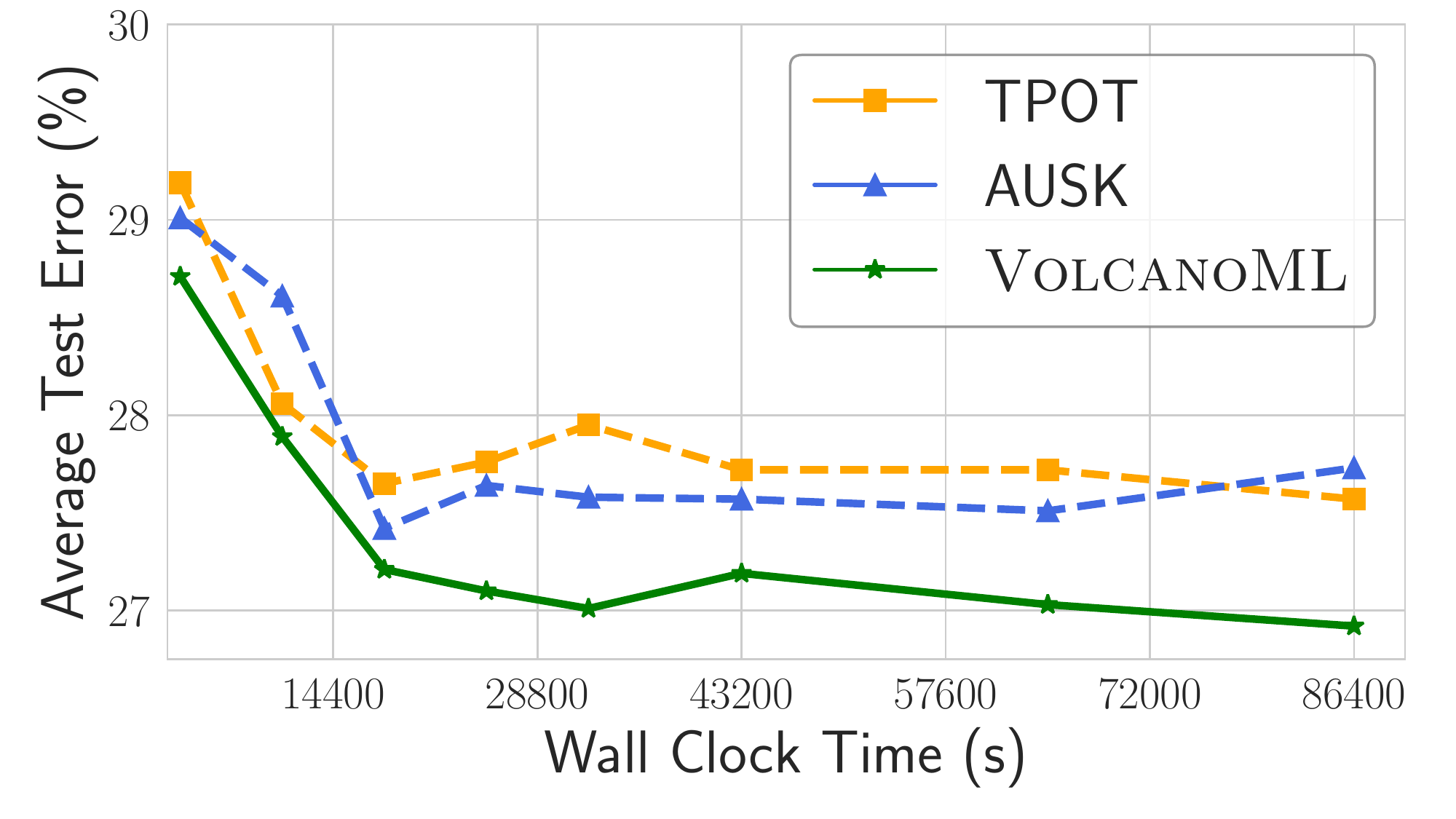}
	}}
	\caption{Average test errors on four large datasets with different time budgets.}
  \label{fig:main_test_speedups}
\end{figure*}

\subsection{Experimental Setup}
\para{\textbf{AutoML Systems.}}
We evaluate \sys as well as two open-source AutoML systems: \texttt{auto-sklearn}~\cite{feurer2015efficient} and \texttt{TPOT}~\cite{olson2019tpot}.
In addition, we also compare \sys with four commercial AutoML platforms from Google, Amazon AWS, Microsoft Azure, and Oracle. 
Both \sys and \texttt{auto-sklearn} support meta-learning, while \texttt{TPOT} does not.
For fair comparison with \texttt{TPOT}, we also use $\sys^{-}$ and AUSK$^{-}$ to denote the versions of \sys and \texttt{auto-sklearn} when meta-learning is disabled.
Our implementation of \sys is available at \url{https://github.com/PKU-DAIR/mindware}.

\para{\textbf{Datasets.}}
To compare \sys with academic baselines, we use 60 real-world ML datasets from the OpenML repository~\cite{10.1145/2641190.2641198}, including 40 for classification (CLS) tasks and 20 for regression (REG) tasks. 
10 of the 40 classification datasets are relatively large, each with 20k to 110k data samples; the other 30 are of medium size, each with 1k to 12k samples. 
In addition, we also use datasets from six Kaggle competitions (See Table~\ref{tbl:kaggle_dataset} for details) to compare \sys with four commercial platforms.

\para{\textbf{AutoML Tasks.}}
We define three kinds of real-world AutoML tasks, including (1) a general classification task on 30 medium datasets, (2) a general regression task on 20 medium datasets, and (3) a large-scale classification task on 10 large datasets. 

To test the scalability of the participating systems, we design three search spaces that include 20, 29, and 100 hyper-parameters, where the smaller search space is a subset of the larger one.
We run \sys and the baseline AutoML systems against each of the three search spaces.
The time budget is 900 seconds for the smallest search space and 1,800 seconds for the other two, when performing the general classification task (1); the time budget is increased to 5,400 and 86,400 seconds respectively, when performing the general regression task (2) and the large-scale classification task (3).


\para{\textbf{Utility Metrics.}}
Following~\cite{feurer2015efficient}, we adopt the metric \emph{balanced accuracy} for all classification tasks --- compared with standard (classification) accuracy, it assigns equal weights to classes and takes the average of class-wise accuracy.
For regression tasks, we use the \emph{mean squared error} (MSE) as the metric.

In our evaluation, we repeat each experiment 10 times and report the average utility metric.
In each experiment, we use four fifths of the data samples in each dataset to search for the best ML pipeline and report the utility metric on the remaining fifth.

\begin{table*}[tb]
\small
\centering
  \caption{Average ranks on 30 classification (CLS) datasets and 20 regression (REG) datasets with three different search spaces. (The lower is the better)}
\begin{tabular}{l|ccccc}
\toprule
Search Space - Task & TPOT & AUSK$^{-}$ & AUSK & $\sys^{-}$  & \sys \\
\hline
Small - CLS   & 3.09  & 3.07 & 3.01 & 2.94 & \textbf{2.89}  \\
Medium - CLS  & 3.2  & 3.32  & 3.27  & 2.78  & \textbf{2.43}  \\
Large - CLS  & 3.29  & 3.77 & 3.57 & 2.72 & \textbf{1.65} \\
\hline
Small - REG  & \textbf{2.98}  & 3.02 & 3.0 & 3.02 & \textbf{2.98} \\
Medium - REG & 2.95  & 3.3  & 3.12  & \textbf{2.75}  & 2.88  \\
Large - REG  & 3.1   & 3.85 & 3.82 & 2.15 & \textbf{2.08} \\
\bottomrule
  \end{tabular}
  \label{scalability_result}
\end{table*}

\para{\textbf{Methodology for Comparing AutoML Systems.}}
To compare the overall test result of each AutoML system on a wide range of datasets, we use the \emph{average rank} as the metric following~\cite{bardenet2013collaborative}.
For each dataset, we rank all participant systems based on the result of the best ML pipeline they have found so far; we then take the average of their ranks across different datasets.
In addition, we use statistical testing
to determine ties and adjust the rankings~\cite{dewancker2016strategy}.

\para{Training Data for Meta-learning.}
The results for meta-learning are obtained from running Bayesian optimization on 90 classification datasets and 50 regression datasets collected from OpenML. 
For classification, we collect the results by optimizing the balanced accuracy, accuracy, f1 score and AUC.
For regression, we collect the results by optimizing the mean squared error, mean absolute error and r2 value.
When \sys receives a new task and the optimization target is one of the above metrics, \sys will use all the evaluation results with this metric to train the RankNet in the conditional block and RGPE in the joint block.
In our experiments, to ensure the current task does not occur in the results for meta-learning, we apply the leave-one-out strategy.
For example, when we optimize Dataset A, we will use all other results except A for meta-learning. 

\para{\textbf{More Details.}}
We include the details of search space and programming API in Appendix \ref{appendix:system_components}, experiment datasets in Appendix~\ref{appendix:results}.

\subsection{End-to-End Comparison}
We first evaluate the participant AutoML systems given the search space explored by \texttt{auto-sklearn}. 
Figure~\ref{end2end_cmp} presents the results of \sys compared to \texttt{auto-sklearn} (AUSK) and \texttt{TPOT} on the 30 classification tasks and the 20 regression tasks, respectively.
For classification tasks, we plot the classification accuracy improvement (\%); for regression tasks, we plot the \emph{relative MSE improvement} $\Delta$, which is defined as     $\Delta(m_1, m_2) = \frac{s(m_2) - s(m_1)}{\operatorname{max}(s(m_2), s(m_1))}$, where $s(\cdot)$ is MSE on the test set. 
Overall, \sys outperforms \texttt{auto-sklearn} and \texttt{TPOT} on 25 and 23 of the 30 classification tasks, and on 17 and 15 of the 20 regression tasks, respectively.

We also conduct experiments to evaluate \sys with different \textit{time budgets}. Figure~\ref{fig:main_test_speedups} presents the results on four large classification datasets.
We observe that \sys exhibits consistent performance over different time budgets. 
Notably, on \texttt{Higgs}, \sys achieves $27.2$\% test error within 4 hours, which is better than the performance of the other two systems given 24 hours.
Furthermore, we also design additional experiments to evaluate the consistency of system performance given different (larger) time budgets and search spaces, and more details can be found in the following sections.

We further study the scalability of the participant systems on the three aforementioned search spaces.
Without meta-learning, \sys achieves the best average rank for both the classification and regression tasks --- on the small search space (with 20 hyper-parameters), \sys performs slightly better than \texttt{auto-sklearn} and \texttt{TPOT}, and it performs significantly better on the medium (with 29 hyper-parameters) and large (with 100 hyper-parameters) search spaces. 
For meta-learning, we present more empirical results and discussions in Section~\ref{sec:meta-learning}.

\subsection{Search Space Enrichment}
We now evaluate the \emph{extensibility} of \sys via two experiments with \emph{enriched} search spaces.

\textit{Adding \texttt{Data\_Balancing} Operator.}
In the first experiment, we implement the feature engineering operator ``\texttt{smote\_balancer}'' -- a popular over-sampling techinique proposed for the overfitting problem, and incorporate it into the aforementioned \textit{balancing} stage of feature engineering (FE) (Section~\ref{sec:building-blocks:search-space}).
Note that \texttt{auto-sklearn} cannot support this fine-grained enrichment of the search space.
Table~\ref{enrichment_smote} presents the results of \texttt{auto-sklearn}, \sys without enrichment, and \sys with enrichment, on five imbalanced datasets.
We observe that enriching the search space brings further improvement, e.g., \sys with enrichment outperforms \texttt{auto-sklearn} by 3.57\% (balanced accuracy) on the dataset \texttt{pc2}.

\textit{Supporting Embedding Selection.}
In the second experiment, we add a new stage ``embedding selection'' into the FE pipeline, with two candidate embedding-extraction operators (i.e., two pre-trained models).
This allows \sys to deal with images, which is difficult for \texttt{auto-sklearn} and \texttt{TPOT} to support using existing code.
We implement two pre-trained models to generate embeddings for images, and we evaluate \sys with the enriched search space on the Kaggle dataset \texttt{dogs-vs-cats}.
We observe that \sys achieves 96.5\% test accuracy, which is significantly better than 70.4\% obtained by \sys without considering embeddings.



\begin{figure*}
	\centering
	\subfigure[Influence Network]{
		\scalebox{0.48}{
			\includegraphics[width=0.95\linewidth]{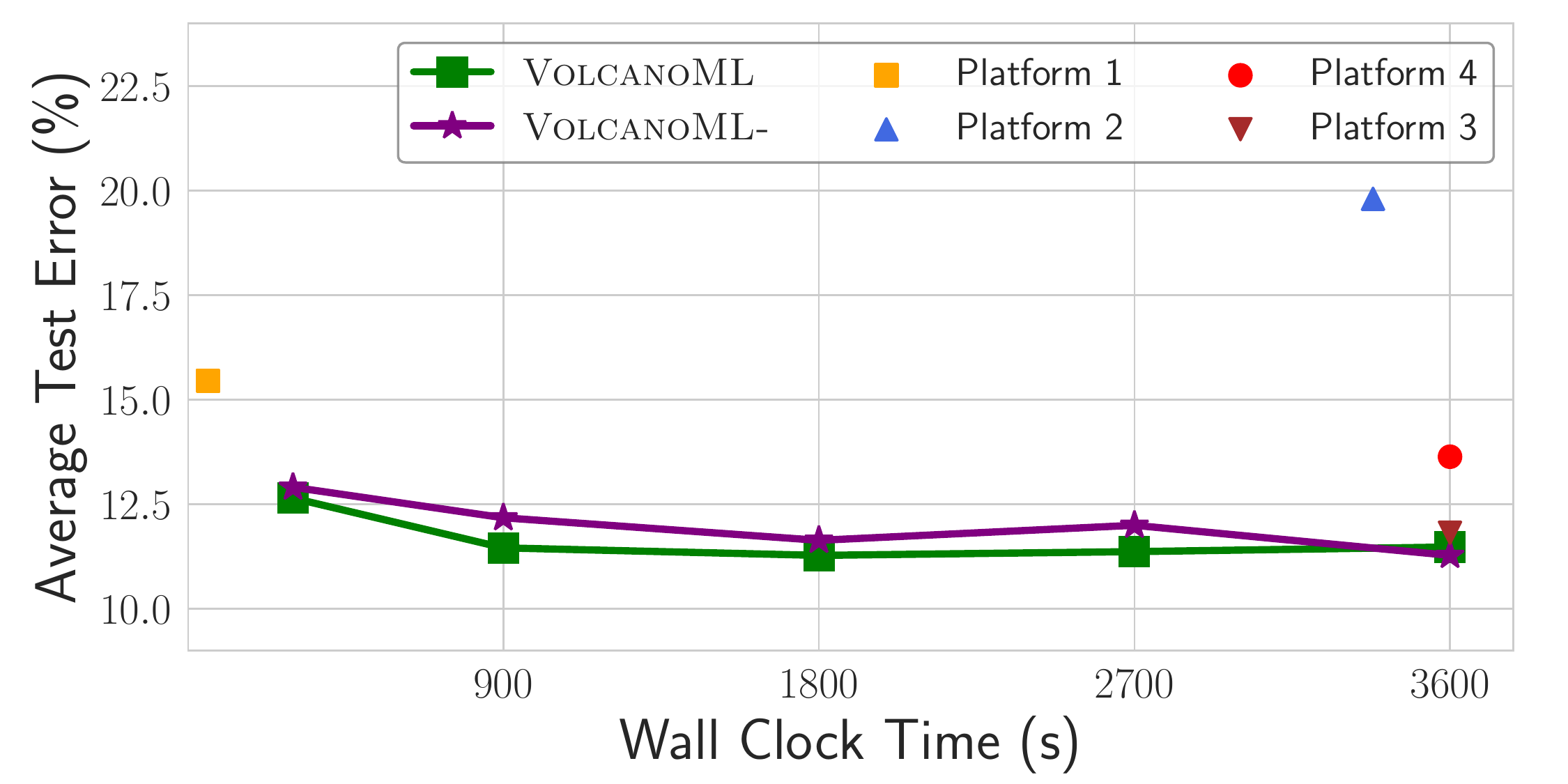}
	}}
	\subfigure[Virus Prediction]{
		\scalebox{0.48}{
			\includegraphics[width=0.95\linewidth]{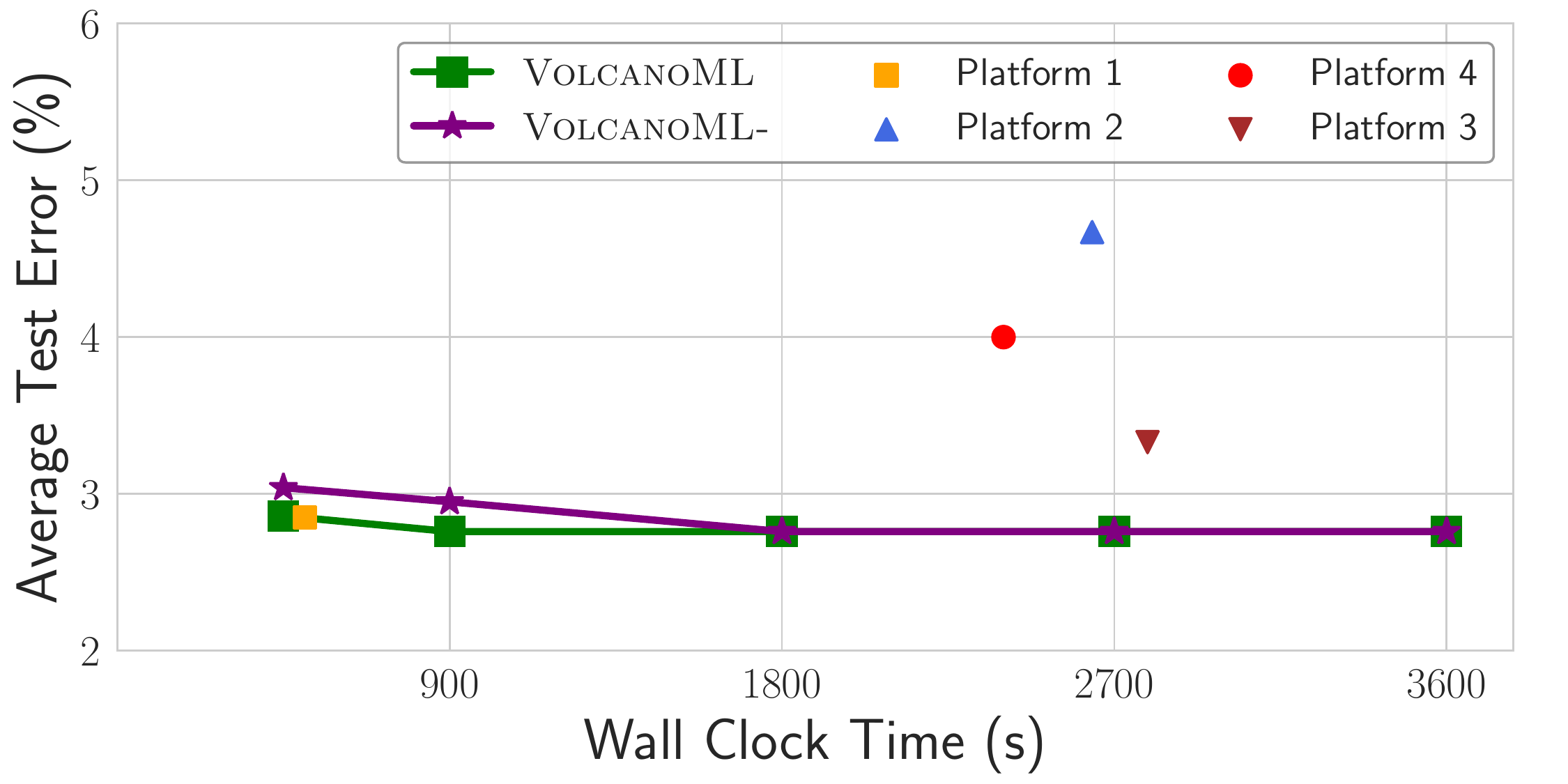}
	}}
	\subfigure[Employee Access]{
		\scalebox{0.48}{
			\includegraphics[width=0.95\linewidth]{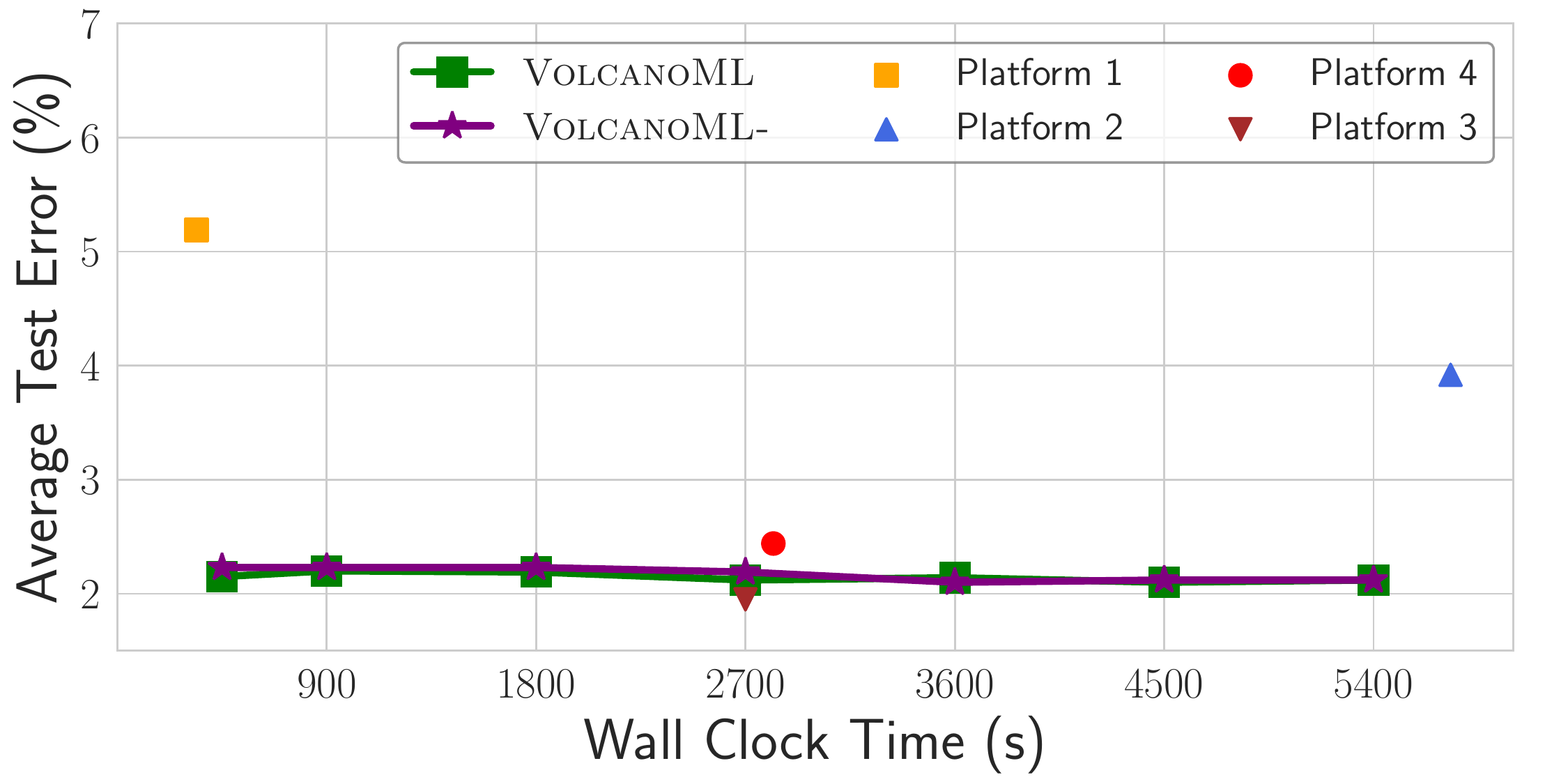}
	}}
    \subfigure[Customer Satisfaction]{
		\scalebox{0.48}{
			\includegraphics[width=0.95\linewidth]{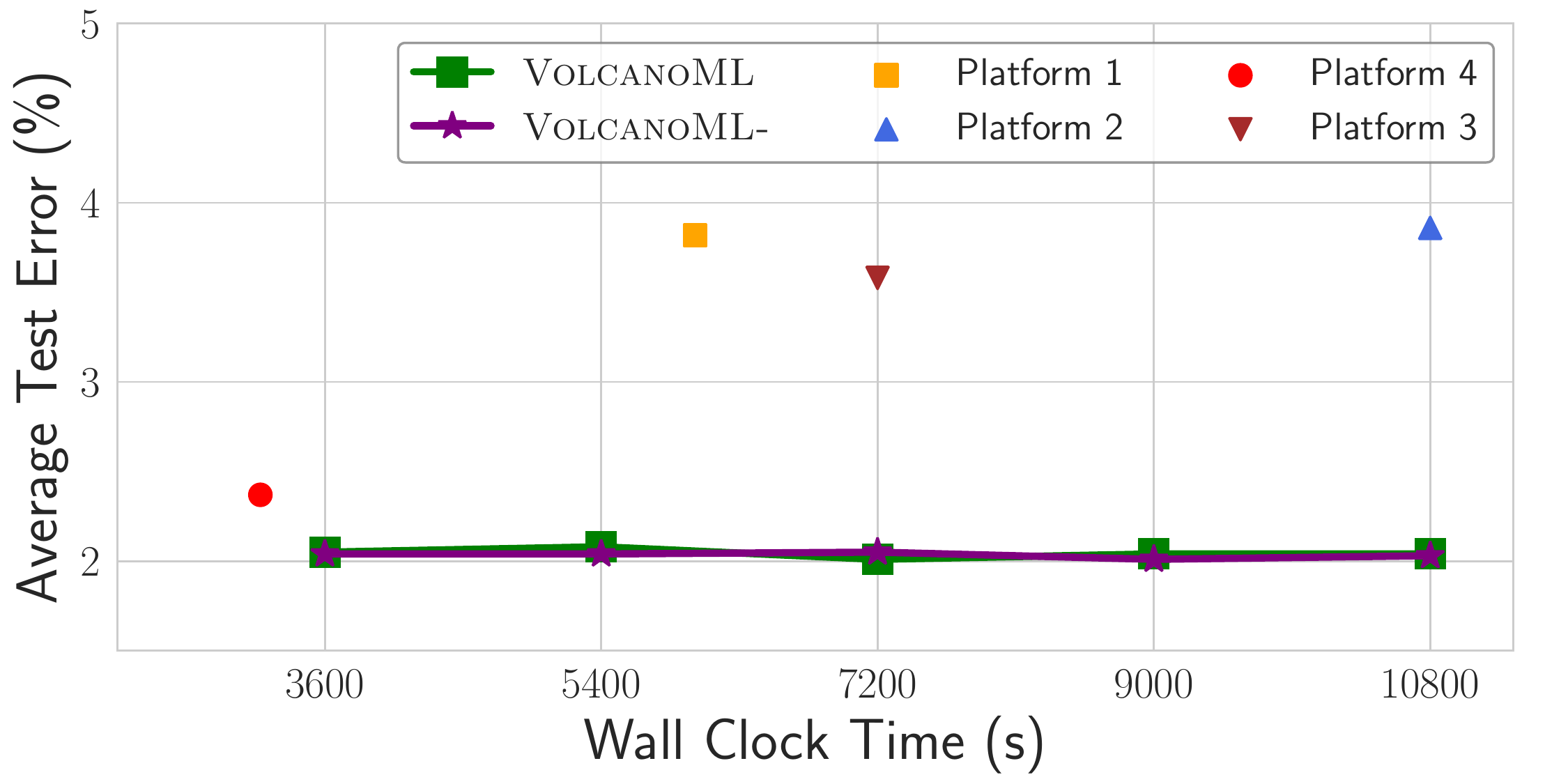}
	}}
	\subfigure[Business Value]{
		\scalebox{0.48}{
			\includegraphics[width=0.95\linewidth]{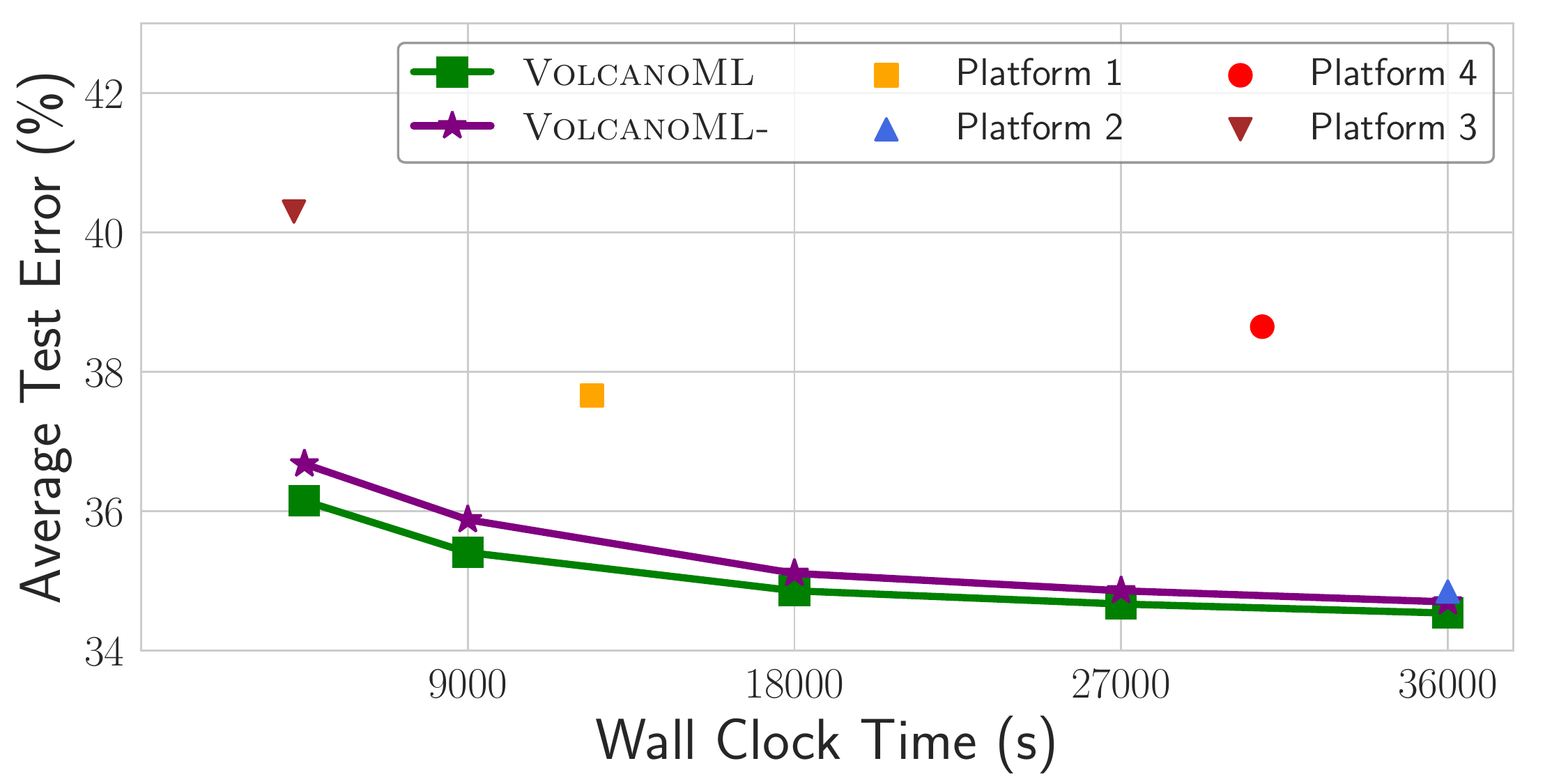}
	}}
	\subfigure[Flavours]{
		\scalebox{0.48}{
			\includegraphics[width=0.95\linewidth]{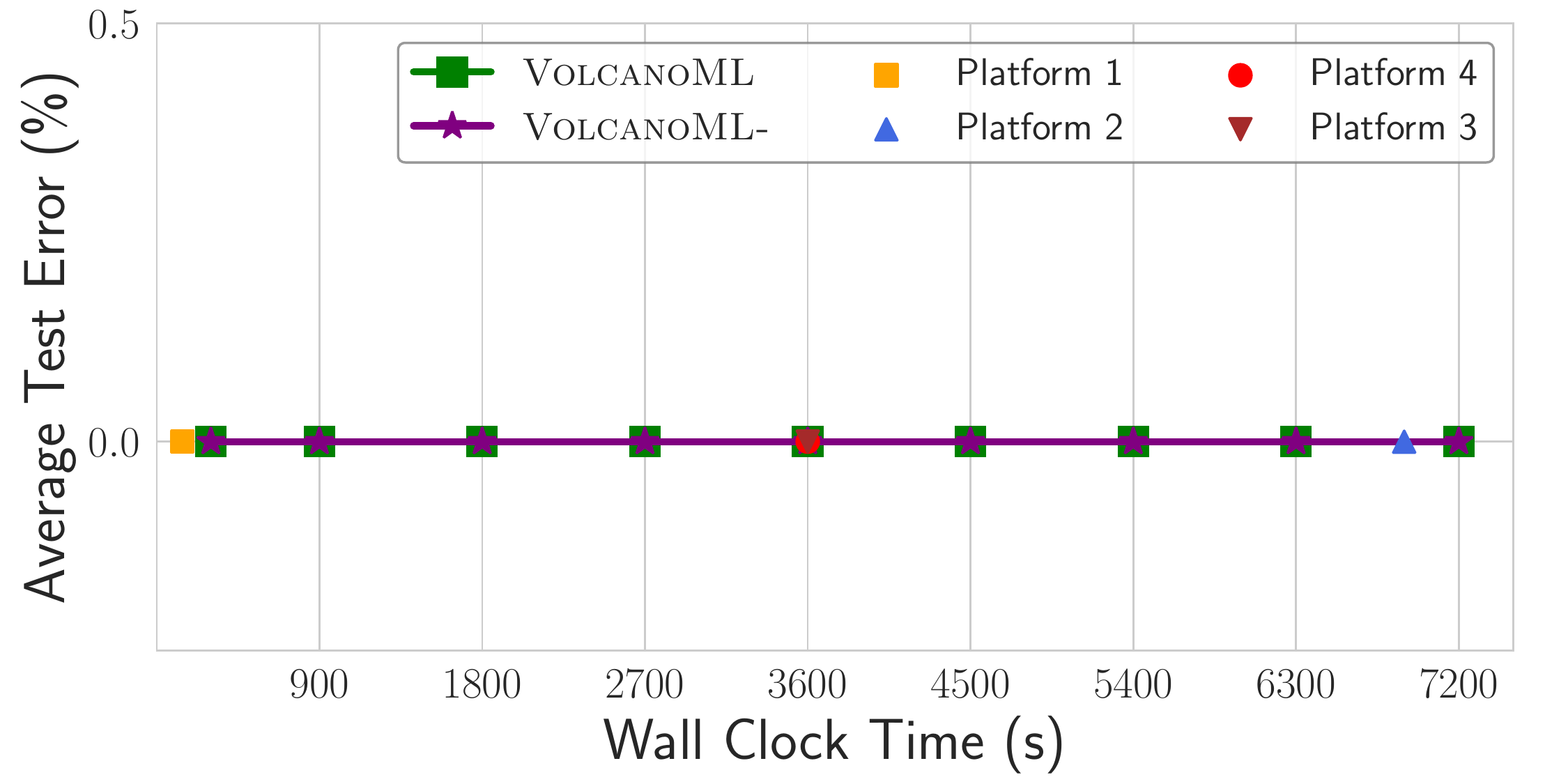}
	}}
	\caption{Test error on Kaggle competitions compared with commericial baselines.}
  \label{fig:industrial_results}
\end{figure*}

\begin{table}[htb]
\centering
\small
\caption{Test accuracy (\%) of \sys with and without the enrichment of ``\texttt{smote\_balancer}'' operator.}
\begin{tabular}{lccc}
    \toprule
    Dataset & AUSK & $\sys^{-}$ & \sys \\ 
    \midrule
    sick            & 97.29 & 97.31 & \textbf{97.34}\\
    pc2             & 86.70  & 86.91  & \textbf{90.27}\\
    abalone         & 66.86 & 65.97 & \textbf{67.32}\\
    page-blocks(2)  & 94.70 & 95.29 & \textbf{96.69}\\
    hypothyroid(2)  & 99.62 & 99.64 & \textbf{99.64}\\
    \bottomrule
\end{tabular}

\label{enrichment_smote}
\end{table}

\subsection{Comparison with 4 Industrial Platforms}
We run additional experiments on six Kaggle competitions (See Table~\ref{tbl:kaggle_dataset} for dataset statistics) over four commercial baselines (AutoML services from Google, AWS, Azure and Oracle) as follows:
\begin{itemize}
    \item Google Cloud AutoML on unknown running environment (not transparent to users).
    \item AWS Sagemaker AutoPilot on an instance `ml.m5.4xlarge' with 16 Intel Xeon® Platinum 8175M processors and 64G memory.
    \item Azure Automated ML on two instances `STANDARD\_D12' with totally 8 unknown processors and 56G memory.
    \item Oracle Data Science on an instance `VM.Standard2.2' with 2 2.0 GHz Intel® Xeon® Platinum 8167M processors and 30G memory.
    \item \sys on an Ali-cloud instance `ecs.hfc6.2xlarge' with 8 3.10 GHz Intel® Xeon® Platinum 8269CY processors and 30G memory.
\end{itemize}

\begin{table}[ht]
\small
\centering
\caption{Kaggle dataset information.}
\resizebox{1\columnwidth}{!}{
\begin{tabular}{lccc}
    \toprule
    Datasets & Classes & Samples & Features\\ 
    \hline
    Influencers in Social Networks & 2 & 5500 & 22 \\
    West-Nile Virus Prediction & 2 & 10506 & 11 \\
    Employee Access Challenge & 2 & 32769 & 9 \\
    Santander Customer Satisfaction & 2 & 76020 & 369 \\
    Predicting Red Hat Business Value & 2 & 2197291 & 12 \\
    Flavors of Physics & 2 & 38012 & 49 \\
    \bottomrule
\end{tabular}
}

\label{tbl:kaggle_dataset}
\end{table}

Due to the different design principles (different hardware and parallelism) of commercial manufacturers, it is very hard to set up exactly the same environment settings. 
We set the the maximal time budget as 10 hours and use cost as an additional metric.
Here, we anonymously refer to these platforms as Platform 1-4.
Figure~\ref{fig:industrial_results} show the results of \sys and the platforms.
We observe that \sys- (without meta-learning) achieves satisfactory results compared with those cloud solutions on the six tasks.
Due to the large initial search space, \sys- performs slightly worse than \sys (with meta-learning) in the beginning on Influence Network, Virus Prediction, and Business Value.
Given more time budget (i.e., fix the x-axis to some time budget), \sys and \sys- show similar results and often outperform the considered commercial platforms.
This demonstrates \sys's effectiveness against the commercial AutoML baselines.


\subsection{Scalability on Different Search Space}
\label{sec:scalability}

\begin{table}
\centering
  \caption{Average ranks on 30 classification (CLS) datasets and 20 regression (REG) datasets with three different search spaces (The lower is the better). The budget is 1800 seconds for classification and 5400 seconds for regression.}
\begin{tabular}{l|ccc}
\toprule
Search Space - Task & TPOT & AUSK & \sys \\
\hline
Small - CLS   & 2.03 & 1.98 & 1.98\\
Medium - CLS  & 1.95 & 2.21 & \textbf{1.83}\\
Large - CLS  & 1.97 & 2.43 & \textbf{1.60} \\
\hline
Small - REG  & 2.00 & 2.00 & 2.00\\
Medium - REG & 2.05 & 2.30 & \textbf{1.65}\\
Large - REG  & 2.10 & 2.20 & \textbf{1.70}\\
\bottomrule
  \end{tabular}

  \vspace{-1em}
  \label{tbl:scalability_result1}
\end{table}

\begin{table}
\centering
  \caption{Average ranks on 30 classification (CLS) datasets and 20 regression (REG) datasets with three different search spaces (The lower is the better). The budget is 3600 seconds for classification and 10800 seconds for regression.}
\begin{tabular}{l|ccc}
\toprule
Search Space - Task & TPOT & AUSK & \sys \\
\hline
Small - CLS   & 2.03 & 1.97 & 2.0\\
Medium - CLS  & 1.90 & 2.27 & \textbf{1.83}\\
Large - CLS  & 2.03 & 2.53 & \textbf{1.43}\\
\hline
Small - REG  & 1.95 & 2.00 & 2.05\\
Medium - REG & 1.95 & 2.30 & \textbf{1.75}\\
Large - REG  & 1.98 & 2.28 & \textbf{1.75}\\
\bottomrule
  \end{tabular}

  \vspace{-1em}
  \label{tbl:scalability_result2}
\end{table}

\begin{table}
\centering
  \caption{Average ranks on 30 classification (CLS) datasets and 20 regression (REG) datasets with three different search spaces (The lower is the better). The budget is 7200 seconds for classification and 21600 seconds for regression.}
\begin{tabular}{l|ccc}
\toprule
Search Space - Task & TPOT & AUSK & \sys \\
\hline
Small - CLS   & 2.00 & 2.00 & 2.00\\
Medium - CLS  & 1.97 & 2.23 & \textbf{1.80}\\
Large - CLS  & 1.92 & 2.40 & \textbf{1.68}\\
\hline
Small - REG & 2.00 & 2.00 & 2.00 \\
Medium - REG & 1.95 & 2.23 & \textbf{1.83}\\
Large - REG  & 2.00 & 2.10 & \textbf{1.90}\\
\bottomrule
  \end{tabular}

  \vspace{-1em}
  \label{tbl:scalability_result3}
\end{table}


To evaluate the scalability of each system, we design three search spaces of different sizes. The small search space only contains four feature selectors (\emph{select percentile}, \emph{select generic univariate}, \emph{extra trees preprocessing}, and \emph{liblinear SVM preprocessing}) and uses \emph{random forest} as the ML algorithm. The medium search space contains the same four feature selectors as the small one and uses \emph{linear\_svc(r)}, \emph{random forest}, and \emph{AdaBoost} as the ML algorithms. The large search space is the entire search space described in Section~\ref{sec:building-blocks:search-space}. 
The three spaces include 20, 29, and 100 hyper-parameters, respectively, and the smaller space is a subset of the larger one.

To further investigate the result of \sys over 1) different time budgets and 2) different search spaces, we conducted additional experiments to run each system given 1800 / 5400 seconds, 3600 / 10800 seconds and 7200 / 21600 seconds for classification / regression tasks over the small, medium and large search spaces respectively.
These numbers are chosen by following the settings in papers of \texttt{auto-sklearn} and \texttt{TPOT}.
The experiments include 50 AutoML tasks (30 for classification and 20 for regression), and we use the metric --- average rank to measure each system.
Tables~\ref{tbl:scalability_result1}, ~\ref{tbl:scalability_result2} and ~\ref{tbl:scalability_result3} show the results over three different search spaces given different time budgets. 
We can observe that, with the increase of time budget and search space, \sys still achieves the best average rank, and performs better compared with \texttt{auto-sklearn} and \texttt{TPOT}.

\begin{figure}
	\centering
	\subfigure[Quake]{
		\scalebox{0.75}{
			\includegraphics[width=1\linewidth]{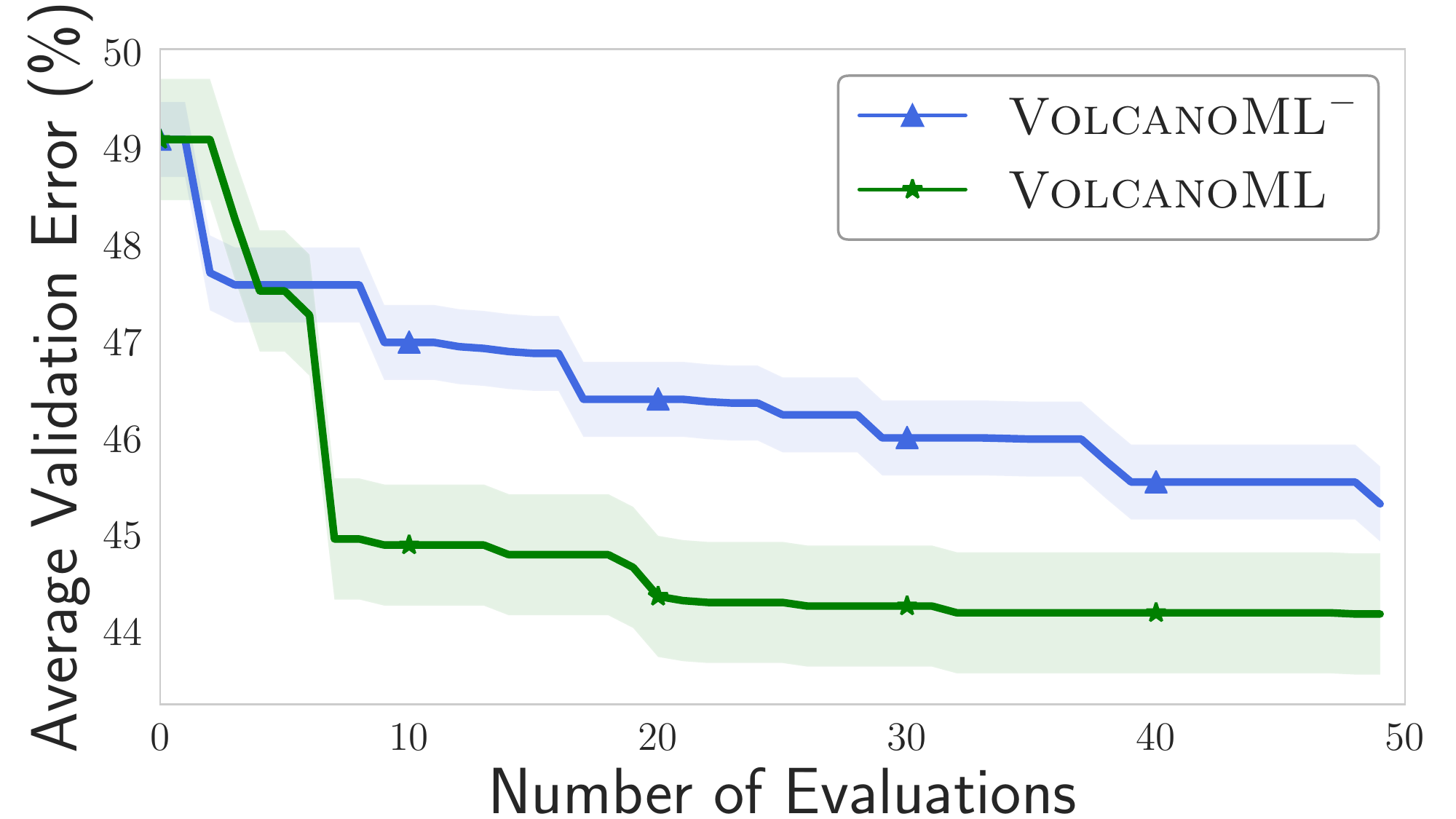}
	}}
	\subfigure[Space\_ga]{
		\scalebox{0.75}{
			\includegraphics[width=1\linewidth]{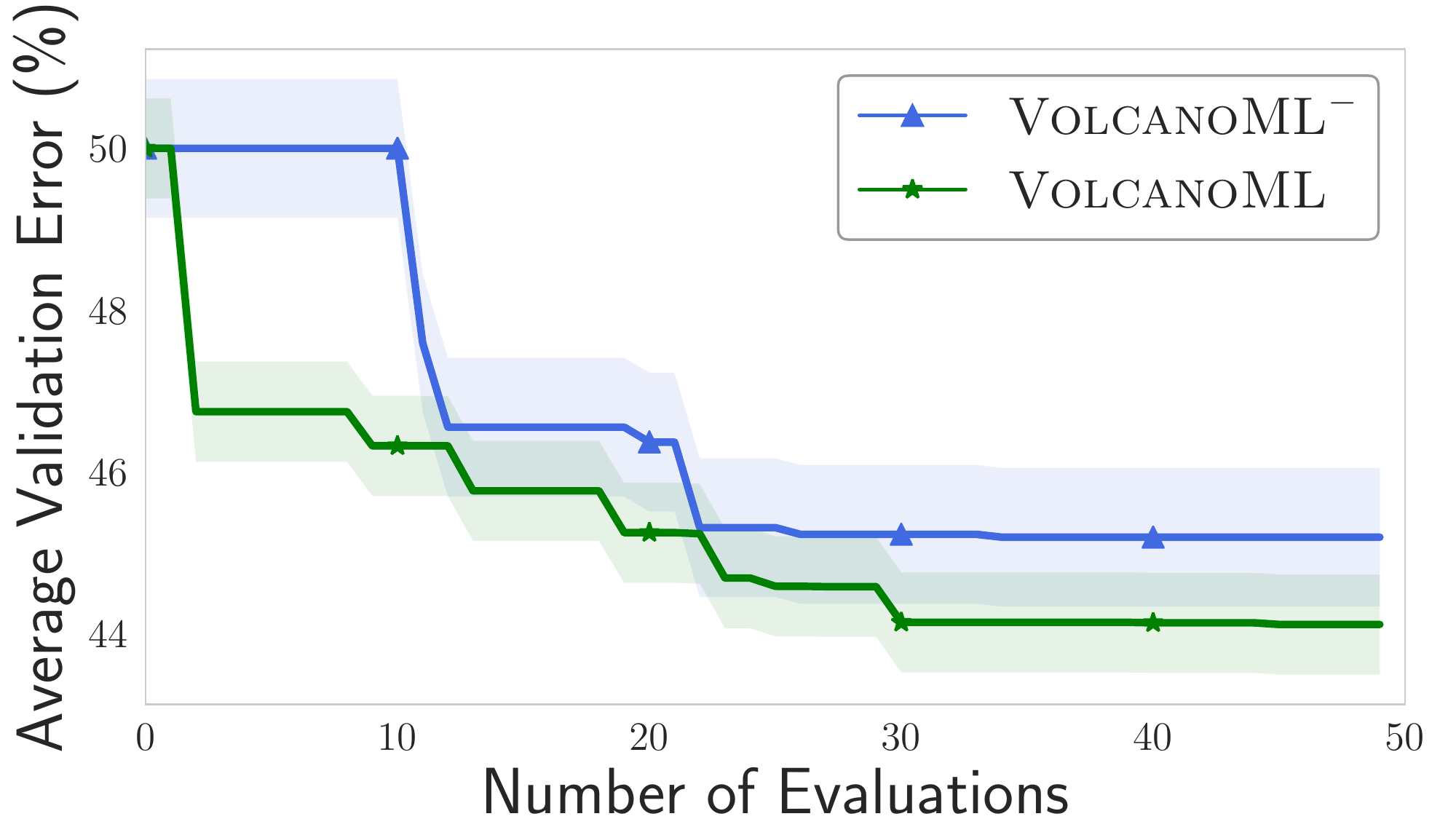}
	}}
	\caption{The result of the first 50 evaluations on `quake' and `space\_ga' using LibSVM. $\sys^{-}$ refers to \sys with meta-learning based BO disabled.}
  \label{tl}
\end{figure}

\subsection{Results about Meta-Learning based Optimization}
\label{sec:meta-learning}
\para{Meta-learning in Joint Block.} Figure~\ref{tl} shows the improvement of meta-learning in a joint block. Compared with $\sys^{-}$, the validation error drops significantly in the first 10 evaluations on \sys, which indicates that meta-learning captures the information of the historical tasks and performs an effective warm-start. When achieving the same validation error as vanilla $\sys^{-}$, \sys reduces the number of evaluations by eight-fold on \texttt{quake} and two-fold on \texttt{space\_ga}. 

\para{Meta-learning in Conditioning Block.} 
To compare the performance of RankNet, we also used \textit{LightGBM} to model the relationship between algorithm performance and tasks by transforming the ranking problem into a binary classification problem.
The input to both \textit{LightGBM} and RankNet is the same. 
We adopted 10-fold validation mechanism to evaluate each method; the meta-learner is learned on the training set, and validated on the validation set.
In addition, we measured the performance of each method using the metric `\emph{mAP@5}', which is the mean Average Precision to predict the top-5 algorithms.
RankNet and \textit{LightGBM} gets 0.87 and 0.62 mAP@5 score respectively. 
This demonstrates the more powerful expressiveness of neural networks (RankNet) than traditional ML algorithms (\textit{LightGBM}).

Table~\ref{scalability_result} also summarizes the performance of meta-learning in terms of the average ranks.
With meta-learning, the average rank of \sys is dramatically improved compared with \texttt{auto-sklearn}.
Overall, \sys with meta-learning achieves the best result over large search space.

\begin{table*}[htb]
    \centering
        \caption{Test accuracy with different execution plans for classification.}
    \resizebox{1.5\columnwidth}{!}{
    \begin{tabular}{l|cccccccc}
       \toprule
       Dataset & Plan 1 & Plan 2 & Plan 3 & Plan 4 & Plan 5 & TPOT & AUSK\\
       \hline
       puma8NH & 0.8275 & 0.8312 & 0.8271 & 0.8280 & 0.8303 & 0.8306 & \textbf{0.8325}\\
       kin8nm & 0.8808 & 0.8886 & 0.8886 & 0.8654 & \textbf{0.8910} & 0.8706 & 0.8834 \\
       cpu\_cmall & 0.9122 & \underline{\textbf{0.9126}} & 0.9126 & 0.9027 & \underline{\textbf{0.9127}} & 0.9121 & 0.9106 \\
       puma32H & 0.8849 & 0.8864 & 0.8848 & 0.8835 & 0.8894 & \textbf{0.8955} & 0.8830 \\
       cpu\_act & 0.9303 & \underline{\textbf{0.9315}} & 0.9305 & 0.9302 & 0.9309 & \underline{\textbf{0.9312}} & 0.9298 \\
       bank32nh & 0.7896 & 0.7889 & 0.7838 & 0.7891 & \underline{\textbf{0.7957}} & 0.7593 & \underline{\textbf{0.7957}}\\
       mc1 & 0.8796 & 0.8904 & 0.8722 & 0.8721 & \textbf{0.8975} & 0.8835 & 0.8896\\
       delta\_elevators & 0.8763 & 0.8760 & 0.8779 & 0.8766 & 0.8790 & \textbf{0.8835} & 0.8743\\
       jm1 & 0.6718 & 0.6721 & 0.6581 & 0.6473 & 0.6692 & 0.6415 & \textbf{0.6772}\\
       pendigits & 0.9932 & 0.9936 & 0.9929 & \textbf{0.9945} & 0.9937 & 0.9931 & \underline{\textbf{0.9944}}\\
       delta\_ailerons & 0.9235 & 0.9240 & 0.9242 & 0.9225 & 0.9259 & \textbf{0.9278} & 0.9259 \\
       wind & 0.8587 & 0.8589 & 0.8566 & 0.8583 & \textbf{0.8593} & 0.8542 & 0.8494\\
       satimage & 0.8961 & 0.8954 & 0.8965 & 0.8946 & \textbf{0.8981} & 0.8961 & 0.8793\\
       optdigits & 0.9889 & 0.9889 & 0.9883 & 0.9889 & 0.9889 & \textbf{0.9902} & 0.9818 \\
       phoneme & 0.8799 & 0.8832 & 0.8808 & 0.8791 & \textbf{0.8866} & 0.8812 & 0.8770\\ 
       spambase & 0.9401 & \textbf{0.9406} & 0.9379 & 0.9387 & 0.9386 & 0.9385 & 0.9358\\
       abalone & 0.6688 & 0.6679 & 0.6618 & 0.6614 & 0.6680 & \underline{\textbf{0.6748}} & \underline{\textbf{0.6751}}\\
       mammography & 0.8740 & 0.8783 & 0.8577 & 0.8755 & \textbf{0.8787} & 0.8568 & 0.8762\\
       waveform & 0.8948 & 0.8961 & 0.8900 & 0.8835 & 0.8952 & 0.8955 & \textbf{0.9040}\\
       pollen & 0.4934 & \underline{\textbf{0.5013}} & 0.5012 & \underline{\textbf{0.5013}} & \underline{\textbf{0.5013}} & 0.4961 & 0.4896\\
       \hline
       Average Rank & 4.30 & 2.98 & 4.80 & 5.13 & \textbf{2.58} & 3.83 & 4.40\\
       \bottomrule
    \end{tabular}
    }

    \label{tab:cls_plan}
\end{table*}

\begin{table*}[htb]
    \centering
        \caption{Test mean square error with different execution plans for regression.}
    \resizebox{1.5\columnwidth}{!}{
    \begin{tabular}{l|ccccccc}
       \toprule
       Dataset & Plan 1 & Plan 2 & Plan 3 & Plan 4 & Plan 5 & TPOT & AUSK\\
       \hline
       bank8FM & \underline{\textbf{0.0008}} & \underline{\textbf{0.0008}} & \underline{\textbf{0.0008}} & \underline{\textbf{0.0008}} & \underline{\textbf{0.0008}} & \underline{\textbf{0.0008}} & 0.0009 \\
       bank32nh & 0.0071 & \underline{\textbf{0.0069}} & 0.0070 & 0.0071 & \underline{\textbf{0.0069}} & \underline{\textbf{0.0069}} & 0.0070 \\
       kin8nm & 0.0067 & 0.0068 & 0.0073 & 0.0076 & \textbf{0.0066} & 0.0092 & 0.0148 \\
       puma8NH & 10.3020 & \textbf{10.0822} & 10.1293 & 10.1091 & 10.1698 & 10.1043 & 10.2109 \\
       cpu\_small & 7.3994 & 7.0854 & 7.0069 & 7.1741 & \textbf{7.0051} & 7.4058 & 8.7286 \\
       wind & 8.9650 & 8.9636 & 8.8993 & 9.2930 & \textbf{8.6976} & 8.8618 & 9.2261 \\
       cpu\_act & 5.0067 & 4.8762 & 4.7950 & 4.7983 & \textbf{4.7790} & 4.8373 & 6.4232 \\
       puma32H & 0.0001 & 0.0001 & 0.0001 & 0.0001 & \textbf{0.0000} & 0.0001 & 0.0001\\
       sulfur & \underline{\textbf{0.0002}} & \underline{\textbf{0.0002}} & \underline{\textbf{0.0002}} & \underline{\textbf{0.0002}} & \underline{\textbf{0.0002}} & 0.0003 & 0.0003\\
       space\_ga & 0.0115 & \textbf{0.0093} & 0.0098 & 0.0099 & 0.0098 & 0.0098 & 0.0108\\
       \hline
       Average Rank & 4.95 & 3.00 & 3.40 & 4.30 & \textbf{2.20} & 4.00 & 6.15\\
       \bottomrule
    \end{tabular}
    }

    \label{tab:rgs_plan}
\end{table*}

\subsection{Evaluations on Different Execution Plans}

To address the inefficiency of automated plan generation, we apply the execution plan that performs well on most tasks. 
Since there are lots of ways to decompose AutoML space, enumerating and evaluating each execution plan is impossible.
To avoid enumerating the entire search space, we list all the execution plans (a small plan set) in the coarse-grained level --- sub-task, and this small plan set covers both the frameworks in existing AutoML systems and the strategies used by human experts.
Concretely, we can obtain this small plan set by decomposing the search space according to the sub-tasks in AutoML: feature engineering, hyperparameter tuning and algorithm selection.
There are in total five execution plans (See Figure~\ref{fig:execution_plans}) --- \textit{J, C, A, AC, and CA}.
If we decompose the search space in a more fine-grained level, the plan set will be larger. 
Note that, most of the existing open-source AutoML systems, e.g., \texttt{auto-sklearn} and \texttt{TPOT} correspond to the execution plan --- plan 1 (\textit{J}).
We evaluate each of them on 20 classification tasks and 10 regression tasks, and the results can be found in Tables~\ref{tab:cls_plan} and ~\ref{tab:rgs_plan}.
We can have that the proposed execution plan used in \sys (i.e., plan 5 - \textit{CA}) outperforms the other alternatives on most tasks (a smaller average rank).
This demonstrates the effectiveness of the proposed execution plan over the potential competitors.

Furthermore, if we look at the performance of the existing open-source AutoML systems, i.e., \texttt{auto-sklearn} and \texttt{TPOT}, which fall into the execution plan 1 - \textit{J}. 
As shown in Tables~\ref{tab:cls_plan} and ~\ref{tab:rgs_plan}, we find that the execution plan used in \sys (i.e., \textit{CA}) outperforms the two AutoML systems on most tasks (a smaller average rank).

The test results are shown in Tables \ref{tab:cls_plan} and \ref{tab:rgs_plan}. 
We can have that the best execution plan varies over datasets. 
Surprisingly, we find that Plan 5, which is the execution plan introduced in \sys, achieves the first place on 16 of the total 30 tasks with an average rank of 2.45 (the lower, the better).
\texttt{TPOT} and \texttt{autosklearn} that use a single joint block achieves an average rank of 3.83 and 4.98 respectively.
Therefore, the proposed execution plan (Plan 5) in \sys sets up a very competitive AutoML baseline for our further research on \sys, e.g., automatic plan generation.

\begin{figure*}[htb]
	\centering
	\subfigure[Wind]{
		\scalebox{0.35}{
			\includegraphics[width=1\linewidth]{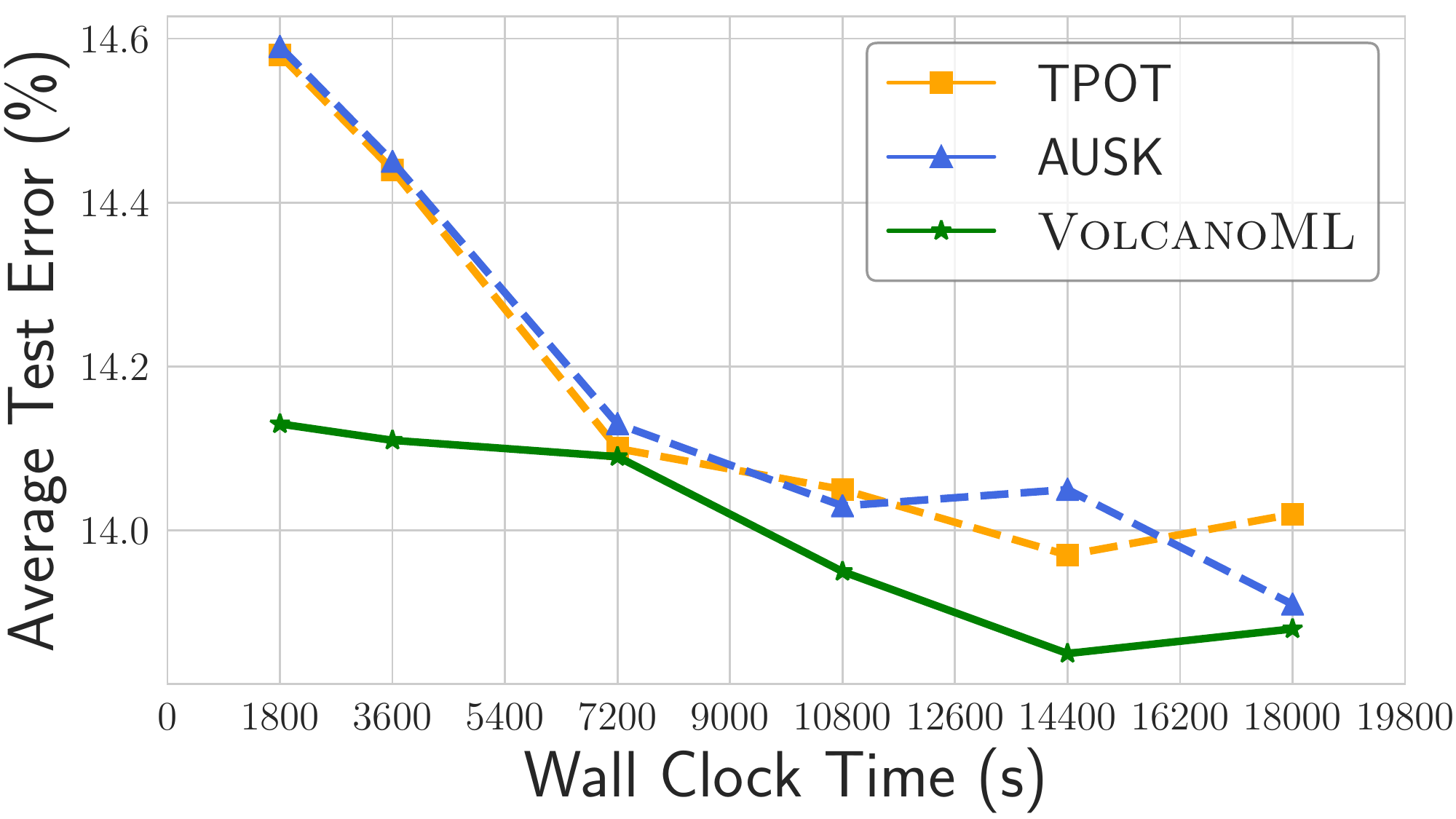}
			
	}}
	\subfigure[Kin8nm]{
		\scalebox{0.35}{
			\includegraphics[width=1\linewidth]{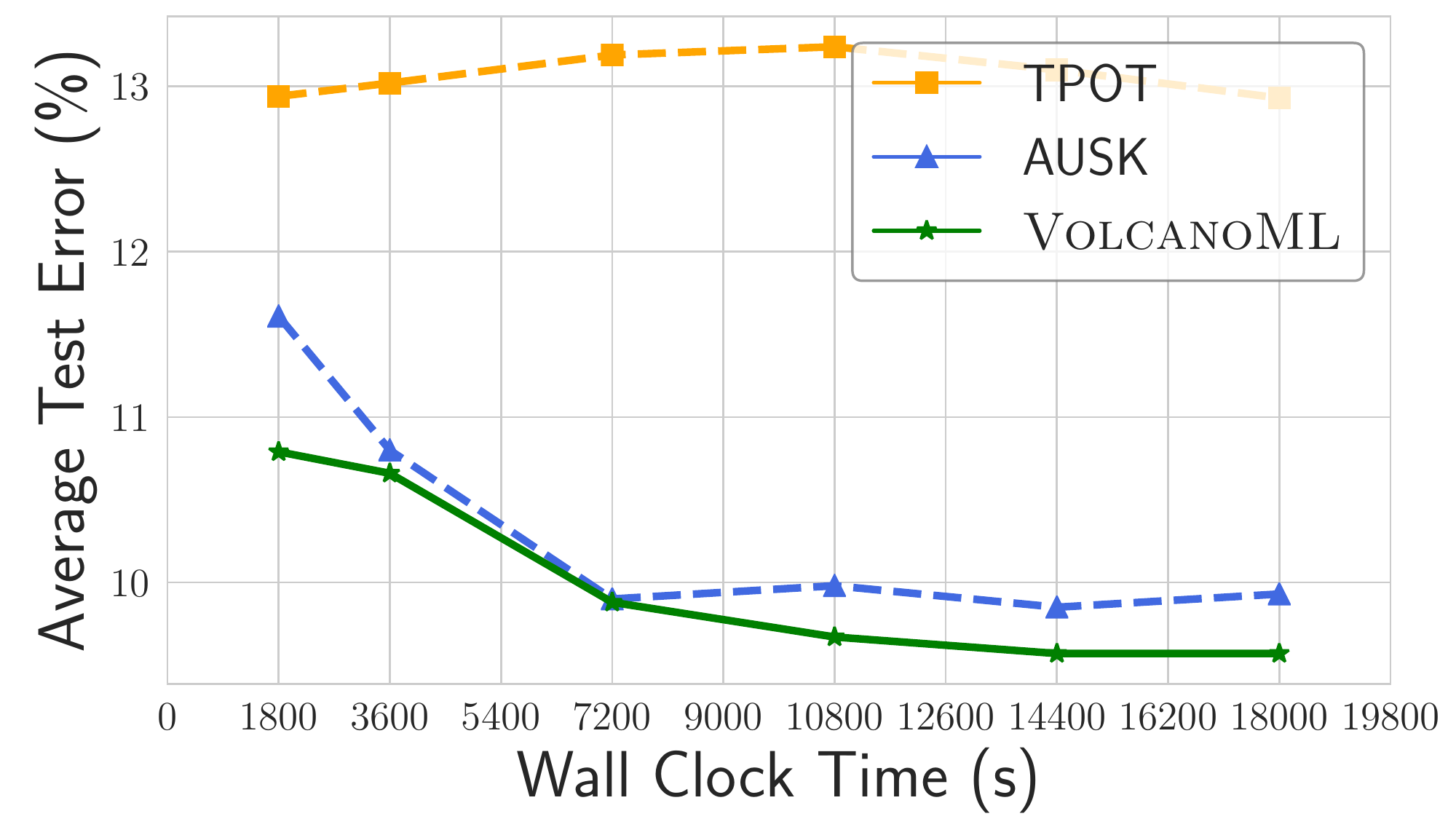}
	}}
	\subfigure[Cpu\_act]{
		\scalebox{0.35}{
			\includegraphics[width=1\linewidth]{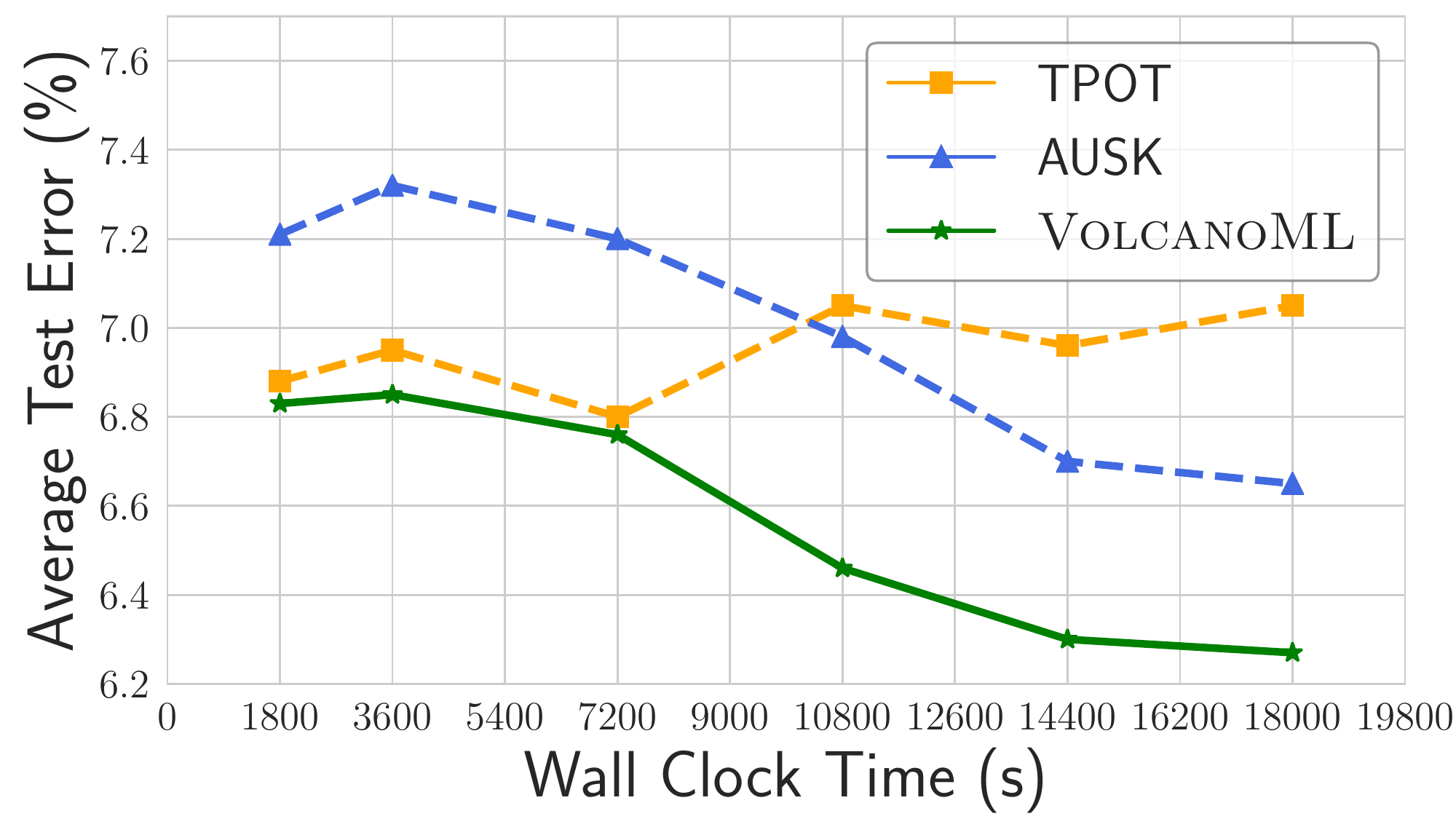}
	}}
    \subfigure[Spambase]{
		\scalebox{0.35}{
			\includegraphics[width=1\linewidth]{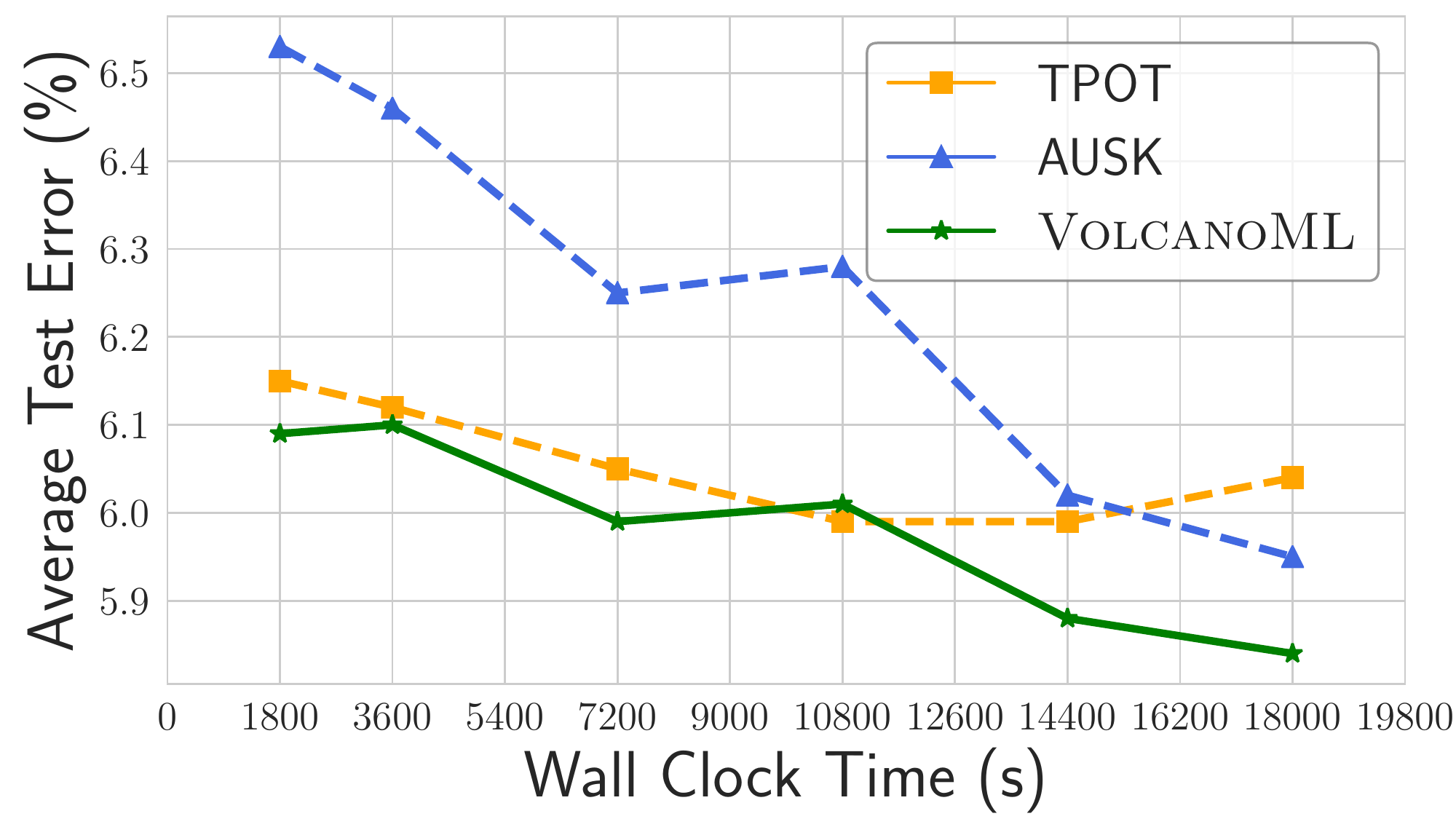}
	}}
	\caption{Test errors on four medium datasets given different time budgets.}
  \label{test_speedups}
\end{figure*}

\subsection{Additional Experiment Results}
\para{Comparison with early-stopping methods.}
Table~\ref{table:early-stopping} show the results for \sys compared with early-stopping based methods on five classification datasets and five regression datasets in Table~\ref{table:early-stopping}.
The datasets are of medium size, each of which contains 8192 samples.
The settings follow Section 6.1, and the compared baselines are Hyperband~\cite{li2018hyperband}, BOHB~\cite{falkner2018bohb}, and MFES-HB~\cite{li2020mfeshb}. 
Remind that \sys uses SMAC~\cite{hutter2011sequential} in the joint block by default.
While the execution plan in \sys is independent of optimization algorithms, we also implement \sys+, which applies the MFES-HB algorithm in the joint block.
From Table~\ref{table:early-stopping}, we observe that \sys with SMAC outperforms the three early-stopping methods, and the performance is further improved when we combine the benefits of both \sys execution plans and early-stopping optimization methods.

\begin{table}[t]
\centering
\resizebox{1\columnwidth}{!}{
\subtable[Classification]{
\begin{tabular}{lccccc}
    \toprule
    Dataset (ID) & \sys & \sys+ & HyperBand & BOHB & MFES-HB \\ 
    \hline
    puma8NH (816) & 83.03 & \textbf{83.12} & 83.01 & 82.91 & 82.96 \\
    kin8nm (807) & 89.10 & \textbf{89.14} & 88.28 & 88.70 & 89.12\\
    cpu\_small (735) & 91.27 & \textbf{91.33} & 90.97 & 91.08 & 91.14 \\
    puma32H (752) & 89.55 & 89.61 & 89.34 & 89.43 & \textbf{89.73} \\
    cpu\_act (761) & \textbf{93.12} & 93.01 & 92.88 & 92.96 & 92.97 \\
    \hline
    Average Rank & 2.2 & \textbf{1.4} & 4.6 & 4.2 & 2.6 \\
    \bottomrule
\end{tabular}
}
}
\resizebox{1\columnwidth}{!}{
\subtable[Regression]{
\begin{tabular}{lccccc}
    \toprule
    Dataset (ID) & \sys & \sys+ & HyperBand & BOHB & MFES-HB \\ 
    \hline
    puma8NH (225) & 10.1698 & \textbf{10.1642} & 10.1843 & 10.1654 & 10.2619 \\
    kin8nm (189) & \textbf{0.0066} & 0.0069 & 0.0081 & 0.0073 & 0.0072 \\
    cpu\_small (227) & \textbf{7.0051} & 7.1341 & 7.4657 & 7.5272 & 7.5363\\
    puma32H (308) & \underline{\textbf{0.0000}} & \underline{\textbf{0.0000}} & \underline{\textbf{0.0000}} & \underline{\textbf{0.0000}} & \underline{\textbf{0.0000}} \\
    cpu\_act (573) & 4.7790 & \textbf{4.7524} & 4.8778 & 5.2856 & 5.1506 \\
    \hline
    Average Rank & 2.0 & \textbf{1.8} & 3.6 & 3.6 & 4.0 \\
    \bottomrule
\end{tabular}
}
}
\caption {Test accuracy (\%) and test mean squared error of \sys compared with early-stopping methods. \sys+ refers to the combination of \sys with MFES-HB.}
\label{table:early-stopping}
\end{table}

\para{Results on large datasets.}
Table~\ref{table_large-cls} shows the results on ten large datasets with a budget of 18,000 seconds. \sys is the best on eight of them. Figure~\ref{test_speedups} shows the validation errors on four of those datasets. When achieving the same validation error compared with \tpot and \ausk, \sys obtains a speed-up of 4.3-10.5$\times$ and 4.8-11$\times$, respectively.

\begin{table}[ht]
\centering
\caption{Test balanced accuracy on 10 large datasets.}
\resizebox{0.75\columnwidth}{!}{
\begin{tabular}{lccc}
    \toprule
    Datasets & TPOT & AUSK & \sys  \\ 
    \midrule
    mnist\_784 & 0.9724 & 0.9701 & \textbf{0.9795} \\
    letter(2) & \underline{\textbf{0.9969}} & 0.9939 & \underline{\textbf{0.9969}} \\
    kropt & 0.8656 & 0.8267 & \textbf{0.8669} \\
    mv & \underline{\textbf{0.9997}} & 0.9994 & \underline{\textbf{0.9997}} \\
    a9a & 0.8129 & \textbf{0.8250} & 0.8215 \\
    covertype & 0.7124 & 0.7098 & \textbf{0.7152}\\
    2dplanes & 0.9291 & \textbf{0.9297} & 0.9293 \\
    higgs & 0.7235 & 0.7258 & \textbf{0.7279} \\
    electricity & \underline{\textbf{0.9327}} & 0.9226 & \underline{\textbf{0.9329}} \\
    fried & \underline{\textbf{0.9296}} & 0.9280 & \textbf{0.9300} \\

    \bottomrule
\end{tabular}
}
\label{table_large-cls}
\end{table}

\begin{figure}[thb]
\centering
\includegraphics[width=0.3\textwidth]{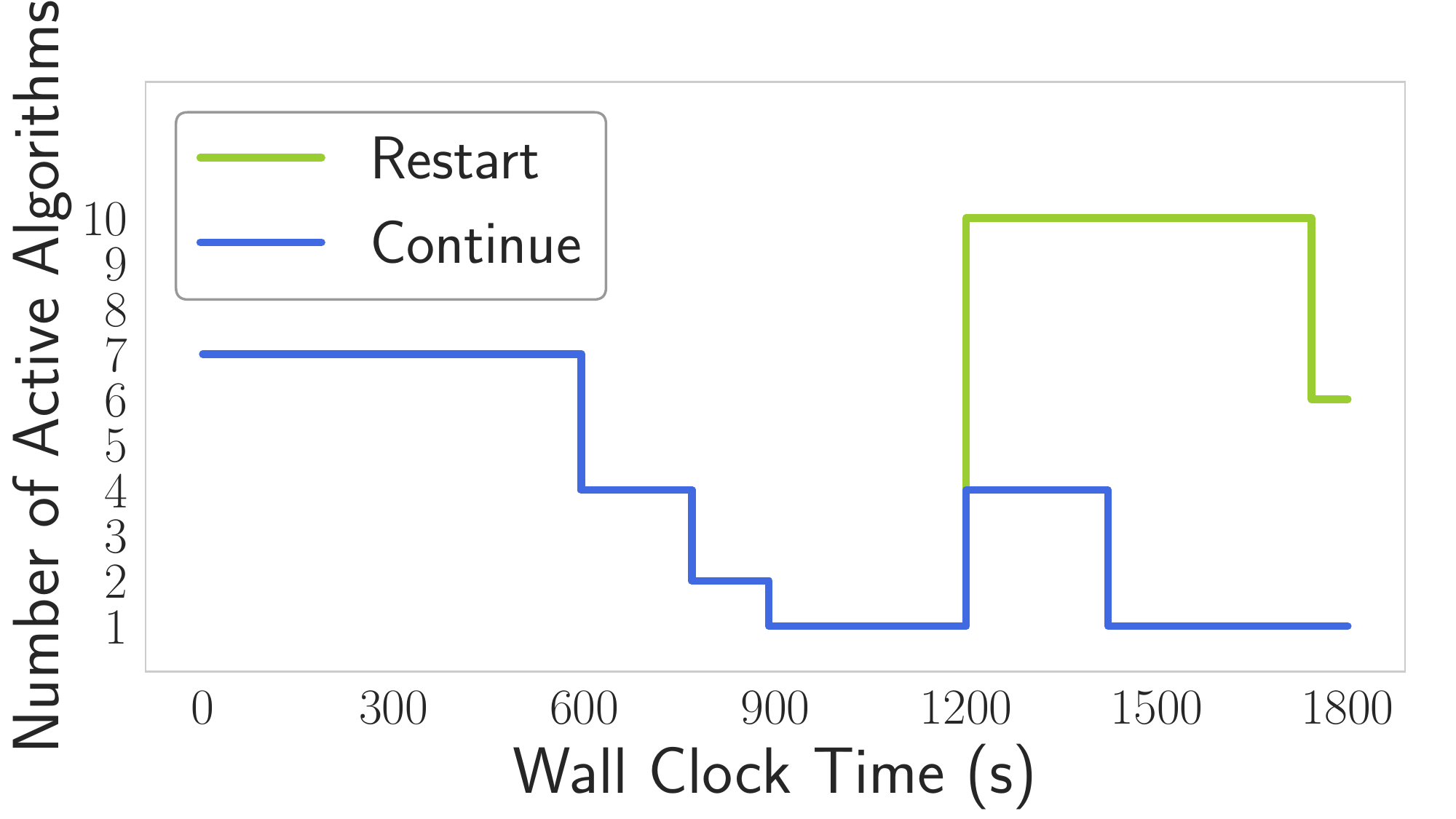}
\caption{The trend of the number of active algorithms in the conditional block on pc4.}
\label{fig:continue}
\end{figure}

\para{Continue tuning in conditional block.}
To show the process of continue tuning, we present a case study on the dataset pc4.
We add three algorithms (LightGBM, Extra Trees, and Liblinear SVC) after tuning 7 other algorithms in \sys for 1200 seconds.
The total budget is 1800 seconds.
The trend of the number of active blocks is plotted in Figure~\ref{fig:continue}.
When new algorithms come, \sys with restarting re-optimizes the extended search space, and it takes another 540s to reduce the number of active algorithms to 6.
For \sys with continue tuning, the number of active algorithms is 4 (1 survived + 3 added) when new algorithms are added, and it takes another 220 seconds to reduce the number to 1, which is LightGBM in the added algorithms.
As continue tuning avoids exploring the search space of those eliminated algorithms, \sys with continue tuning improves the test accuracy on pc4 to 86.44\% compared with 84.74\% achieved by \sys with restarting.

\para{Comparison with progressive methods.}
We also compare the progressive strategies with original ones on five classification tasks and five regression tasks. 
The settings follow Section 6.1 and the results are shown in Table~\ref{table:strategy}.
We observe that the original strategy outperforms the progressive one on 8 of the 10 tasks.
As a result, we apply it as the original strategy by default for \sys.

\begin{table}[t]
\centering
\resizebox{1\columnwidth}{!}{
\subtable[Classification]{
\begin{tabular}{lcc}
    \toprule
    Dataset (ID) & Original & Progressive \\ 
    \hline
    puma8NH (816) & \textbf{83.03} & 82.99  \\
    kin8nm (807) & \textbf{89.10} & 88.72 \\
    cpu\_small (735) & \textbf{91.27} & \textbf{91.27} \\
    puma32H (752) & \textbf{89.55} & 88.97 \\
    cpu\_act (761) & \textbf{93.12} & 93.09 \\
    \bottomrule
\end{tabular}
}
\subtable[Regression]{
\begin{tabular}{lcc}
    \toprule
    Dataset (ID) & Original & Progressive \\ 
    \hline
    puma8NH (225)& \textbf{10.1698} & 10.2437  \\
    kin8nm (189) & 0.0066 & \textbf{0.0065} \\
    cpu\_small (227)& \textbf{7.0051} & 7.2181 \\
    puma32H (308) & \textbf{0.0000} & 0.0001 \\
    cpu\_act (573) & \textbf{4.7790} & 4.8321 \\
    \bottomrule
\end{tabular}
}
}
\caption {Test accuracy (\%) and test mean squared error for two optimization strategies on classification and regression tasks.}
\label{table:strategy}
\end{table}
\section{Conclusion}
\label{sec:conclusion}

In this paper, we have presented \sys, a scalable and extensible framework that allows users to design decomposition strategies for large AutoML search spaces in an expressive and flexible manner.
\sys introduces novel building blocks akin to relational operators in database systems that enable expressing search space decomposition strategies in a \emph{structured} fashion -- similar to relational execution plans.
Moreover, \sys introduces a Volcano-style execution model, inspired by its classic counterpart that has been widely used for relational query evaluation, to execute the decomposition strategies it yields.
Experimental evaluation demonstrates that \sys can generate more efficient decomposition strategies that also lead to performance-wise better ML pipelines, compared to state-of-the-art AutoML systems.

\begin{acknowledgements}
This work is supported by the National Natural Science Foundation of China (NSFC No.61832001, U1936104), Beijing Academy of Artificial Intelligence (BAAI) and PKU-Tencent Joint Research Lab. Bin Cui is the corresponding authors. 

Ce Zhang and the DS3Lab gratefully acknowledge the support from the Swiss National Science Foundation (Project Number 200021\_184628), Innosuisse/SNF BRIDGE Discovery (Project Number 40B2-0\_187132), European Union Horizon 2020 Research and Innovation Programme (DAPHNE, 957407), Botnar Research Centre for Child Health, Swiss Data Science Center, Alibaba, Cisco, eBay,
Google Focused Research Awards, Oracle Labs.
\end{acknowledgements}

\appendix
\section{Appendix}
\label{sec:appendix}

In this section, we describe more details about the background, system design and implementations.

\subsection{AutoML Formulations and Motivations}

\subsubsection{Formulations}

\para{Definition and Notation.} 
There are $K$ candidate algorithms $\mathcal{A}=\{A^1, ..., A^K\}$. 
Each algorithm $A^i$ has a corresponding hyper-parameter space $\Lambda_i$.
The algorithm $A^i$ with hyper-parameter configuration $\lambda$ and new feature set $F$ is denoted by $A^i_{(\lambda,F)}$. 
Given the dataset $D=\{D_{train}, D_{valid}\}$ of a learning problem, the AutoML problem is to find the joint algorithm, feature, and hyper-parameter configuration $A^{\ast}_{(\lambda^{\ast},F^{\ast})}$ that minimizes the loss metric (e.g., the validation error on $D_{valid}$):
\begin{equation}
    A^\ast_{(\lambda^\ast, F^{\ast})} = \operatornamewithlimits{argmin}_{A^i\in\mathcal{A},\lambda\in\Lambda^i,F\in\mathcal{F}^i} \mathcal{L}(A_{(\lambda, F)}^i; D),
\label{automl_joint}
\end{equation}
where $\mathcal{F}^i=Gen(A^i, D, \mathbf{op})$ is the feature space of $A^i$ that can be generated from the raw feature (data) set $D$, and $\mathbf{op}$ is the set of available FE operators.

\begin{figure}
\begin{center}
\centerline{\includegraphics[width=0.65\columnwidth]{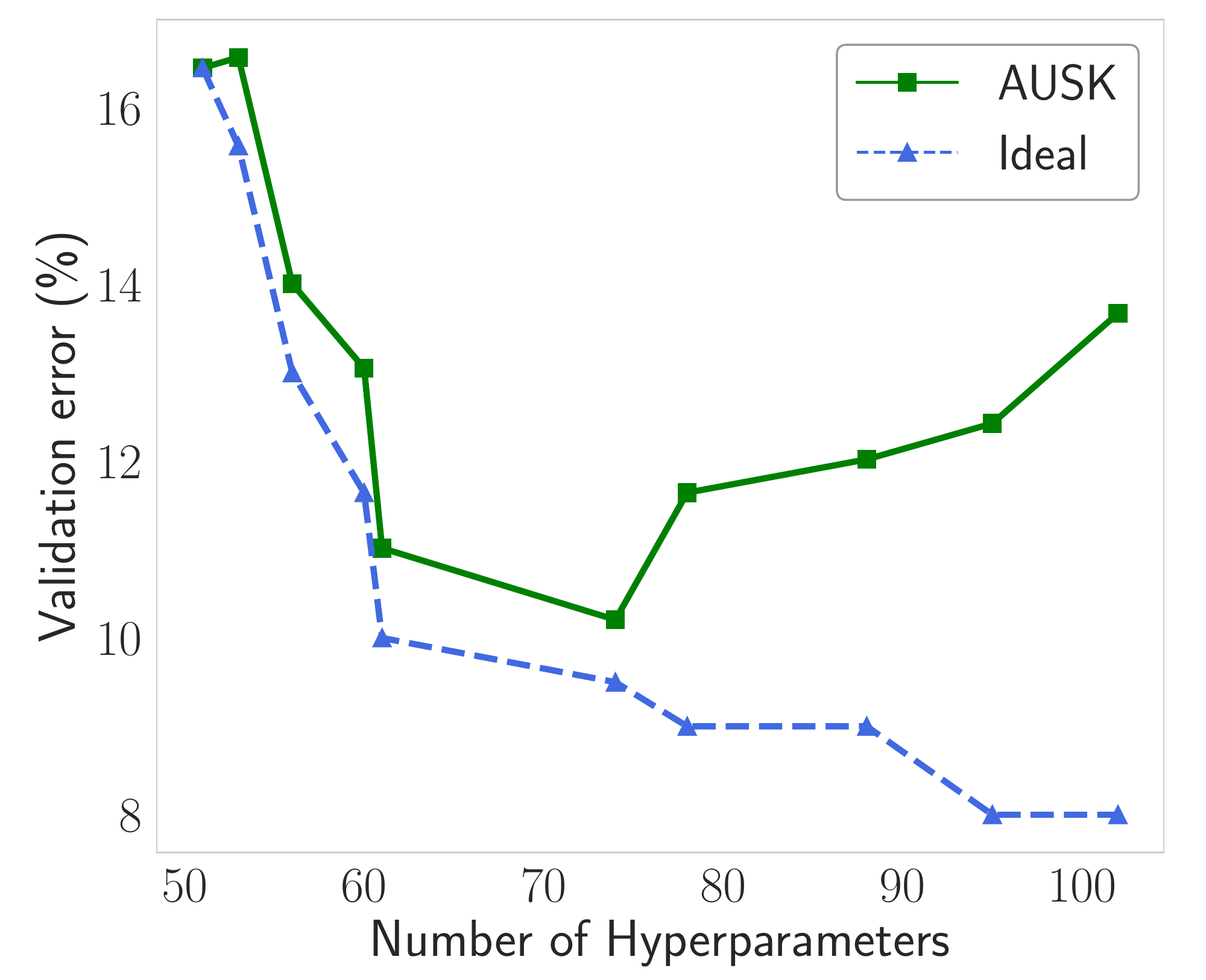}}
\caption{Validation error on \texttt{pc4} when increasing the number of hyper-parameters in \ausk given the same time budget.}
\label{fig:trend}
\end{center}
\end{figure}

\para{Challenge: Ever-growing Search Space.} 
Enriching the search space can lead to performance improvement since the enriched search space may bring better configurations. 
However, an ever-growing search space can significantly increase the complexity of searching for ML pipelines.
Existing AutoML systems usually can only explore very limited configurations in a huge search space, and thus suffer from the low-efficiency issue~\cite{li2020efficient} that hampers the effectiveness of AutoML systems.
In Figure~\ref{fig:trend}, we provide a brief example of \ausk, one state-of-the-art system AutoML system.
Its search algorithm cannot scale to a high-dimensional search space~\cite{li2020efficient}. 
To alleviate this issue, in this paper we focus on developing a scalable AutoML system.

\begin{figure}
\begin{center}
\centerline{\includegraphics[width=0.65\columnwidth]{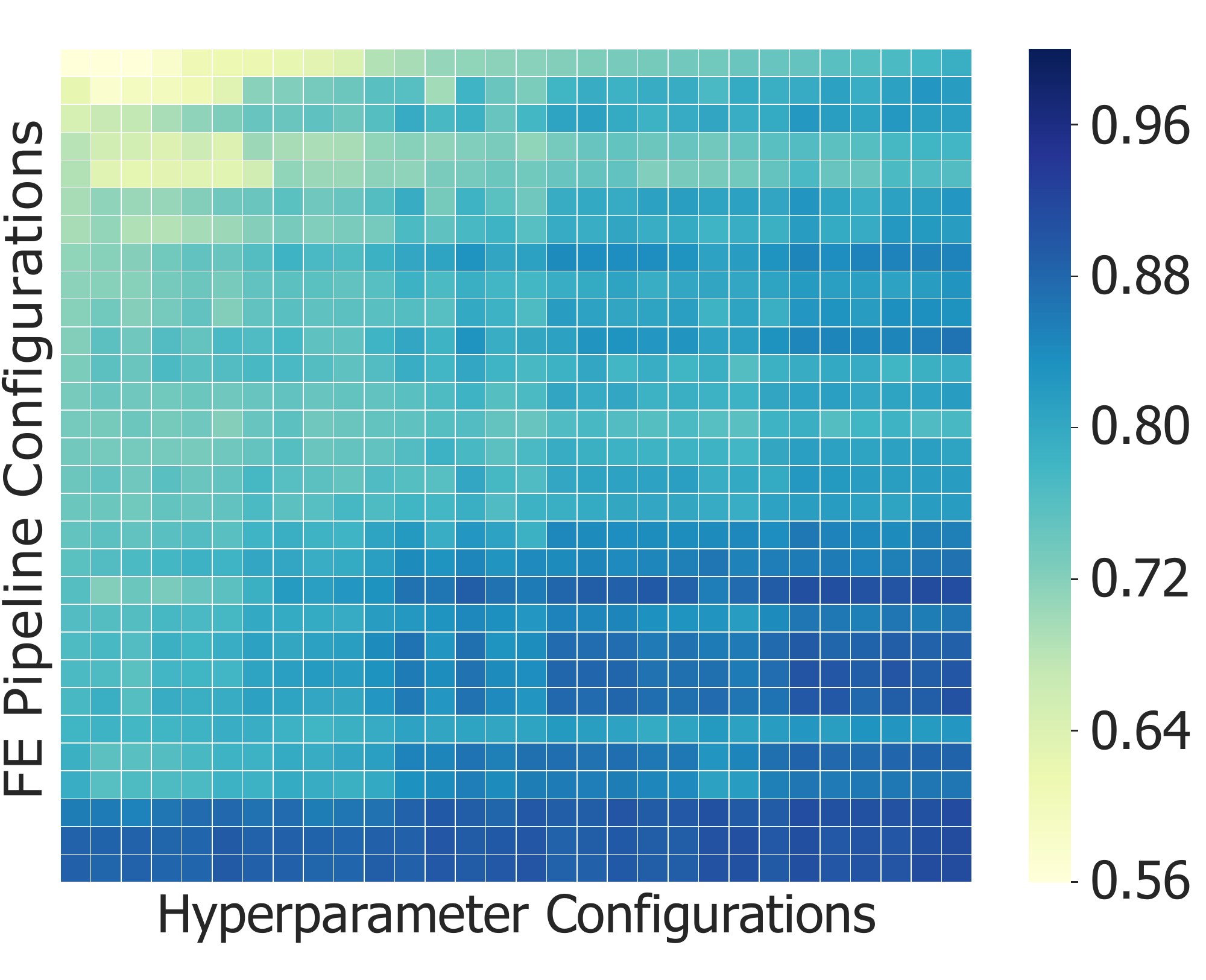}}
\caption{The performance distribution of ML pipelines constructed by 30 FE and HPO configurations on \texttt{fri\_c1} using Random Forest. For FE configurations, the performance increases from top to down; for HPO configurations, the performance increases from left to right (The deeper, the better).}
\label{fig:heatmap}
\end{center}
\end{figure}

\subsubsection{Observations and Motivations about AutoML}
\label{appendix:obs}
We now present several important observations that inspired the design of \sys.

\para{Observation 1.} The search space can be partitioned according to ML algorithms. 
The entire search space is the union of the search spaces of individual algorithms, i.e., $\Omega=\{S^1, ..., S^K\}$, where $S^i$ is the joint space of features and hyper-parameters, i.e., $S^i=(\Lambda^i \times \mathcal{F}^i)$.

\para{Observation 2.} The sub-space of algorithm $A^i$ can be very large, e.g., in \ausk, $S^i$ usually includes more than $50$ hyper-parameters.
When exploring the search spaces via extensive experiments, we observe the following:
\begin{itemize}
    \item If hyper-parameter configuration $\lambda_1$ performs better than $\lambda_2$, i.e., $\lambda_1 \le \lambda_2$, then it often holds that $(\lambda_1, F) \le (\lambda_2, F)$ for the joint configuration $(\lambda, F)$ with $F$ fixed; 
    \item If FE pipeline configuration $F_1$ performs better than $F_2$, i.e., $F_1\le F_2$, then it often holds that $(\lambda, F_1) \le (\lambda, F_2)$ for the joint configuration $(\lambda, F)$ with $\lambda$ fixed.
\end{itemize}
Figure~\ref{fig:heatmap} presents an example for these observations.
This motivates us to solve the joint FE and HPO problem via \emph{alternating} optimization.
That is, we can alternate between optimizing FE and HPO, and we can fix the FE configuration (resp. HPO configuration) when optimizing for HPO (resp. FE).
This alternating manner is indeed similar to how human experts solve the joint optimization problem manually. 
One obvious advantage of alternating optimization is that each time only a much smaller subspace ($\Lambda^i$ or $\mathcal{F}^i$) needs to be optimized, instead of the joint space $S^i=(\Lambda^i \times \mathcal{F}^i)$.


\para{Observation 3.} The sensitivity of ML algorithms to FE and HPO is often different.
Taking Figure~\ref{fig:heatmap} for example, compared to HPO, FE has a larger influence on the performance of `\textit{Random Forest}' on `\texttt{fri\_c1}'; in this case, optimizing FE \emph{more frequently} can bring more performance improvement.

\para{Observation 4.} The above observations motivate the use of \emph{meta-learning}. We can learn (1) the algorithm performance across ML tasks and (2) the configuration selection of each ML algorithm across tasks. 
Such meta-knowledge obtained from historical tasks can greatly improve the efficiency of ML pipeline search.

Therefore, a scalable AutoML system should include two basic components: (1) an efficient framework that can navigate in a huge search space, and (2) a meta-learning module that can extract knowledge from previously ML tasks and apply it to new tasks.

\subsection{\sys Components and Implementations}
\label{appendix:system_components}

\subsubsection{Compenents and Search Space}
\para{Feature Engineering.} The feature engineering pipeline is shown in Figure~\ref{fe_pipeline}. It comprises four sequential stages: \emph{preprocessors} (compulsory), \emph{scalers} (5 possible operators), \emph{balancers} (1 possible operators) and \emph{feature transformers} (13 possible operstors).
For each of the latter three stages, \sys picks one operator and then execute the entire pipeline.
Table~\ref{fe_space} presents the details of each operator.
The total number of hyper-parameters for FE is 52.

We follow the design of the search space for feature engineering in the existing AutoML systems, e.g.,  \texttt{autosklearn} and \texttt{TPOT}. 
It limits the search space for feature engineering by adopting a fixed pipeline including different stages, and each stage is equipped with an operation (featurizer) that is selected from a pool of featurizers.
The pool of featurizers at each stage is relatively small, and Bayesian optimization can be used to choose the proper featurizers for each stage.
When the pool of featurizers is very large, high-dimensional Bayesian optimization algorithms could work better.
In many real-world cases, though this architecture is not good enough for feature engineering (effectiveness), there still remains space to explore to conduct feature engineering effectively and efficiently.
To support real scenarios, \sys provides API for user-defined feature engineering operators and we recommend users to add domain-specific feature engineering operators to the search space for better search performance. 
In addition, users can replace the original feature engineering part in \sys with other iterative feature engineering methods easily.

\para{ML Algorithms.} \sys implements 11 algorithms for classification and 10 algorithms for regression, with a total of 50 and 49 hyper-parameters respectively. 
The built-in algorithms include \emph{linear models}, \emph{support vector machine}, \emph{discriminant analysis}, \emph{nearest neighbors}, and \emph{ensembles}.
Table~\ref{algo_space} presents the details.

\para{Ensemble Methods.} Ensembles that combine predictions from multiple base models have been known to outperform individual models, often drastically reducing the variance of the final predictions~\cite{Dietterich2000}. 
\sys provides four ensemble methods: \emph{bagging}, \emph{blending}, \emph{stacking}, and \emph{ensemble selection}~\cite{Caruana2004}.
During the search process, the top $N_{\text{top}}$ configurations for each algorithm are recorded and the corresponding models are stored. 
After the optimization budget exhausts, 
the saved models are treated as the base models for the ensemble method. 
We use \emph{ensemble selection} as the default method and build an ensemble of size 50.

\begin{table}[h]
\centering
\caption {Hyper-parameters of ML algorithms in \sys. We distinguish categorical (cat) hyper-parameters from numerical (cont) ones. The numbers in the brackets are conditional hyper-parameters.}
\resizebox{0.9\columnwidth}{!}{
\begin{tabular}{p{95pt}ccc}
    \toprule
    Type of Classifier & \#$\lambda$ & cat (cond) & cont (cond) \\ 
    \midrule
    AdaBoost & 4 & 1 (-) & 3 (-) \\
    Random forest & 5 & 2 (-) & 3 (-) \\
    Extra trees & 5 & 2 (-) & 3 (-) \\
    Gradient boosting & 7 & 1 (-) & 6 (-) \\
    KNN & 2 & 1 (-) & 1 (-) \\
    LDA & 4 & 1 (-) & 3 (1) \\
    QDA & 1 & - & 1 (-) \\
    Logistic regression & 4 & 2 (-) & 2 (-) \\
    Liblinear SVC & 5 & 2 (2) & 3 (-) \\
    LibSVM SVC & 7 & 2 (2) & 5 (-) \\
    LightGBM & 6 & - & 6 (-) \\
    \bottomrule
\end{tabular}
}
\\
\resizebox{0.9\columnwidth}{!}{
\begin{tabular}{p{95pt}ccc}
    \toprule
    Type of Regressor & \#$\lambda$ & cat (cond) & cont (cond) \\ 
    \midrule
    AdaBoost & 4 & 1 (-) & 3 (-) \\
    Random forest & 5 & 2 (-) & 3 (-) \\
    Extra trees & 5 & 2 (-) & 3 (-) \\
    Gradient boosting & 7 & 1 (-) & 6 (-) \\
    KNN & 2 & 1 (-) & 1 (-) \\
    Lasso & 3 & - & 3 (-) \\
    Ridge & 4 & 1 (-) & 3 (-) \\
    Liblinear SVC & 5 & 2 (2) & 3 (-) \\
    LibSVM SVC & 8 & 3 (3) & 5 (-) \\
    LightGBM & 6 & - & 6 (-) \\
    \bottomrule
\end{tabular}
}

\label{algo_space}
\end{table}

\begin{table}[h]
\centering
\caption{Hyper-parameters of FE operators in \sys.}
\resizebox{0.9\columnwidth}{!}{
\begin{tabular}{p{95pt}ccc}
    \toprule
    Type of Operator & \#$\lambda$ & cat (cond) & cont (cond) \\ 
    \midrule
    One-hot encoder & 0 & - & - \\
    Imputer & 1 & - & 1 (-) \\
    \midrule
    Minmax & 0 & - & - \\
    Normalizer & 0 & - & - \\
    Quantile & 2 & 1 (-) & 1 (-) \\
    Robust & 2 & - & 2 (-) \\
    Standard & 0 & - & - \\
    \midrule
    Weight Balancer & 0 & - & - \\
    \midrule
    Cross features & 1 & - & 1 (-) \\
    Fast ICA & 4 & 3 (1) & 1 (1) \\
    Feature agglomeration & 4 & 3 (2) & 1 (-) \\
    Kernel PCA & 5 & 1 (1) & 4 (3) \\
    Rand. kitchen sinks & 2 & - & 2 (-) \\
    LDA decomposer & 1 & 1 (-) & - \\
    Nystroem sampler & 5 & 1 (1) & 4 (3) \\
    PCA & 2 & 1 (-) & 1 (-) \\
    Polynomial & 2 & 1 (-) & 1 (-) \\
    Random trees embed. & 5 & 1 (-) & 4 (-) \\
    \midrule
    SVD & 1 & - & 1 (-) \\
    Select percentile & 2 & 1 (-) & 1 (-) \\
    Select generic univariate & 3 & 2 (-) & 1 (-) \\
    Extra trees preprocessing & 5 & 2 (-) & 3 (-) \\
    Linear SVM preprocessing & 5 & 3 (3) & 2 (-) \\
    \bottomrule
\end{tabular}
}

\label{fe_space}
\end{table}

\subsubsection{Programming Interface}
Consider a tabular dataset of raw values in a CSV file, named \texttt{train.csv}, where the last column represents the label.
We take a classification task as an example. With \sys, only six lines of code are needed for searching and model evaluation.

\begin{lstlisting}[language=Python]
from ... import DataManager, Classifier
dm = DataManager()
train_node = dm.load_train('train.csv')
test_node = dm.load_test('test.csv')
clf = Classifier(**params).fit(train_node)
predictions = clf.predict(test_node)
\end{lstlisting}

By calling \texttt{load\_train} and \texttt{load\_test}, the \emph{data manager} automatically identifies the type of each feature (continuous, discrete, or categorical), imputes missing values, and converts string-like features to one-hot vectors. 
By calling \texttt{fit}, \sys splits the dataset into folds for training and validation, evaluates various configurations, and generates an ensemble from each individual configuration. For users who need to customize the search process, \texttt{Classifier} provides additional parameters to specify:
\begin{itemize}
    \item \texttt{time\_limit} controls the total runtime of the search process;
    \item \texttt{include\_algorithms} specifies which algorithms are included (if not specified, all built-in algorithms are included);
    \item \texttt{ensemble\_method} chooses which ensemble strategy to use;
    \item \texttt{enable\_meta} determines whether to use meta-learning to accelerate the search process;
    \item \texttt{metric} specifies the metric used to evaluate the performance of each configuration.
\end{itemize}

\para{Customized Components.} \sys provides APIs to easily enrich the search space, such as the stage in FE pipeline, FE operators, and ML algorithms. 
The following is the syntax of defining customized components:

\begin{lstlisting}[language=Python]
from ... import add_classifier
from ... import update_FEPipeline
from ... import BaseModel, BaseOperator

# Add new ML algorithm.
class CustomizedModel(BaseModel):
    def fit(x,y): ...
    def predict(x): ...
    def get_search_space(): ...

add_classifier(CustomizedModel)

# Add new FE operator.
class CustomizedOP(BaseOperator):
    def operate(x): ...
    def get_search_space(): ...

# Customize FE pipeline.
update_FEPipeline([`new_stage', ...],
    {`new_stage': [CustomizedOP], 
    ...}) 
\end{lstlisting}

It is important to note that, \ausk does not support adding a new stage or updating the existing stages in the FE pipeline.
In addition, \ausk cannot add an operator for any stage (e.g., adding \texttt{smote\_balancer} to the stage \texttt{balancer}), while \sys supports this.

\subsection{Experiment Datasets}
\label{appendix:results}

In our experiments, we splitted each dataset into five folds. Four are used for training  and the remaining one is used for testing.
The 60 OpenML datasets used are presented as follows (in the form of ``dataset\_name (OpenML id)''):

\para{Classification Datasets.}
kc1 (1067), quake (772), segment (36), ozone-level-8hr (1487), space\_ga (737), sick (38), pollen (871), analcatdata\_supreme (728), abalone (183), spambase (44), waveform(2) (979), phoneme (1489), page-blocks(2)  (1021), optdigits(28), satimage (182), wind (847), delta\_ailerons (803), puma8NH (816), kin8nm (807),
puma32H (752), cpu\_act (761), bank32nh (833), mc1 (1056), delta\_elevators (819), jm1 (1053), pendigits (32), mammography (310), ailerons (734), eeg (1471), letter(2) (977), kropt (184), mv (881), fried (901), 2dplanes (727), electricity (151), a9a (A2), mnist\_784 (554), higgs (23512), covertype (180).

\para{Regression Datasets.}
stock (223), socmob (541), Moneyball (41021), insurance (A1), weather\_izmir (42369), us\_crime (315), debutanizer (23516), space\_ga (507), pollen (529), wind (503), bank8FM (572), bank32nh (558), kin8nm (189), puma8NH (225), cpu\_act (573), puma32H (308), cpu\_small (227), visualizing\_soil (668), sulfur (23515), rainfall\_bangladesh (41539).

Since the datasets \texttt{insurance} and \texttt{a9a} are not collected in OpenML, we use A1 and A2 as their OpenML ID instead.

%
%

\bibliographystyle{spmpsci}
\bibliography{reference}

\begin{thebibliography}{10}
\providecommand{\url}[1]{{#1}}
\providecommand{\urlprefix}{URL }
\expandafter\ifx\csname urlstyle\endcsname\relax
  \providecommand{\doi}[1]{DOI~\discretionary{}{}{}#1}\else
  \providecommand{\doi}{DOI~\discretionary{}{}{}\begingroup
  \urlstyle{rm}\Url}\fi

\bibitem{aguilar2021ease}
Aguilar~Melgar, L., Dao, D., Gan, S., G{\"u}rel, N.M., Hollenstein, N., Jiang,
  J., Karla{\v{s}}, B., Lemmin, T., Li, T., Li, Y., et~al.: Ease. ml: A
  lifecycle management system for machine learning.
\newblock In: Proceedings of the Annual Conference on Innovative Data Systems
  Research (CIDR), 2021. CIDR (2021)

\bibitem{bai2022autodc}
Bai, Y., Li, Y., Shen, Y., Yang, M., Zhang, W., Cui, B.: Autodc: an automatic
  machine learning framework for disease classification.
\newblock Bioinformatics  (2022)

\bibitem{bardenet2013collaborative}
Bardenet, R., Brendel, M., K{\'e}gl, B., Sebag, M.: Collaborative
  hyperparameter tuning.
\newblock In: International conference on machine learning, pp. 199--207. PMLR
  (2013)

\bibitem{barnes2015azure}
Barnes, J.: Azure machine learning.
\newblock Microsoft Azure Essentials. 1st ed, Microsoft  (2015)

\bibitem{baylor2017tfx}
Baylor, D., Breck, E., Cheng, H.T., Fiedel, N., Foo, C.Y., Haque, Z., Haykal,
  S., Ispir, M., Jain, V., Koc, L., et~al.: Tfx: A tensorflow-based
  production-scale machine learning platform.
\newblock In: Proceedings of the 23rd ACM SIGKDD International Conference on
  Knowledge Discovery and Data Mining, pp. 1387--1395 (2017)

\bibitem{bergstra2012random}
Bergstra, J., Bengio, Y.: Random search for hyper-parameter optimization.
\newblock Journal of Machine Learning Research \textbf{13}, 281--305 (2012)

\bibitem{bergstra2011algorithms}
Bergstra, J.S., Bardenet, R., Bengio, Y., K{\'e}gl, B.: Algorithms for
  hyper-parameter optimization.
\newblock In: Advances in neural information processing systems, pp. 2546--2554
  (2011)

\bibitem{Boehm2019}
Boehm, M., Antonov, I., Baunsgaard, S., Dokter, M., Ginth{\"o}r, R.,
  Innerebner, K., Klezin, F., Lindstaedt, S., Phani, A., Rath, B., et~al.:
  Systemds: A declarative machine learning system for the end-to-end data
  science lifecycle.
\newblock arXiv preprint arXiv:1909.02976  (2019)

\bibitem{TFX}
Breck, E., Polyzotis, N., Roy, S., Whang, S., Zinkevich, M.: Data validation
  for machine learning.
\newblock In: MLSys (2019)

\bibitem{ranknet}
Burges, C.: From ranknet to lambdarank to lambdamart: An overview.
\newblock Learning \textbf{11} (2010)

\bibitem{caroe1999dual}
Car{\o}E, C.C., Schultz, R.: Dual decomposition in stochastic integer
  programming.
\newblock Operations Research Letters \textbf{24}(1-2), 37--45 (1999)

\bibitem{Caruana2004}
Caruana, R., Niculescu-Mizil, A., Crew, G., Ksikes, A.: {Ensemble selection
  from libraries of models}.
\newblock In: Proceedings, Twenty-First International Conference on Machine
  Learning, ICML 2004 (2004).
\newblock \doi{10.1145/1015330.1015432}

\bibitem{Chen2018}
Chen, B., Wu, H., Mo, W., Chattopadhyay, I., Lipson, H.: Autostacker: A
  compositional evolutionary learning system.
\newblock In: Proceedings of the Genetic and Evolutionary Computation
  Conference, pp. 402--409 (2018)

\bibitem{DeepDive}
De~Sa, C., Ratner, A., R{\'e}, C., Shin, J., Wang, F., Wu, S., Zhang, C.:
  Deepdive: Declarative knowledge base construction.
\newblock ACM SIGMOD Record \textbf{45}(1), 60--67 (2016)

\bibitem{dechter1998bucket}
Dechter, R.: Bucket elimination: A unifying framework for probabilistic
  inference.
\newblock In: Learning in graphical models, pp. 75--104. Springer (1998)

\bibitem{dewancker2016strategy}
Dewancker, I., McCourt, M., Clark, S., Hayes, P., Johnson, A., Ke, G.: A
  strategy for ranking optimization methods using multiple criteria.
\newblock In: Workshop on Automatic Machine Learning, pp. 11--20. PMLR (2016)

\bibitem{Dietterich2000}
Dietterich, T.G.: {Ensemble methods in machine learning}.
\newblock In: Lecture Notes in Computer Science (including subseries Lecture
  Notes in Artificial Intelligence and Lecture Notes in Bioinformatics) (2000).
\newblock \doi{10.1007/3-540-45014-9_1}

\bibitem{Drori2018}
Drori, I., Krishnamurthy, Y., Rampin, R., De, R., Lourenco, P., Ono, J.P., Cho,
  K., Silva, C., Freire, J.: {AlphaD3M: Machine Learning Pipeline Synthesis}.
\newblock AutoML Workshop at ICML  (2018)

\bibitem{efimova2017fast}
Efimova, V., Filchenkov, A., Shalamov, V.: Fast automated selection of learning
  algorithm and its hyperparameters by reinforcement learning.
\newblock In: International Conference on Machine Learning AutoML Workshop
  (2017)

\bibitem{eggensperger2013towards}
Eggensperger, K., Feurer, M., Hutter, F., Bergstra, J., Snoek, J., Hoos, H.,
  Leyton-Brown, K.: Towards an empirical foundation for assessing bayesian
  optimization of hyperparameters.
\newblock In: NIPS workshop on Bayesian Optimization in Theory and Practice,
  vol.~10, p.~3 (2013)

\bibitem{falkner2018bohb}
Falkner, S., Klein, A., Hutter, F.: Bohb: Robust and efficient hyperparameter
  optimization at scale.
\newblock In: International Conference on Machine Learning, pp. 1437--1446.
  PMLR (2018)

\bibitem{feurer2015efficient}
Feurer, M., Klein, A., Eggensperger, K., Springenberg, J., Blum, M., Hutter,
  F.: Efficient and robust automated machine learning.
\newblock In: Advances in neural information processing systems, pp. 2962--2970
  (2015)

\bibitem{feurer2018scalable}
Feurer, M., Letham, B., Bakshy, E.: Scalable meta-learning for bayesian
  optimization using ranking-weighted gaussian process ensembles.
\newblock In: AutoML Workshop at ICML (2018)

\bibitem{10.5555/1450931}
Garcia-Molina, H., Ullman, J.D., Widom, J.: Database Systems: The Complete
  Book, 2 edn.
\newblock Prentice Hall Press, USA (2008)

\bibitem{Ghoting2011}
Ghoting, A., Krishnamurthy, R., Pednault, E., Reinwald, B., Sindhwani, V.,
  Tatikonda, S., Tian, Y., Vaithyanathan, S.: Systemml: Declarative machine
  learning on mapreduce.
\newblock In: 2011 IEEE 27th International Conference on Data Engineering, pp.
  231--242. IEEE (2011)

\bibitem{golovin2017google}
Golovin, D., Solnik, B., Moitra, S., Kochanski, G., Karro, J., Sculley, D.:
  Google vizier: A service for black-box optimization.
\newblock In: Proceedings of the 23rd ACM SIGKDD International Conference on
  Knowledge Discovery and Data Mining, pp. 1487--1495. ACM (2017)

\bibitem{GooPre}
Google: Google prediction api,.
\newblock \url{https://developers.google.com/prediction} (2020)

\bibitem{Graefe1994}
Graefe, G.: {Volcano—An Extensible and Parallel Query Evaluation System}.
\newblock IEEE Transactions on Knowledge and Data Engineering  (1994)

\bibitem{he2020automl}
He, X., Zhao, K., Chu, X.: Automl: A survey of the state-of-the-art.
\newblock Knowledge-Based Systems \textbf{212}, 106622 (2021)

\bibitem{hu2019multi}
Hu, Y.Q., Yu, Y., Tu, W.W., Yang, Q., Chen, Y., Dai, W.: Multi-fidelity
  automatic hyper-parameter tuning via transfer series expansion.
\newblock AAAI  (2019)

\bibitem{Hutter2014}
Hutter, F., Hoos, H., Leyton-Brown, K.: {An efficient approach for assessing
  hyperparameter importance}.
\newblock In: 31st International Conference on Machine Learning, ICML 2014
  (2014)

\bibitem{hutter2011sequential}
Hutter, F., Hoos, H.H., Leyton-Brown, K.: Sequential model-based optimization
  for general algorithm configuration.
\newblock In: International Conference on Learning and Intelligent
  Optimization, pp. 507--523. Springer (2011)

\bibitem{automl_book}
Hutter, F., Kotthoff, L., Vanschoren, J. (eds.): Automated Machine Learning:
  Methods, Systems, Challenges.
\newblock Springer (2018).
\newblock In press, available at http://automl.org/book.

\bibitem{Hutter2015}
Hutter, F., L{\"u}cke, J., Schmidt-Thieme, L.: Beyond manual tuning of
  hyperparameters.
\newblock KI-K{\"u}nstliche Intelligenz \textbf{29}(4), 329--337 (2015)

\bibitem{IBMc}
IBM: Ibmwatson studio autoai.
\newblock \url{https://www.ibm.com/cloud/watson-studio/autoai} (2020)

\bibitem{jamieson2016non}
Jamieson, K., Talwalkar, A.: Non-stochastic best arm identification and
  hyperparameter optimization.
\newblock In: Artificial Intelligence and Statistics, pp. 240--248 (2016)

\bibitem{jiang2021automated}
Jiang, H., Shen, Y., Li, Y.: Automated hyperparameter optimization challenge at
  cikm 2021 analyticcup.
\newblock arXiv preprint arXiv:2111.00513  (2021)

\bibitem{jones1998efficient}
Jones, D.R., Schonlau, M., Welch, W.J.: Efficient global optimization of
  expensive black-box functions.
\newblock Journal of Global optimization \textbf{13}(4), 455--492 (1998)

\bibitem{kandasamy2017multi}
Kandasamy, K., Dasarathy, G., Schneider, J., P{\'o}czos, B.: Multi-fidelity
  bayesian optimisation with continuous approximations.
\newblock In: International Conference on Machine Learning, pp. 1799--1808.
  PMLR (2017)

\bibitem{kanter2015deep}
Kanter, J.M., Veeramachaneni, K.: Deep feature synthesis: Towards automating
  data science endeavors.
\newblock In: 2015 {IEEE} International Conference on Data Science and Advanced
  Analytics, DSAA 2015, Paris, France, October 19-21, 2015, pp. 1--10. IEEE
  (2015)

\bibitem{Katz2017}
Katz, G., Shin, E.C.R., Song, D.: Explorekit: Automatic feature generation and
  selection.
\newblock In: 2016 IEEE 16th International Conference on Data Mining (ICDM),
  pp. 979--984. IEEE (2016)

\bibitem{Kaul2017}
Kaul, A., Maheshwary, S., Pudi, V.: Autolearn—automated feature generation
  and selection.
\newblock In: 2017 IEEE International Conference on data mining (ICDM), pp.
  217--226. IEEE (2017)

\bibitem{Khurana2018}
Khurana, U., Samulowitz, H., Turaga, D.: {Feature engineering for predictive
  modeling using reinforcement learning}.
\newblock In: 32nd AAAI Conf. Artif. Intell. AAAI 2018 (2018)

\bibitem{Khurana2016}
Khurana, U., Turaga, D., Samulowitz, H., Parthasrathy, S.: Cognito: Automated
  feature engineering for supervised learning.
\newblock In: 2016 IEEE 16th International Conference on Data Mining Workshops
  (ICDMW), pp. 1304--1307. IEEE (2016)

\bibitem{kleinfbhh17}
Klein, A., Falkner, S., Bartels, S., Hennig, P., Hutter, F.: Fast bayesian
  optimization of machine learning hyperparameters on large datasets.
\newblock In: Proceedings of the 20th International Conference on Artificial
  Intelligence and Statistics, pp. 528--536 (2017)

\bibitem{komer2014hyperopt}
Komer, B., Bergstra, J., Eliasmith, C.: Hyperopt-sklearn: automatic
  hyperparameter configuration for scikit-learn.
\newblock In: ICML workshop on AutoML, vol.~9. Citeseer (2014)

\bibitem{NorthStar}
Kraska, T.: Northstar: An interactive data science system.
\newblock Proceedings of the VLDB Endowment \textbf{11}(12), 2150--2164 (2018)

\bibitem{ActiveClean}
Krishnan, S., Wang, J., Wu, E., Franklin, M.J., Goldberg, K.: Activeclean:
  Interactive data cleaning for statistical modeling.
\newblock Proceedings of the VLDB Endowment \textbf{9}(12), 948--959 (2016)

\bibitem{ledell2020h2o}
LeDell, E., Poirier, S.: H2o automl: Scalable automatic machine learning.
\newblock In: Proceedings of the AutoML Workshop at ICML, vol. 2020 (2020)

\bibitem{levine2017rotting}
Levine, N., Crammer, K., Mannor, S.: Rotting bandits.
\newblock In: Advances in NIPS, pp. 3074--3083 (2017)

\bibitem{li2018hyperband}
Li, L., Jamieson, K., DeSalvo, G., Rostamizadeh, A., Talwalkar, A.: Hyperband:
  A novel bandit-based approach to hyperparameter optimization.
\newblock Proceedings of the International Conference on Learning
  Representations pp. 1--48 (2018)

\bibitem{automl2}
Li, T., Zhong, J., Liu, J., Wu, W., Zhang, C.: Ease. ml: Towards multi-tenant
  resource sharing for machine learning workloads.
\newblock Proceedings of the VLDB Endowment \textbf{11}(5), 607--620 (2018)

\bibitem{li2020efficient}
Li, Y., Jiang, J., Gao, J., Shao, Y., Zhang, C., Cui, B.: Efficient automatic
  cash via rising bandits.
\newblock In: AAAI, pp. 4763--4771 (2020)

\bibitem{li2022transfer}
Li, Y., Shen, Y., Jiang, H., Bai, T., Zhang, W., Zhang, C., Cui, B.: Transfer
  learning based search space design for hyperparameter tuning.
\newblock Proceedings of the 28th ACM SIGKDD Conference on Knowledge Discovery
  \& Data Mining  (2022)

\bibitem{li2022hyper}
Li, Y., Shen, Y., Jiang, H., Zhang, W., Li, J., Liu, J., Zhang, C., Cui, B.:
  Hyper-tune: Towards efficient hyper-parameter tuning at scale.
\newblock Proceedings of the VLDB Endowment \textbf{15} (2022)

\bibitem{li2022transbo}
Li, Y., Shen, Y., Jiang, H., Zhang, W., Yang, Z., Zhang, C., Cui, B.: Transbo:
  Hyperparameter optimization via two-phase transfer learning.
\newblock Proceedings of the 28th ACM SIGKDD Conference on Knowledge Discovery
  \& Data Mining  (2022)

\bibitem{li2020mfeshb}
Li, Y., Shen, Y., Jiang, J., Gao, J., Zhang, C., Cui, B.: Mfes-hb: Efficient
  hyperband with multi-fidelity quality measurements.
\newblock In: Proceedings of the AAAI Conference on Artificial Intelligence,
  vol.~35, pp. 8491--8500 (2021)

\bibitem{openbox}
Li, Y., Shen, Y., Zhang, W., Chen, Y., Jiang, H., Liu, M., Jiang, J., Gao, J.,
  Wu, W., Yang, Z., et~al.: Openbox: A generalized black-box optimization
  service.
\newblock In: Proceedings of the 27th ACM SIGKDD Conference on Knowledge
  Discovery \& Data Mining, pp. 3209--3219 (2021)

\bibitem{li2021volcanoml}
Li, Y., Shen, Y., Zhang, W., Jiang, J., Ding, B., Li, Y., Zhou, J., Yang, Z.,
  Wu, W., Zhang, C., et~al.: Volcanoml: Speeding up end-to-end automl via
  scalable search space decomposition.
\newblock Proceedings of the VLDB Endowment  (2021)

\bibitem{liaw2018tune}
Liaw, R., Liang, E., Nishihara, R., Moritz, P., Gonzalez, J.E., Stoica, I.:
  Tune: A research platform for distributed model selection and training.
\newblock arXiv preprint arXiv:1807.05118  (2018)

\bibitem{liberty2020elastic}
Liberty, E., Karnin, Z., Xiang, B., Rouesnel, L., Coskun, B., Nallapati, R.,
  Delgado, J., Sadoughi, A., Astashonok, Y., Das, P., et~al.: Elastic machine
  learning algorithms in amazon sagemaker.
\newblock In: Proceedings of the 2020 ACM SIGMOD International Conference on
  Management of Data, pp. 731--737 (2020)

\bibitem{liu2018progressive}
Liu, C., Zoph, B., Neumann, M., Shlens, J., Hua, W., Li, L.J., Fei-Fei, L.,
  Yuille, A., Huang, J., Murphy, K.: Progressive neural architecture search.
\newblock In: Proceedings of the European Conference on Computer Vision (ECCV),
  pp. 19--34 (2018)

\bibitem{liu2019automated}
Liu, S., Ram, P., Bouneffouf, D., Bramble, G., Conn, A.R., Samulowitz, H.,
  Gray, A.G.: An admm based framework for automl pipeline configuration pp.
  4892--4899 (2020)

\bibitem{Mohr2018}
Mohr, F., Wever, M., H{\"u}llermeier, E.: Ml-plan: Automated machine learning
  via hierarchical planning.
\newblock Machine Learning \textbf{107}(8), 1495--1515 (2018)

\bibitem{Moritz2007}
Moritz, P., Nishihara, R., Wang, S., Tumanov, A., Liaw, R., Liang, E., Elibol,
  M., Yang, Z., Paul, W., Jordan, M.I., et~al.: Ray: A distributed framework
  for emerging $\{$AI$\}$ applications.
\newblock In: 13th $\{$USENIX$\}$ Symposium on Operating Systems Design and
  Implementation ($\{$OSDI$\}$ 18), pp. 561--577 (2018)

\bibitem{Krypton}
Nakandala, S., Kumar, A., Papakonstantinou, Y.: Incremental and approximate
  inference for faster occlusion-based deep cnn explanations.
\newblock In: Proceedings of the 2019 International Conference on Management of
  Data, pp. 1589--1606 (2019)

\bibitem{Cerebro}
Nakandala, S., Zhang, Y., Kumar, A.: Cerebro: A data system for optimized deep
  learning model selection.
\newblock Proceedings of the VLDB Endowment \textbf{13}(12), 2159--2173 (2020)

\bibitem{Nargesian2017}
Nargesian, F., Samulowitz, H., Khurana, U., Khalil, E.B., Turaga, D.S.:
  Learning feature engineering for classification.
\newblock In: Ijcai, pp. 2529--2535 (2017)

\bibitem{olson2019tpot}
Olson, R.S., Moore, J.H.: Tpot: A tree-based pipeline optimization tool for
  automating machine learning.
\newblock In: Automated Machine Learning, pp. 151--160. Springer (2019)

\bibitem{poloczek2017multi}
Poloczek, M., Wang, J., Frazier, P.: Multi-information source optimization.
\newblock In: Advances in Neural Information Processing Systems, pp. 4288--4298
  (2017)

\bibitem{snorkel}
Ratner, A., et~al.: Snorkel: Rapid training data creation with weak
  supervision.
\newblock PVLDB  (2017)

\bibitem{Holoclean}
Rekatsinas, T., Chu, X., Ilyas, I.F., R{\'e}, C.: Holoclean: Holistic data
  repairs with probabilistic inference.
\newblock Proceedings of the VLDB Endowment \textbf{10}(11) (2017)

\bibitem{nni}
Research, M.: Microsoft nni.
\newblock \url{https://github.com/Microsoft/nni} (2020)

\bibitem{DeSa2017}
de~S{\'{a}}, A.G., Pinto, W.J.G., Oliveira, L.O.V., Pappa, G.L.: {RECIPE: A
  grammar-based framework for automatically evolving classification pipelines}.
\newblock In: Lecture Notes in Computer Science (including subseries Lecture
  Notes in Artificial Intelligence and Lecture Notes in Bioinformatics) (2017)

\bibitem{app1}
Schawinski, K., et~al.: Generative adversarial networks recover features in
  astrophysical images of galaxies beyond the deconvolution limit.
\newblock MNRAS Letters  (2017)

\bibitem{sen2018noisy}
Sen, R., Kandasamy, K., Shakkottai, S.: Noisy blackbox optimization with
  multi-fidelity queries: A tree search approach.
\newblock arXiv preprint arXiv:1810.10482  (2018)

\bibitem{bo_survey}
Shahriari, B., Swersky, K., Wang, Z., Adams, R.P., De~Freitas, N.: Taking the
  human out of the loop: A review of bayesian optimization.
\newblock Proceedings of the IEEE \textbf{104}(1), 148--175 (2015)

\bibitem{smith2020machine}
Smith, M.J., Sala, C., Kanter, J.M., Veeramachaneni, K.: The machine learning
  bazaar: Harnessing the ml ecosystem for effective system development.
\newblock In: Proceedings of the 2020 ACM SIGMOD International Conference on
  Management of Data, pp. 785--800 (2020)

\bibitem{snoek2012practical}
Snoek, J., Larochelle, H., Adams, R.P.: Practical bayesian optimization of
  machine learning algorithms.
\newblock In: Advances in neural information processing systems, pp. 2951--2959
  (2012)

\bibitem{swersky2013multi}
Swersky, K., Snoek, J., Adams, R.P.: Multi-task bayesian optimization.
\newblock In: Advances in neural information processing systems, pp. 2004--2012
  (2013)

\bibitem{takeno2020multifidelity}
Takeno, S., Fukuoka, H., Tsukada, Y., Koyama, T., Shiga, M., Takeuchi, I.,
  Karasuyama, M.: Multi-fidelity bayesian optimization with max-value entropy
  search and its parallelization.
\newblock In: International Conference on Machine Learning, pp. 9334--9345.
  PMLR (2020)

\bibitem{ThoHutHooLey13-AutoWEKA}
Thornton, C., Hutter, F., Hoos, H.H., Leyton-Brown, K.: Auto-weka: Combined
  selection and hyperparameter optimization of classification algorithms.
\newblock In: Proceedings of the 19th ACM SIGKDD international conference on
  Knowledge discovery and data mining, pp. 847--855 (2013)

\bibitem{VanRijn2018}
Van~Rijn, J.N., Hutter, F.: Hyperparameter importance across datasets.
\newblock In: Proceedings of the 24th ACM SIGKDD International Conference on
  Knowledge Discovery \& Data Mining, pp. 2367--2376 (2018)

\bibitem{DBLP:journals/corr/abs-1810-03548}
Vanschoren, J.: Meta-learning: {A} survey.
\newblock CoRR \textbf{abs/1810.03548} (2018).
\newblock \urlprefix\url{http://arxiv.org/abs/1810.03548}

\bibitem{10.1145/2641190.2641198}
Vanschoren, J., Van~Rijn, J.N., Bischl, B., Torgo, L.: Openml: networked
  science in machine learning.
\newblock ACM SIGKDD Explorations Newsletter \textbf{15}(2), 49--60 (2014)

\bibitem{modeldb}
Vartak, M., et~al.: Modeldb: A system for machine learning model management.
\newblock In: HILDA (2016)

\bibitem{Vilalta2002}
Vilalta, R., Drissi, Y.: {A perspective view and survey of meta-learning}.
\newblock Artificial Intelligence Review  (2002).
\newblock \doi{10.1023/A:1019956318069}

\bibitem{wang2013bayesian}
Wang, Z., Zoghi, M., Hutter, F., Matheson, D., De~Freitas, N.: Bayesian
  optimization in high dimensions via random embeddings.
\newblock In: Twenty-Third International Joint Conference on Artificial
  Intelligence (2013)

\bibitem{wistuba2016two}
Wistuba, M., Schilling, N., Schmidt-Thieme, L.: Two-stage transfer surrogate
  model for automatic hyperparameter optimization.
\newblock In: Joint European conference on machine learning and knowledge
  discovery in databases, pp. 199--214. Springer (2016)

\bibitem{wu2019practical}
Wu, J., Toscano-Palmerin, S., Frazier, P.I., Wilson, A.G.: Practical
  multi-fidelity bayesian optimization for hyperparameter tuning.
\newblock In: Uncertainty in Artificial Intelligence, pp. 788--798. PMLR (2020)

\bibitem{ZeroER}
Wu, R., Chaba, S., Sawlani, S., Chu, X., Thirumuruganathan, S.: Zeroer: Entity
  resolution using zero labeled examples.
\newblock In: Proceedings of the 2020 ACM SIGMOD International Conference on
  Management of Data, pp. 1149--1164 (2020)

\bibitem{Query20}
Wu, W., Flokas, L., Wu, E., Wang, J.: Complaint-driven training data debugging
  for query 2.0.
\newblock In: Proceedings of the 2020 ACM SIGMOD International Conference on
  Management of Data, pp. 1317--1334 (2020)

\bibitem{automl}
Yao, Q., Wang, M., Chen, Y., Dai, W., Li, Y.F., Tu, W.W., Yang, Q., Yu, Y.:
  Taking human out of learning applications: A survey on automated machine
  learning.
\newblock arXiv preprint arXiv:1810.13306  (2018)

\bibitem{mlflow}
Zaharia, M., et~al.: Accelerating the machine learning lifecycle with mlflow.
\newblock IEEE Data Eng. Bull.  (2018)

\bibitem{zhang2021facilitating}
Zhang, X., Chang, Z., Li, Y., Wu, H., Tan, J., Li, F., Cui, B.: Facilitating
  database tuning with hyper-parameter optimization: A comprehensive
  experimental evaluation.
\newblock Proceedings of the VLDB Endowment  (2021)

\bibitem{zhang2022towards}
Zhang, X., Wu, H., Li, Y., Tan, J., Li, F., Cui, B.: Towards dynamic and safe
  configuration tuning for cloud databases.
\newblock Proceedings of the 2022 ACM SIGMOD International Conference on
  Management of Data  (2022)

\bibitem{DBLP:journals/corr/abs-1904-12054}
Z{\"o}ller, M.A., Huber, M.F.: Benchmark and survey of automated machine
  learning frameworks.
\newblock Journal of artificial intelligence research \textbf{70}, 409--472
  (2021)

\end{thebibliography}


%
%


\end{document}